\documentclass{article}

\usepackage{microtype}
\usepackage{graphicx}
\usepackage{booktabs} 

\usepackage{hyperref}
\usepackage{multicol}

 \usepackage[accepted]{icml2023}

\usepackage{amsmath}
\usepackage{amssymb}
\usepackage{mathtools}
\usepackage{amsthm}

\usepackage[capitalize,noabbrev]{cleveref}
\usepackage{caption}
\usepackage{subcaption}

\theoremstyle{plain}
\newtheorem{theorem}{Theorem}[section]

\theoremstyle{definition}
\newtheorem{definition}[theorem]{Definition}

\theoremstyle{remark}

\DeclareMathOperator*{\argmax}{arg\,max}
\DeclareMathOperator*{\argmin}{arg\,min}

\usepackage[textsize=tiny]{todonotes}
\usepackage{enumitem}
\setlist[itemize]{leftmargin=*}

\newcommand{\benchname}{MultiRobustBench}

\usepackage{color}

\icmltitlerunning{\benchname: Benchmarking Robustness Against Multiple Attacks}

\begin{document}

\twocolumn[
\icmltitle{\benchname: Benchmarking Robustness Against Multiple Attacks}



\icmlsetsymbol{equal}{*}

\begin{icmlauthorlist}
\icmlauthor{Sihui Dai}{princeton}
\icmlauthor{Saeed Mahloujifar}{princeton}
\icmlauthor{Chong Xiang}{princeton}
\icmlauthor{Vikash Sehwag}{princeton}
\icmlauthor{Pin-Yu Chen}{ibm}
\icmlauthor{Prateek Mittal}{princeton}
\end{icmlauthorlist}

\icmlaffiliation{princeton}{Electrical and Computer Engineering, Princeton University}
\icmlaffiliation{ibm}{IBM Research}

\icmlcorrespondingauthor{Sihui Dai}{sihuid@princeton.edu}

\icmlkeywords{Machine Learning, ICML}

\vskip 0.3in
]



\printAffiliationsAndNotice{}  

\begin{abstract}
The bulk of existing research in defending against adversarial examples
focuses on defending against a single (typically bounded $\ell_p$-norm) attack, but for a practical setting, machine learning (ML) models should be robust to a wide variety of attacks. In this paper, we present the first unified framework for considering multiple attacks against ML models. Our framework is able to model different levels of learner's knowledge about the test-time adversary, allowing us to model robustness against unforeseen attacks and robustness against unions of attacks. Using our framework, we present the first leaderboard, \benchname~ (\url{https://multirobustbench.github.io}), for benchmarking multiattack evaluation which captures performance across attack types and attack strengths. We evaluate the performance of 16 defended models for robustness against a set of 9 different attack types, including $\ell_p$-based threat models, spatial transformations, and color changes, at 20 different attack strengths (180 attacks total). Additionally, we analyze the state of current defenses against multiple attacks.  Our analysis shows that while existing defenses have made progress in terms of average robustness across the set of attacks used, robustness against the worst-case attack is still a big open problem as all existing models perform worse than random guessing.
\end{abstract}

\section{Introduction}

\begin{figure*}[th]
    \centering
    \begin{subfigure}[t]{0.48\textwidth}
         \centering
         \includegraphics[width=\textwidth]{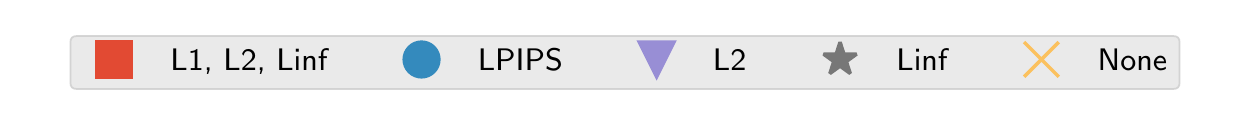}
         \vspace{-30pt}
     \end{subfigure}
     
    \begin{subfigure}[t]{0.32\textwidth}
    \includegraphics[width=\textwidth]{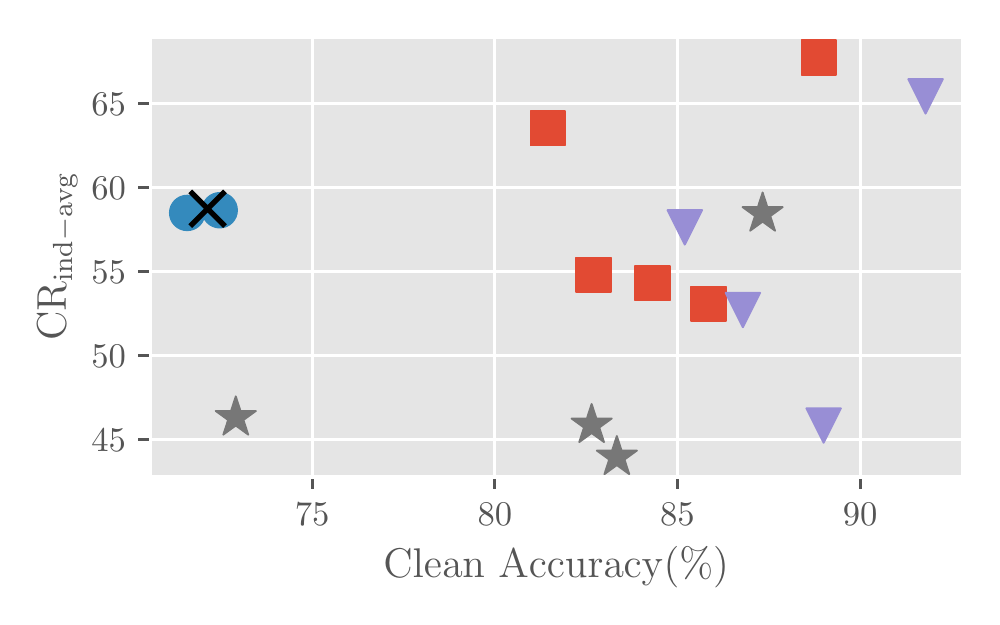}
    \caption{Clean accuracy vs $\text{CR}_{\text{ind-avg}}$}
    \label{subfig:clean_avg}
    \end{subfigure}
    \begin{subfigure}[t]{0.32\textwidth}
    \includegraphics[width=\textwidth]{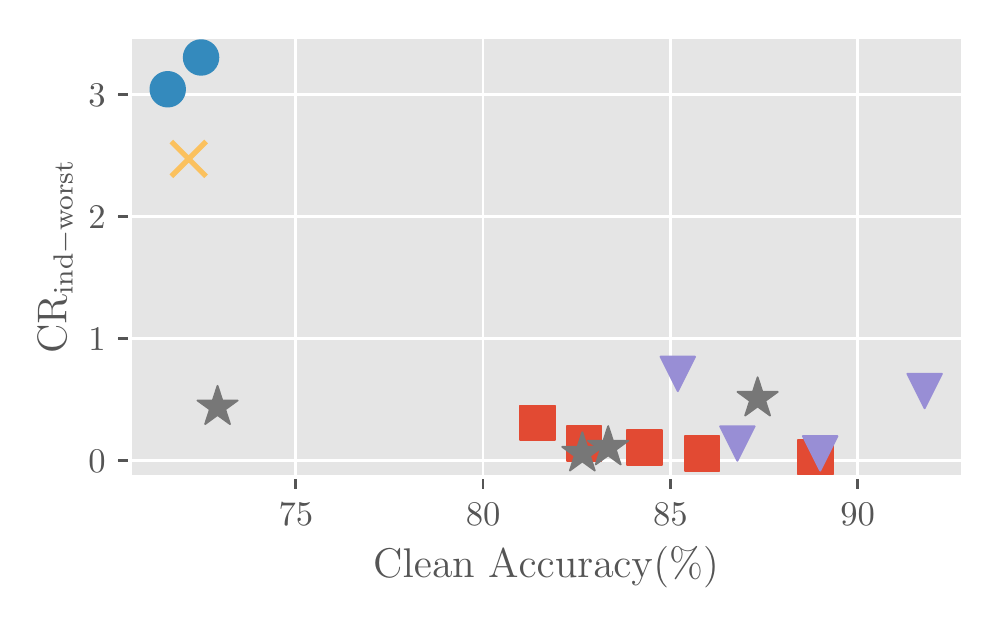}
    \caption{Clean accuracy vs $\text{CR}_{\text{ind-worst}}$}
        \label{subfig:clean_worst}
    \end{subfigure}
    \begin{subfigure}[t]{0.32\textwidth}
    \includegraphics[width=\textwidth]{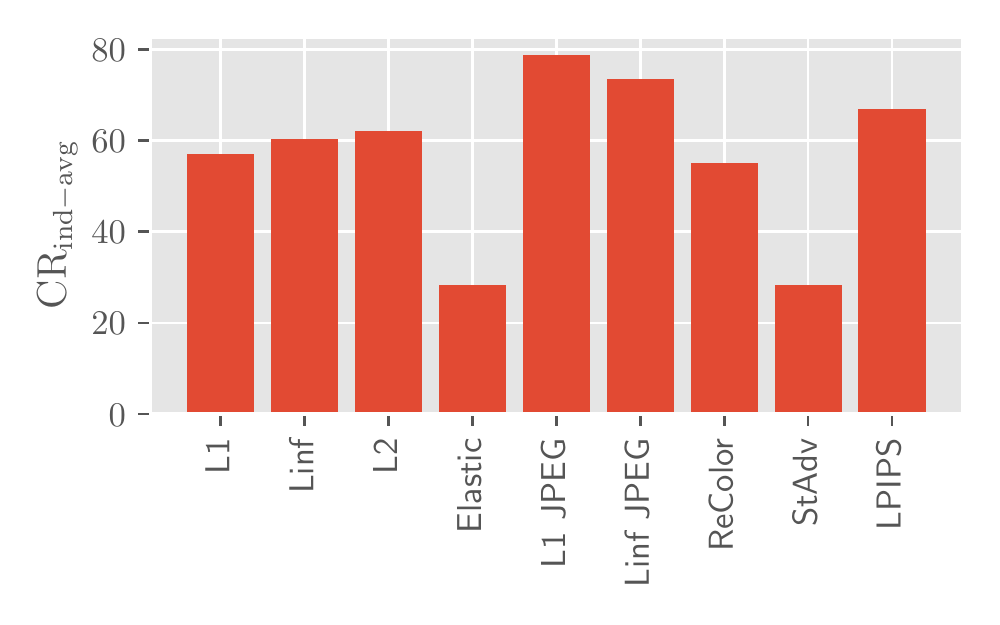}
    \caption{Average single-$\text{CR}_{\text{ind-avg}}$}
    \label{subfig:attack_diff}
    \end{subfigure}
    \caption{\noindent \textbf{State of current defenses.} Figures (a) and (b): Clean accuracy and CR of existing techniques on our leaderboard in terms. Each marker represents a single defense and the marker shape/color represents the types of attacks used by the defense during training ($K_{\text{learner}}$). Figure (c): $\text{CR}_{\text{ind-avg}}$ taken across each attack type averaged across all 16 defenses.}
    \label{fig:defense_performance}
\end{figure*}

For safety-critical applications, it is important that machine learning (ML) models are robust against test-time adversaries.  These test-time adversaries can potentially use multiple (and  unforeseen) attack types, motivating the need to study multiattack robustness.  Several works \citep{MainiWK20, TB19, Croce020, dai2022formulating, jin2020manifold, laidlaw2020perceptual,Hsiung2022towards} design defenses for multiattack robustness, but these works lack a unified evaluation framework: these works utilize different small sets of attacks and attack strengths.  The lack of a standardized benchmark for evaluating multiattack robustness is an obstacle to understanding and improving upon the current progress made by the community towards multiattack robustness.

To improve our understanding of current progress in multiattack robustness, we introduce \benchname~ (available at \url{multirobustbench.github.io}), which provides a leaderboard for multiattack robustness based on two new metrics that we introduce: \textit{competitiveness ratio} (CR) and \textit{stability constant} (SC).  CR measures how close the robust accuracy of a defense on each attack type is to the robust accuracy of the best performing model for each specific attack type.  SC measures  robustness degradation across attacks of different strengths.  Our benchmark evaluates 16 defended models based on a set of 9 different attacks across 20 levels of attack strengths (180 attacks total, 2880 evaluations overall), making it the largest multiattack evaluation to date. 
Our benchmark allows us to draw important insights on the state of research in multiattack robustness; specifically, we find that while existing research has made progress on average robustness over this set of attacks, all existing defenses perform worse than random guessing in worst-case multiattack robustness. 

Our contributions are as follows:

\textbf{We introduce an adversarial game framework for multiattack robustness.} This framework unifies previously studied settings such as robustness against unions of known attacks and robustness against unforeseen attacks by introducing \textit{knowledge sets} which capture mismatch in threat models used during training and test-time. Using this framework, we define a taxonomy of settings in multiattack robustness.

\textbf{We introduce metrics (competitiveness ratio and stability constant) for measuring multiattack performance.} Competitiveness ratio (CR) can be interpreted as an aggregated percentage representing how close the accuracy of the defense is to the accuracy of the best performing models, while stability constant (SC) measures how much robustness changes when switching to a different attack strengths.  We introduce 2 different variants of CR, one for measuring average multiattack robustness ($\text{CR}_{\text{ind-avg}}$) and one for measuring worst case multiattack robustness $\text{CR}_{\text{ind-worst}}$.

\textbf{Using our proposed metrics, we provide a leaderboard containing evaluations of existing defenses targeting muliattack robustness.} Our leaderboard evaluates existing models on a wide variety of attacks including bounded $\ell_p$ norm attacks, color changes \citep{laidlaw2020perceptual}, spatial transformations \citep{XiaoZ0HLS18}, elastic attacks, JPEG attacks \citep{kang2019robustness}, and bounded LPIPS attacks \citep{laidlaw2020perceptual}. Our leaderboard also provides features such as performance visualizations, which can be a useful diagnostic tool for understanding weaknesses of individual defenses.

\textbf{We analyze the state of current defenses for multiattack robustness.}  We find that while current models have decent performance in terms of average case robustness across multiple (imperceptible) attacks, there is significant room for improvement in terms of worst-case performance.  Additionally, using our metrics, we analyze how factors such as architecture size, use of additional training data, and number of training epochs influence multiattack robustness.

Overall, benchmarks have the potential to revolutionize ML by enabling comparable research and highlighting an open research problem for our community. We hope that our benchmark inspires research in multiattack robustness and accelerates the development of stronger defenses.

\section{Prior Work}
\textbf{Adversarial robustness.} Prior works have demonstrated a vulnerability of existing ML models: imperceptible perturbations during test-time can cause models to misclassify \citep{szegedy2013intriguing}.  These imperceptible perturbations can be of many different types including norm-bounded perturbations, small spatial transformations \citep{XiaoZ0HLS18}, color changes \citep{LaidlawF19}, and their compositions \citep{hsiung2022carben}.  Although a variety of attack types exist, the majority of current research in adversarial robustness focuses on defending models against perturbations that are bounded in $\ell_2$ or $\ell_\infty$ norm.  Adversarial training \citep{madry2017towards, zhang2019theoretically, gowal2020uncovering}, a popular defense framework in which the model is trained with adversarial examples, has been mainly studied for $\ell_p$ robustness.  For example, prior works have studied the impact of architecture size \citep{wu2021wider, huang2021exploring}, early stopping \citep{rice2020overfitting}, and additional training data \citep{carmon2019unlabeled, rebuffi2021fixing, gowal2021improving, sehwag2021robust} on robustness on the $\ell_{\infty}$ and $\ell_2$ attacks used during adversarial training. However, it is unclear how applicable these findings and defenses are in practice, where the adversary can potentially use multiple attacks which might not be known in advance to the defender. In our work, we find that general trends observed for robustness against single (known) attacks do not necessarily hold for multiattack robustness.

\textbf{Multiattack robustness.} Several prior works have studied robustness against multiple attacks.  One line of works focuses on improving robustness against the union of known attacks (typically the union of $\ell_p$-balls) \citep{MainiWK20, TB19, Croce020, madaan2020learning}.  Another line of works looks at defending against attacks that are not used during training \citep{laidlaw2020perceptual, dai2022formulating, jin2020manifold}.  We provide a framework that unifies both of these research directions and provides metrics and a leaderboard for benchmarking these defenses.

\textbf{Benchmarking adversarial robustness.} RobustBench \citep{croce2020robustbench} is a standardized benchmark for adversarial robustness which provides leaderboards for $\ell_{\infty}$ and $\ell_2$ robustness measured via AutoAttack \citep{croce2020reliable}.  CARBEN \citep{hsiung2022carben} is a benchmark for measuring robustness against compositions of $\ell_{\infty}$ attacks with global spatial transformations (i.e. rotation) and global color shifts (i.e. hue shift).  These global transformations are weaker than existing attacks like StAdv \citep{XiaoZ0HLS18} and ReColor \citep{LaidlawF19} which optimize over pixels, making it unclear how well evaluations from CARBEN reflect true multiattack robustness.

As a research community, we lack a standardized benchmark that accurately reflects robustness against multiple attacks.  Current works in multiattack robustness deploy different methods for evaluating the performance of their defense, such as reporting accuracies for different sets of attacks or using custom metrics (such as mUAR \citep{kang2019robustness}) for measuring performance.  To standardize evaluation for multiattack robustness, we introduce a leaderboard which ranks existing defenses for multiattack robustness based on metrics motivated by our proposed framework for multiattack robustness.  Our leaderboard also provides performance visualizations (Appendix \ref{app:performance_visualizations}) for each defense so that researchers can understand the weaknesses of existing defenses in detail, which can potentially lead to improvements in the direction of multiattack robustness.

\section{Robustness against multiple attacks}
We begin by first providing a framework for modeling problems in multiattack robustness.  We then discuss goals in multiattack robustness and metrics for measuring it.

\textbf{Notations} We use $\mathcal{D} = X \times Y$ to denote the data distribution where $X$ is the support of the images and $Y$ is the support of the labels. We use $D_{\text{train}}$ to denote the training set with data sampled from $\mathcal{D}$ in an i.i.d. manner.  We will refer to the defense as a learning algorithm which we will denote with $\mathcal{A}$. We use $\mathcal{H}$ to denote the hypothesis class used by $\mathcal{A}$ (ie. $\mathcal{A}$ outputs a function $h \in \mathcal{H}$).

\subsection{A unified adversarial game framework for modeling robustness against multiple attacks}
\label{sec:adv_game}
While prior works have studied problems relating to robustness against multiple attacks \citep{TB19, MainiWK20, Croce020, laidlaw2020perceptual, dai2022formulating, jin2020manifold,Hsiung2022towards}, these works have studied specific instances of multiattack robustness (i.e. robustness against unions of known threat models, unforeseen attack robustness), but have not provided a unified framework for modelling problems in multiattack robustness. In this section, we propose a unified framework for multiattack robustness by providing an adversarial game formulation.

We begin by introducing a perturbation function which maps inputs to adversarial examples.

\begin{definition}[Perturbation Function] Let $C:X \to 2^X$ define the constraint of the adversary and $\ell: Y \times Y \to \mathbb{R}$ 
be a loss function.  A perturbation function $P_C: X \times Y \times \mathcal{H} \to X$ maps input and hypothesis to adversarially perturbed versions of the input:
$$P_C(x, y, h) = \argmax_{x' \in C(x)} \ell(h(x), y)$$
\end{definition}



To capture multiattack settings such as robustness against unforeseen attacks where the learner does not know what type of attacks are present during test time, we introduce a knowledge set.
\begin{definition}[Knowledge Set] A knowledge set $K_{\text{learner}}$ is a set of perturbation functions.  We say that the defender is restricted to knowledge set $K_{\text{learner}}$ if the learning algorithm optimizes model selection by using information about perturbation functions only within $K_{\text{learner}}$. \end{definition}

The learner and attacker knowledge sets allow us to model robustness against multiple perturbations as an adversarial game:

\begin{definition}[Adversarial Game for Multiple Attacks] \label{def:adv_game}
\end{definition}
\vspace{-15pt}
\begin{enumerate}
\itemsep0em
    \item Environment specifies a robustness threshold $\gamma$ and specifies a (possibly infinite) set $K$ of perturbation functions that can occur during test-time.  The environment also specifies the learner's knowledge set $K_{\text{learner}}$ where $|K_{\text{learner}}| \le |K|$. 
    \item The learner then chooses learning algorithm $\mathcal{A}$ and obtains model $h = \mathcal{A}(D_{\text{train}}, K_{\text{learner}})$.  Here, $ \mathcal{A}(D_{\text{train}}, K_{\text{learner}})$ denotes that the learning algorithm is restricted to using information about perturbation functions within $K_{\text{learner}}$.
    \item If $\frac{\text{err}_{\text{multi}}(h; K)}{\min_{h^* \in \mathcal{H}} \text{err}_{\text{multi}}(h^*; K)}  \le \gamma$,
    then the learner wins and $\mathcal{A}$ produces a model that is close to optimal against $K$. Otherwise the attacker wins.
\end{enumerate}

The definition of $\text{err}_{\text{multi}}$ and relationship between $K$ and $K_{\text{learner}}$ can lead to different forms of robustness against multiple attacks.

\textbf{Relationship between $K$ and $K_{\text{learner}}$.} The relationship between $K$ and $K_{\text{learner}}$ leads to different settings for robustness against multiple attacks.  The setting where $K = K_{\text{learner}}$ models the commonly studied setting where the learner knows the attacks used during evaluation in advance and can optimize their model directly with respect to those attacks.  For example, works studying robustness against unions of $\ell_p$ attacks \citep{TB19, MainiWK20, Croce020} fall under this category.  We call the setting where $K = K_{\text{learner}}$ the \textit{full knowledge} setting.  We note that when $|K| = 1$, the adversarial game for multiple attacks reduces to the adversarial game for a single attack.

When $K \ne K_{\text{learner}}$, there is a mismatch between the attacks that the learner is aware about and can use for the learning algorithm.  We call this the \textit{knowledge mismatch} setting.  The types of mismatches can be divided into several cases: 1) $K \cap K_{\text{learner}} = \emptyset$, 2) $K \cap K_{\text{learner}} \ne \emptyset$.

The first case represents settings where the learner has no knowledge of the true space of attacks.  An example of this is if the adversary is constrained to using patch attacks, while the learner is under the impression that there will only be imperceptible attacks at test-time so $K_{\text{learner}}$ consists of a selection of imperceptible attacks (ie. bounded $\ell_p$ attacks).  We call this the setting of \textit{no knowledge}.

The second case represents settings where the learner knows only a subset of attacks that will be used during test-time.  We call this setting the \textit{partial knowledge} setting and contains the problem of unforeseen attacks.  An example of this is when the test-time adversary is restricted to attacks that do not change a human's classification of the image, but the learner does not know how to model the full space of these attacks and is aware of only a subset of those attacks (ie. bounded $\ell_p$ attacks).  For the task of image classification, the setting of learner knowledge models a more realistic learning setting compared to the full knowledge since we would like our model to be robust against attacks developed in the future.  The partial knowledge setting is also more realistic in comparison to the no knowledge setting for image classification since existing known attacks are also valid attacks that can be used the adversary during test-time.

In Appendix \ref{app:categorization}, we categorize existing defenses against multiple attacks into full, partial, and no knowledge settings.

\textbf{Definition of $\text{err}_{\text{multi}}$.} Choosing the definition of err for multiple attacks also leads to different problems in multiattack robustness. For example, if the learner knows the distribution $\mathcal{P}(K)$ of frequency at which attacker chooses each attack in $K$, then this can be modeled with $\text{err}_{\text{multi-exp}}(h;K) := \mathbb{E}_{P \sim \mathcal{P}(K)} \text{err}(h; P)$.  The learner can also consider using the worst case error across all $P \in K$ as a measure of multiattack performance: $\text{err}_{\text{multi-max}}(h;K) := \max_{P \in K} \text{err}(h; P)$.

Another possibility is to let $\gamma$ be a vector of length equal to the number of perturbation functions in $K$ and letting $\text{err}_{\text{multi}}$ output a vector of errors with respect to each individual perturbation $P \in K$ (ie. $\text{err}_{\text{multi-ind}} := [\text{err}(h;P)]_{P \in K}$).  In this case, the learner only wins the game if the losses on each individual attack lies within the corresponding robustness threshold in the vector $\gamma$.  $\text{err}_{\text{multi-ind}}$ allows us to model the problem of achieving robustness against the union of attacks in $K$ while also allowing us to specify how much tradeoff in performance across attacks we are willing to tolerate.

\subsection{Metrics for evaluating multiattack robustness} 
\label{sec:metrics}
Using the adversarial game formulation in Definition \ref{def:adv_game}, we now design metrics which aggregate accuracy across each individual attack into a single number.  In this section, we discuss two potential criteria that we would like to achieve when designing a good defense: 1) competitive performance and 2) stability across attack difficulty and introduce the metrics we use for measuring each criterion. 


\textbf{Competitive performance across attacks.} 
In the multiattack adversarial game formulation in Definition \ref{def:adv_game}, we saw that the objective of the learner is to choose a learning algorithm which allows the learner to obtain a model $h$ whose performance is competitive with the best model in the hypothesis set with respect to the choice of $\text{err}_{\text{multi}}$.  We introduce a family of metrics which we call competitiveness ratio (CR), which measures how close $h$ is to the best model in the hypothesis set.

\begin{definition}[Competitiveness Ratio (CR)]
Let $\text{acc}^*_{\text{multi}}(K) := 1 - \min_{h^* \in \mathcal{H}} \text{err}_{\text{multi}}(h^*; K)$ and $\text{acc}_{\text{multi}}(h, K):= 1 - \text{err}_{\text{multi}}(h; K)$.  Then, the competitiveness ratio (CR) of a defended model $h$ is given by:
\begin{equation}\text{CR}(h; K) = 100 \times \frac{\text{acc}_{\text{multi}}(h, K)}{\text{acc}^*_{\text{multi}}(K)}
\end{equation}
\end{definition}
 
In practice, we approximate $\text{acc}^*$ through adversarial training and will discuss this in more depth in Section \ref{sec:eval_setup}.  We note that CR can be used in all knowledge settings since metrics are taken with respect to $K$ which can differ from $K_{\text{learner}}$.  Using different definitions of $\text{err}_{\text{multi}}$ leads to different variants of CR.  



For example, if we use $\text{err}_{\text{multi-ind}}$ 
as the multi-attack error function, then CR compares each attack within $K$ to the best accuracy on that specific attack.  We can then aggregate all of these scores by either taking the expectation or worst case, leading to the following variants of CR:

\begin{definition}($\text{CR}_{\text{ind-avg}}$ and $\text{CR}_{\text{ind-worst}}$) \label{def:crind}
For a single $P \in K$, let $\text{acc}^*(P):= 1 - \min_{h \in \mathcal{H}} \text{err}(h; P)$ and $\text{acc}(h, P):= 1 - \text{err}(h; P)$. Then,
\begin{equation}
    \text{CR}_{\text{ind-avg}}(h; K) := 100 \times  \mathbb{E}_{P\sim \mathcal{P}(K)} \left[ \frac{\text{acc}(h, P)}{\text{acc}^*(P)} \right] \label{eq:crindavg}
  \end{equation}
  \begin{equation}
    \text{CR}_{\text{ind-worst}}(h; K) := 100 \times \min_{P \in K} \frac{\text{acc}(h, P)}{\text{acc}^*(P)}
  \end{equation}
\end{definition}
  
We discuss using other choices for $\text{err}_{\text{multi}}$ in Appendix \ref{app:multiattack_err}.

For choices of $\text{err}_{\text{multi}}$, high CR indicates that the model $h$ is closer to optimal with regards to our chosen definition of $\text{err}_\text{multi}$.  When $K$ contains only attacks of the same type (ie. $\ell_2$ perturbations) at different strengths (ie. radii of $\ell_2$ ball), we call this metric \textit{single-CR}.  When $K$ contains other attack types.  Comparing single-CR values for a set of attacks allow us to understand whether there are some specific attack types that the model performs poorly on.  

\textbf{Stability across attack strength.} 
As discussed in Section \ref{sec:adv_game}, there can exist a knowledge mismatch between $K$ and $K_{\text{learner}}$.  For example,  $K_{\text{learner}}$ can $\ell_2$ perturbations with radius up to $0.50$ while $K$ contains $\ell_2$ perturbations with radius up to $0.51$.  In this case, another goal of the learner is to have a graceful degradation of robustness in the vicinity of attacks in $K_{\text{learner}}$: since $0.51$ is close to $0.50$, we should not see a drastic difference in robustness from $K_{\text{learner}}$ to $K$.


We now define an attack strength function, which measures difficulty of attacks.  We will then use this definition to define \textit{stability}, which is our criterion for measuring smooth degradation of robustness across attacks of similar difficulty.

\begin{definition}[Attack strength function]  An \textit{attack strength function} $s:K \to \mathbb{R}^+$ maps perturbation functions in attack set $K$ to a number representing the difficulty of the attack. For example, if $K$ contains a single attack type at different perturbation sizes $\epsilon$, we can consider $s$ to output the value of $\epsilon$ corresponding to the attack. As another example, with multiple attack types, we can consider an attack strength function $s(P) = \min_{h \in \mathcal{H}} \text{err}(h;P)$. 
\end{definition}

Using the attack strength function definition, we now define stability across perturbations.

\begin{definition}[Stability across perturbations] \label{def:stability} A model $h$ is $(L, \alpha)$-locally stable across perturbations with respect to attack strength function $s$ if we have that for all $P_1 \in K_{\text{learner}}$ and $P_2 \in K$ such that $|s(P_1) - s(P_2)| \le \alpha$, $|\text{acc}(h, P_1) - \text{acc}(h, P_2)| \le L|s(P_1) - s(P_2)|$.  Equivalently, for a given $\alpha$ and model $h$, we can compute the corresponding constant $L$, which we call the \textit{stability constant (SC)} as follows:

\begin{equation}
    L_{\alpha}(h) = \max_{\substack{P_1 \in K_{\text{learner}}, P_2 \in K \\ |s(P_1) - s(P_2)| \le \alpha \\ P_1 \ne P_2}} \frac{|\text{acc}(h, P_1) - \text{acc}(h, P_2)|}{|s(P_1) - s(P_2)|}
\end{equation}

\end{definition}
In the above definitions of stability and SC, since $\alpha$ represents the difference in difficulty between attacks, we are interested in the regime of small $\alpha$. Ideally, we would like SC to be small at small values of $\alpha$, since that would suggest robust performance does not change much for attacks of similar difficulty.

\section{Description of \benchname}
\begin{figure*}[th]
     \centering
         \includegraphics[width=0.75\textwidth]{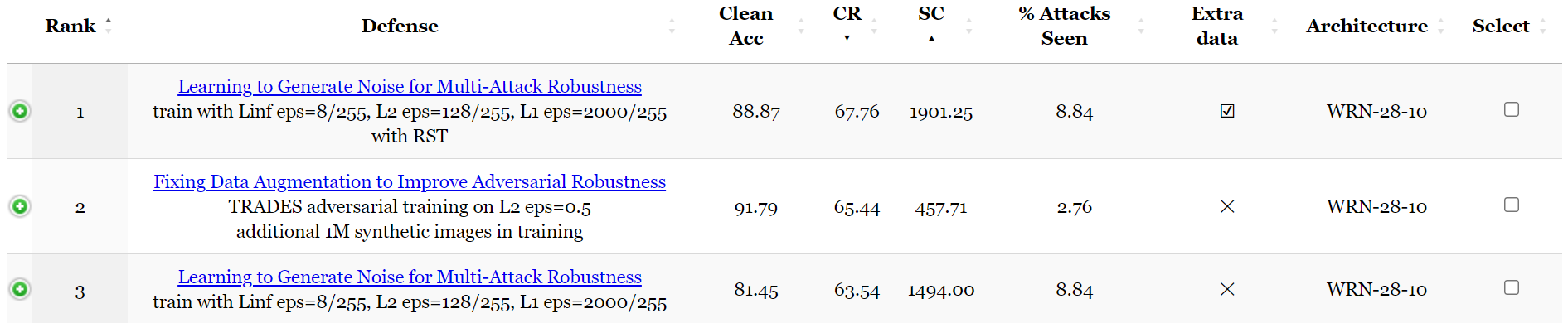}
         \caption{Top three entries on our leaderboard for \textit{average} case multiattack robustness on CIFAR-10}
         \label{fig:top3avg}
        \vspace{-10pt}
\end{figure*}

\begin{figure*}
    \centering
    \includegraphics[width=0.75\textwidth]{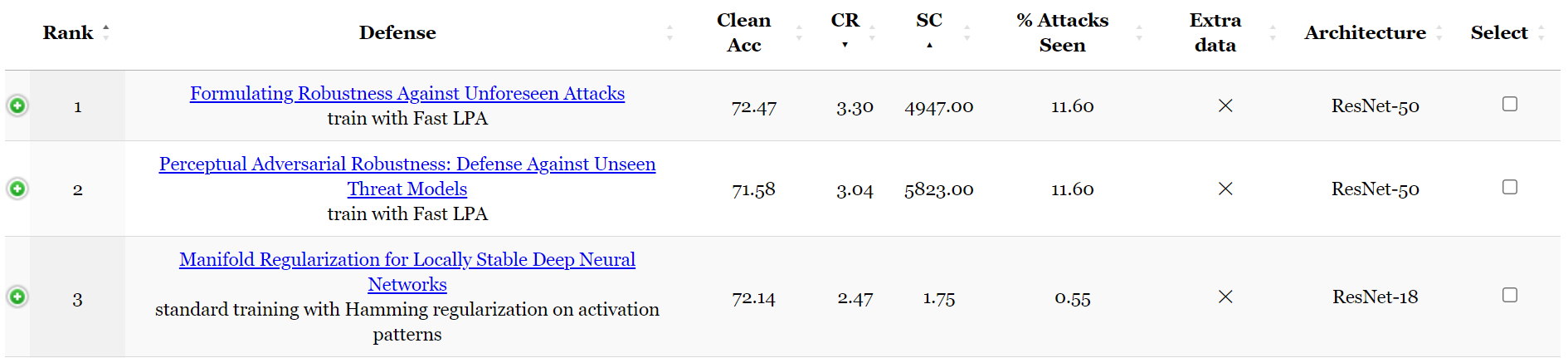}
         \caption{Top three entries on our leaderboard for \textit{worst} case multiattack robustness on CIFAR-10}
    \label{fig:top3worst}
\end{figure*}

Using CR and SC introduced in Section \ref{sec:metrics}, we provide a leaderboard that ranks existing defenses against multiple adversarial perturbations in order to standardize evaluation of defenses against multiple attacks. Our leaderboard also provides visualizations of performance across individual attack types for researchers to analyze and understand strengths and weaknesses of their defenses.  This leaderboard is available at \url{https://multirobustbench.github.io/}.
\subsection{Evaluation Setup}
\label{sec:eval_setup}

\textbf{Restrictions} Similar to RobustBench \citep{croce2020robustbench}, we focus on models that have a fully deterministic forward pass, nonzero gradients, and no optimization loop in the forward pass.  Given that the bulk of attacks used in benchmarking are white-box attacks, our evaluations may be inaccurate for any model which does not satisfy these requirements.

\textbf{Attack Space}  The space of attacks $K$ which do not visually change the class of the original image is infinite, so it is important to define a subset of these attacks to use for evaluation. For benchmarking, we consider 9 different attack types at 20 different attack strengths ($\epsilon$).  We provide detailed descriptions of each attack and range of $\epsilon$ used per attack in Appendix \ref{app:attack_desc}.
\vspace{-5pt}
\begin{itemize}
\itemsep0em
    \item \textbf{Bounded $\ell_p$ attacks} We consider $\ell_1$, $\ell_2$, and $\ell_{\infty}$ attacks.  To measure robustness, we use apgd-t and fab-t from the AutoAttack package \citep{croce2020reliable}.  We restrict to using this subset of attacks to reduce evaluation time (since we evaluate a total of 60 $\ell_p$ attacks per model).
    
    \item \textbf{Color shift} For leaderboard rankings, we consider pixel-wise color shifts via ReColor attacks \citep{LaidlawF19}.  
    
    \item \textbf{Spatial transformations} For leaderboard rankings, we consider small shifts in pixel positions using StAdv attacks \citep{XiaoZ0HLS18}. 
    
    \item \textbf{UAR attacks \citep{kang2019robustness}} \citet{kang2019robustness} introduced a set of attacks for measuring unforeseen robustness including elastic, Linf JPEG, and L1 JPEG attacks. We incorporate these 3 attacks into our benchmark. 

\item \textbf{Bounded LPIPS attacks \citep{laidlaw2020perceptual}} \citet{laidlaw2020perceptual} introduced 2 attacks (PPGD and LPA) based on LPIPS distance \citep{zhang2018lpips}, which is a more perceptually aligned distance metric than $\ell_p$ norms. For evaluation, we measure robustness against LPIPS attacks by taking the accuracy against the union of PPGD and LPA attacks.

\end{itemize}
\vspace{-5pt}

\textbf{Approximating optimal single attack accuracy.}  In Section \ref{sec:metrics}, we defined CR in terms of optimal single attack accuracy $\text{acc}^*(P)$.  In practice, we do not know these optimal values, so we approximate these by using the accuracies of ResNet-18 models that are trained using adversarial training directly on to attack of interest $P$. We choose to use ResNet-18 models for training efficiency and use the robust accuracy averaged over 3 runs for $\text{acc}^*(P)$.  We note that by using ResNet-18 accuracies for $\text{acc}^*(P)$, our metrics are also able to capture improvements in multiattack performance due to changes in architecture.

\textbf{Attack strength function.} In Section \ref{sec:metrics}, we defined SC in terms of an attack strength function.  
For our leaderboard, we choose to use the error of a ResNet-18 model trained directly on the attack as the attack strength function.  For computing SC as in Definition \ref{def:stability}, we use $\alpha=3\%$.


\subsection{Leaderboard}
We provide 2 leaderboards for the CIFAR-10 dataset, one for average case performance, which ranks defenses based on $\text{CR}_{\text{ind-avg}}$, and one for worst case performance, which ranks defenses based on $\text{CR}_{\text{ind-worst}}$. Our leaderboard contains evaluations for 16 pretrained models, all of which use training-based defenses, including techniques for training on unions of $\ell_p$ norms \citep{MainiWK20, TB19, madaan2020learning}, training with novel threat models \citep{laidlaw2020perceptual}, regularization based approaches \citep{jin2020manifold, dai2022formulating}, and $\ell_p$ norm adversarial training \citep{madry2017towards, zhang2019theoretically, rebuffi2021fixing}.   We include details of the models present on the leaderboard in Appendix \ref{app:defenses}.  We note that these models are trained with either $\ell_2$ attacks with $\epsilon=0.5$, $\ell_{\infty}$ attacks with $\epsilon = \frac{8}{255}$, LPIPS attacks with $\epsilon=1$, or the union of $\ell_1, \epsilon=\frac{2000}{255}$, $\ell_2, \epsilon=\frac{128}{255}$ and $\ell_{\infty}, \epsilon=\frac{8}{255}$ attacks.

We compute ranks based on the set of attacks described in \ref{sec:eval_setup}. We note that none of the models evaluated use all the attacks in our evaluation set, so \textit{all models are evaluated for performance in a partial knowledge or no knowledge setting}.  The top 3 entries on each leaderboard are shown in Figure \ref{fig:top3avg} and Figure \ref{fig:top3worst}. Our leaderboard site also provides features such as performance visualizations.  We discuss these further in Appendix \ref{app:performance_visualizations}.


\section{Analysis}
\label{sec:analysis}

\begin{figure*}[ht]
    \centering
     
    \begin{subfigure}[t]{0.32\textwidth}
    \includegraphics[width=\textwidth]{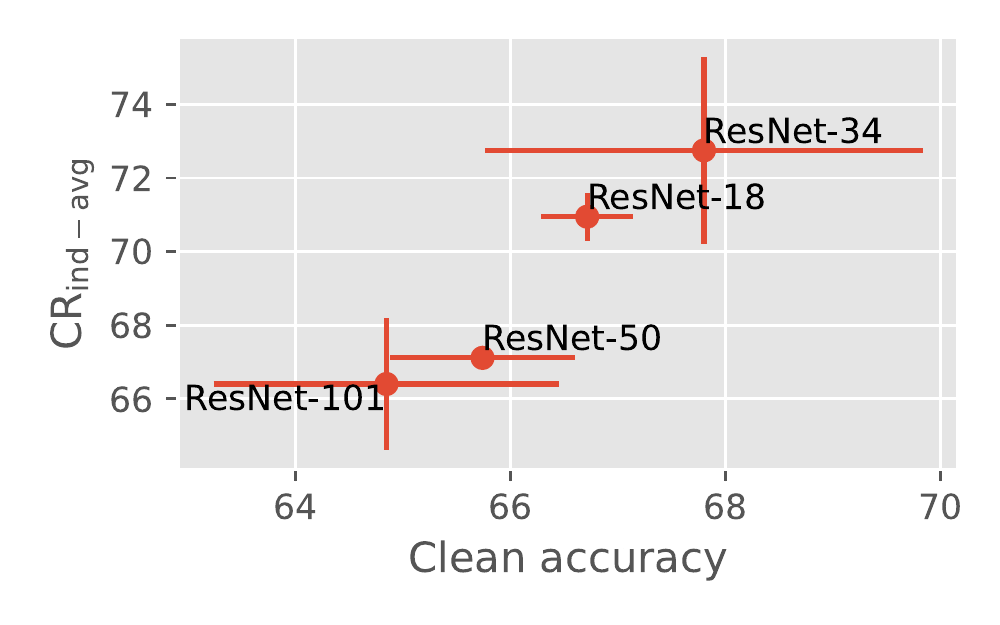}
    \caption{Clean accuracy vs $\text{CR}_{\text{ind-avg}}$}
    \label{subfig:lpips_arch_avg}
    \end{subfigure}
    \begin{subfigure}[t]{0.32\textwidth}
    \includegraphics[width=\textwidth]{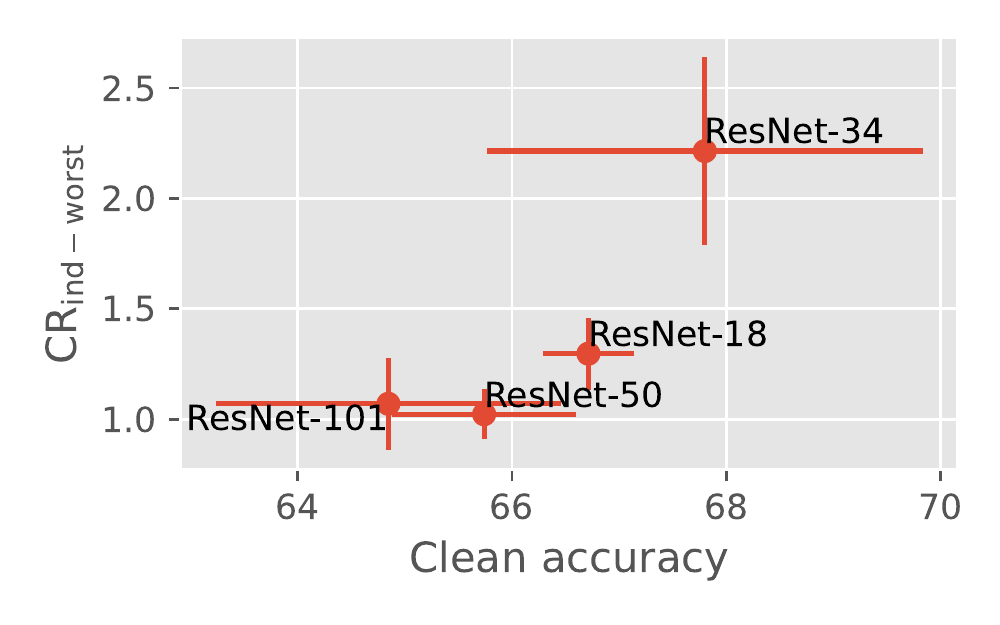}
    \caption{Clean accuracy vs $\text{CR}_{\text{ind-worst}}$}
        \label{subfig:lpips_arch_worst}
    \end{subfigure}
    \begin{subfigure}[t]{0.32\textwidth}
    \includegraphics[width=\textwidth]{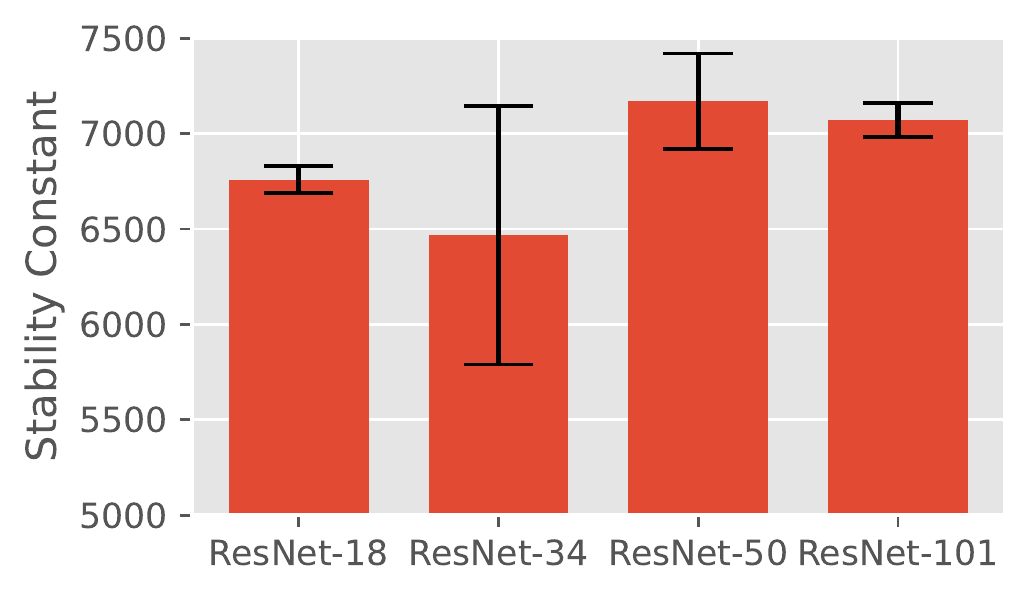}
    \caption{SC}
    \label{subfig:lpips_arch_stab}
    \end{subfigure}
    \caption{\noindent \textbf{Impact of architecture size.} Figures (a) and (b): Clean accuracy vs CR for models trained with PAT \citep{laidlaw2020perceptual} (LPIPS threat model).  Results are averaged over 3 trials and error bars are shown.  Higher values of CR indicate better performance. Figure (c): SC computed for models of each architecture. Lower SC indicates better performance.
    }
    \label{fig:lpips_arch_impact}
\end{figure*}

\begin{figure*}[ht]
    \centering
     
    \begin{subfigure}[t]{0.32\textwidth}
    \includegraphics[width=\textwidth]{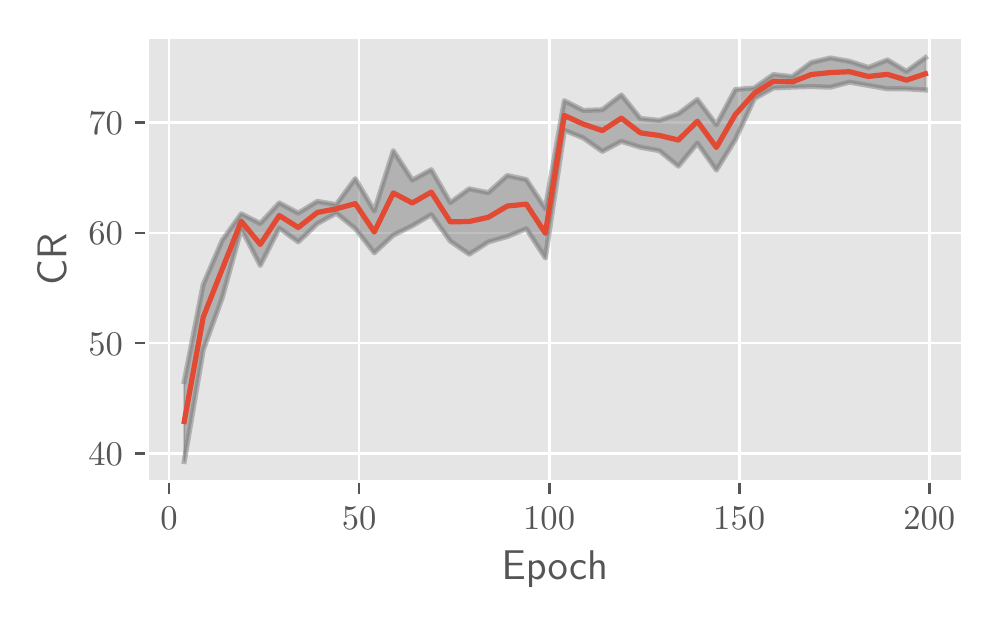}
    \caption{$\text{CR}_{\text{ind-avg}}$}
    \label{subfig:lpips_es_avg}
    \end{subfigure}
    \begin{subfigure}[t]{0.32\textwidth}
    \includegraphics[width=\textwidth]{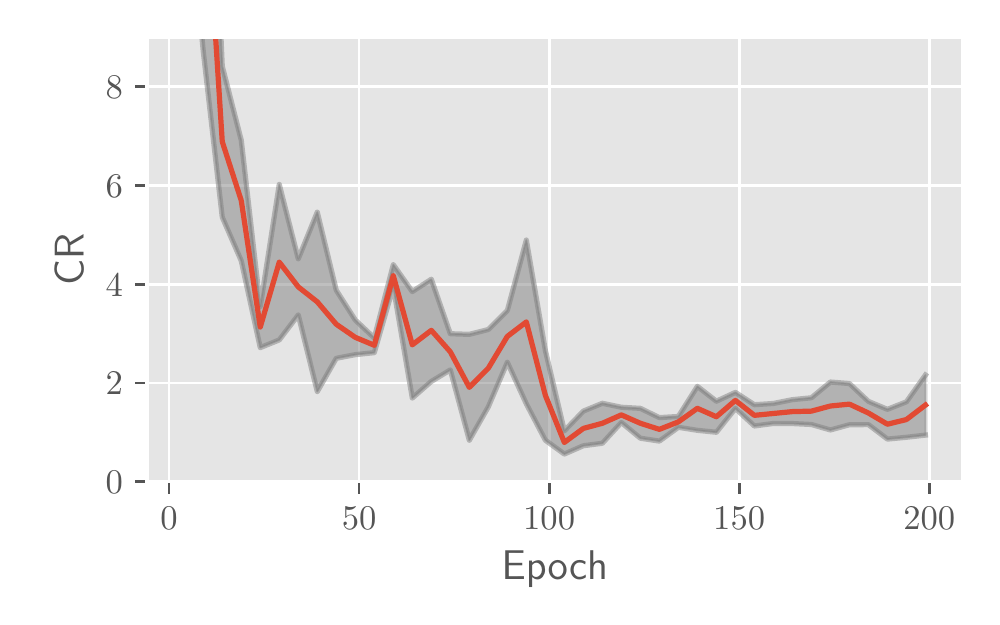}
    \caption{$\text{CR}_{\text{ind-worst}}$}
        \label{subfig:lpips_es_worst}
    \end{subfigure}
    \begin{subfigure}[t]{0.32\textwidth}
    \includegraphics[width=\textwidth]{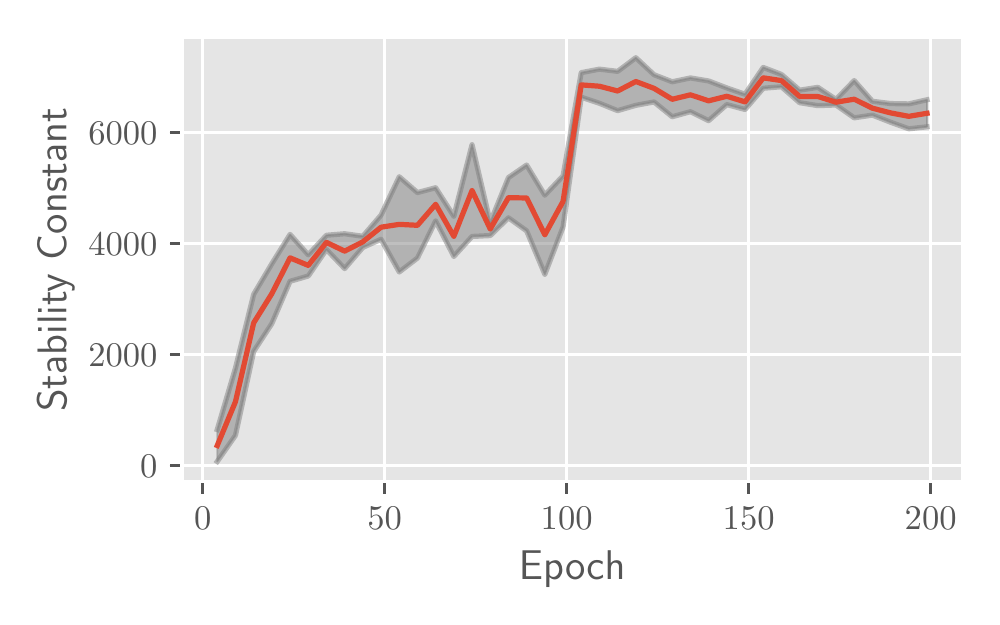}
    \caption{SC}
    \label{subfig:lpips_es_stab}
    \end{subfigure}
    \caption{\noindent \textbf{Impact of number of training epochs.} CR and SC over epoch for models trained using PAT \citep{laidlaw2020perceptual} (LPIPS threat model). The red line indicates the average over 3 runs while the grey band highlights indicate 1 standard deviation from the mean. Higher values of CR and lower values of SC indicate better performance.}
    \label{fig:lpips_early_stop_impact}
\end{figure*}

\begin{figure}[ht]
    \centering
     
    \begin{subfigure}[t]{0.4\textwidth}
    \includegraphics[width=\textwidth]{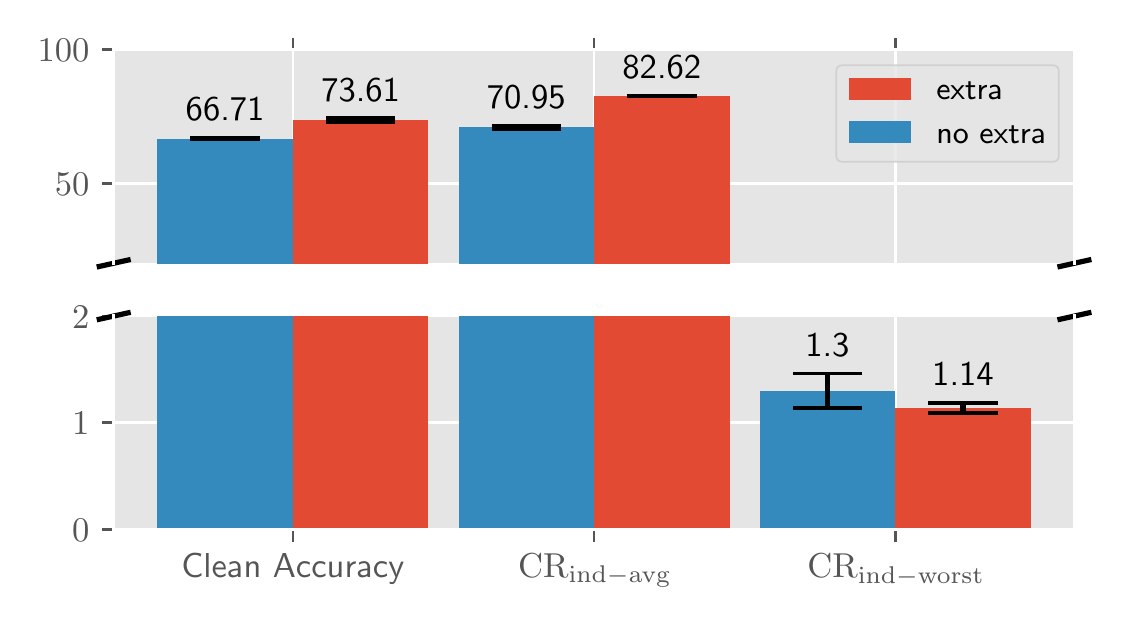}
    \caption{Clean Accuracy and CR}
    \label{subfig:lpips_extra_avg}
    \label{subfig:lpips_extra_worst}
    \end{subfigure}
    \begin{subfigure}[t]{0.34\textwidth}
    \includegraphics[width=\textwidth]{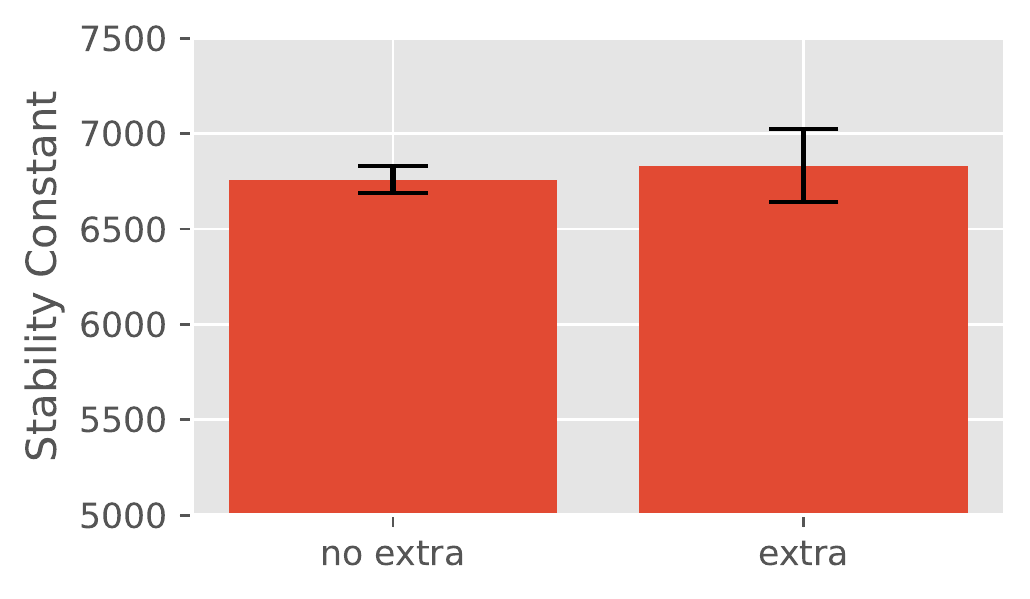}
    \caption{SC}
    \label{subfig:lpips_extra_stab}
    \end{subfigure}
    \caption{\noindent \textbf{Impact of additional training data.} Figure (a): Clean accuracy and CR for ResNet-18 models trained using PAT \citep{laidlaw2020perceptual} (LPIPS threat model).  Higher CR indicates better performance.  Results are averaged over 3 trials and error bars are shown. Figure (c): SC computed for models with and without additional training data. Lower SC indicates better performance.}
    \label{fig:lpips_extra_impact}
    \vspace{-20pt}
\end{figure}

Using our proposed metrics and leaderboard evaluations, we now analyze the performance of existing techniques for multiattack robustness (specifically under a partial or no knowledge setting).  Additionally, since some entries on our leaderboard utilize larger architecture size and additional training data, we separately study the impact of these factors on CR and stability to provide deeper insights as to how these design choices influence multiattack robustness.

\subsection{Evaluating existing techniques for robustness against multiple perturbations}
To understand the performance of existing defenses for multiattack robustness, we plot clean accuracy, $\text{CR}_{\text{ind-avg}}$, and $\text{CR}_{\text{ind-worst}}$ across defenses in Figure \ref{fig:defense_performance}. 

\textbf{Average case vs worst case multiattack performance.} Interestingly, we find that while many defenses can reach high values of $\text{CR}_{\text{ind-avg}}$ (the highest being 67.76), the scores for $\text{CR}_{\text{ind-worst}}$ are much lower (the highest being 3.30).  This suggests that for all existing defenses, there are some attacks (which may lie outside of the learner knowledge set) that can significantly reduce the accuracy of the defended model.  \textit{Thus, for the task of robustness against the worst-case imperceptible attack, designing defenses that a robust to multiple attacks is a significant open problem for the research community.}  This trade-off between clean accuracy and robust accuracy has been noted in prior works \citep{tsipras2019robustness, zhang2019theoretically, TB19}.

\textbf{Clean accuracy vs average case multiattack performance.}  From Figure \ref{subfig:clean_avg}, we find that some existing defenses achieve both high clean accuracy and high $\text{CR}_{\text{ind-avg}}$.  Interestingly, we find that the 2 best models in terms of $\text{CR}_{\text{ind-avg}}$ and clean accuracy (\citep{madaan2020learning} with robust self-training and \citep{rebuffi2021fixing} with $\ell_2$ threat model) both incorporate additional training data which suggests that additional training data may improve average performance over attacks tested.  Overall, we find that $\text{CR}_{\text{ind-avg}}$ are uncorrelated; for example, the rank 4, 5, and 6 models in terms of $\text{CR}_{\text{ind-avg}}$ (models using LPIPS threat model \citep{laidlaw2020perceptual, dai2022formulating} and no knowledge \citep{jin2020manifold}) have the lowest clean accuracies out of all defenses present on the leaderboard.

\textbf{Clean accuracy vs worst-case multiattack performance.} From Figure \ref{subfig:clean_worst}, we observe that the models with highest $\text{CR}_{\text{ind-worst}}$ also have the lowest clean accuracy, which differs from trends observed for $\text{CR}_{\text{ind-avg}}$.  We note that the models achieving the top 3 $\text{CR}_{\text{ind-worst}}$ scores are in fact the rank 4, 5, and 6 models in terms of $\text{CR}_{\text{ind-avg}}$. The state of current defenses in $\text{CR}_{\text{ind-worst}}$ also suggests that \textit{there may be some trade-off between worst-case multiattack performance and clean accuracy}.

\textbf{CR across individual attacks.} In Figure \ref{subfig:attack_diff}, we plot single-$\text{CR}_{\text{ind-avg}}$ (CR computed across individual attack types) averaged over all 16 defenses.  We find that out of all attack types, attacks that spatially perturb pixels (elastic attacks and StAdv attacks) are generally the most challenging to defend against.  In fact, the best performing model on elastic attacks can only achieve single-$\text{CR}_{\text{ind-avg}}$ score of 38.48 for elastic attacks.  Meanwhile, for StAdv attacks, the highest single-$\text{CR}_{\text{ind-avg}}$ score is 50.35.  We note that these scores are not obtained by the top 3 models ranked by $\text{CR}_{\text{ind-avg}}$, and are obtained by rank 7 (\citep{rebuffi2021fixing} with $\ell_{\infty}$ threat model) and rank 6 (\citep{laidlaw2020perceptual}) respectively.  This suggests that \textit{designing defenses that have improved performance on elastic and StAdv attacks can improve the state of current defenses for multiattack robustness}.

\subsection{Understanding the impact of architecture size, additional training data, and early stopping on multiattack robustness}
\label{sec:design_choice_impact}
While all evaluated models on our leaderboard use training-based defenses which can be applied to any architecture and training dataset, the entries differ in choice of architecture, use of additional training data, and number of training epochs used.  To investigate the impact of how these factors influence multiattack robustness, we train ResNet models using adversarial training with 3 different threat models (LPIPS with radius 0.5 \citep{laidlaw2020perceptual}, $\ell_{\infty}$ with radius $\frac{8}{255}$ and $\ell_2$ with radius 0.5) and analyze $\text{CR}_{\text{ind-avg}}$, $\text{CR}_{\text{ind-worst}}$, and SC of trained models.  We present results for LPIPS threat model in this section and provide corresponding analysis for $\ell_{\infty}$ and $\ell_2$ threat models in Appendix \ref{app:adv_train_analysis}.  Details about experimental setup are available in Appendix \ref{app:setup}.  We note that to reduce computational cost, we evaluate $\ell_{\infty}$ and $\ell_2$ robustness using PGD and $\ell_1$ robustness using APGD-CE \citep{croce2020reliable} instead of running APGD-T and FAB-T attacks as done for leaderboard evaluations, so CR scores present in this section are not directly comparable to those on the leaderboard.

\textbf{Impact of architecture size.} We present results on the impact of architecture size on CR and SC in Figure \ref{fig:lpips_arch_impact}.  We find that out of the architectures tested (ResNet-18, ResNet-34, ResNet-50, ResNet-101), smaller architectures (ResNet-18 and ResNet-34) generally have higher clean accuracy and higher CR compared to ResNet-50 and ResNet-101.  While previous studies \citep{gowal2020uncovering} demonstrated that larger architecture can improve robust performance for $\ell_p$ robustness, we find that this is not always the case for multiattack robustness (even with $\ell_p$ training as shown in Appendix \ref{app:adv_train_analysis}).  The higher CR values suggest that these \textit{smaller models have better generalization to unseen attacks while larger models are more likely to overfit to seen attack types}. We find that SC is also on average lower for smaller architectures which indicates that \textit{smaller architectures have less change in robust accuracy across attack types}.

\textbf{Impact of number of epochs.} We present results on the impact of number of training epochs on CR and SC in Figure \ref{fig:lpips_early_stop_impact}.  We observe that while $\text{CR}_{\text{ind-avg}}$ generally increases over training epochs, $\text{CR}_{\text{ind-worst}}$ decreases over epochs, indicating that average robustness increases, but worst-case robustness does not.  \textit{This suggests that while more training improves average performance across the set of tested attacks, there may be a few attacks in this set for which more training degrades  performance.}  When we investigate this further, we find that for all attacks except for elastic attacks, $\text{CR}_{\text{ind-worst}}$ increases over epochs.  In Appendix \ref{app:lpips_per_attack_epochs}, we plot the $\text{CR}_{\text{ind-worst}}$ for each attack type and investigate the impact of including elastic attacks in training.  The trend in worst-case robustness is also reflected by Figure \ref{subfig:lpips_es_stab} which shows that SC increases over training epochs, meaning there is a large drop in robustness when evaluated on unseen attacks. 

\textbf{Impact of additional training data.}  We now investigate the impact of using additional (synthetic) training data.  Specifically, we incorporate the 1M DDPM samples from \citep{gowal2021improving} for CIFAR-10 into training.  We present results on the impact of additional data on CR and SC in Figure \ref{fig:lpips_extra_impact}.  From Figure \ref{subfig:lpips_extra_avg}, we find that for $\text{CR}_{\text{ind-avg}}$, using additional data significantly improves clean accuracy and CR scores, suggesting that the \textit{extra training data can improve average robustness across attacks}.  In fact, we achieve SOTA $\text{CR}_{\text{ind-avg}}$ (69.14) when compared to other models on the leaderboard.  For worst-case performance ($\text{CR}_{\text{ind-worst}}$), we find that CR with and without extra data is comparable.  This suggests that while on average extra data helps, extra data does not uniformly improve performance across all attacks.  In Appendix \ref{app:lpips_per_attack_data}, we plot the impact of additional data on $\text{CR}_{\text{ind-worst}}$ for each attack type.  Similar to the overall trend for $\text{CR}_{\text{ind-worst}}$, we find that extra training data does not have much impact on stability; Figure \ref{subfig:lpips_extra_stab} shows that the SCs are comparable with and without extra data.

\section{Limitations, Discussion, and Conclusion}
The need for a benchmark is imperative to better understand and standardize the progress in multiattack robustness. In our benchmark, we introduce new metrics (CR and SC) and a leaderboard which ranks models based on a set of 180 attacks using these metrics.  Currently, our leaderboard contains 16 models, and as new defenses for multiattack robustness are proposed, we plan to update it with new defenses.  Additionally, as new attack types and stronger attacks are introduced, we plan to incorporate these into our evaluation pipeline.

One challenge with our benchmark is the runtime of evaluation, which makes it very computationally expensive to evaluate large architectures and large-scale image datasets, such as ImageNet.  Future improvements in attack efficiency can improve the scalability of our evaluation pipeline.  Currently, our leaderboard only contains evaluations for CIFAR-10; the future, we hope to include additional leaderboards for other image datasets.

Our benchmark and analyses highlight the weaknesses of current defenses on the task of worst-case multiattack robustness; in particular, we find that no defense can outperform random guessing.  In addition, we demonstrate that trends for single (known) attack robustness do not necessarily hold for the multiattack robustness.  We hope that our benchmark inspires future research in multiattack robustness.

\section*{Acknowledgements}
This work was supported in part by the National Science Foundation under grants CNS-2131938, the ARL’s Army Artificial Intelligence Innovation Institute (A2I2), Schmidt DataX award, and Princeton E-ffiliates Award.  This material is also based upon work supported by the National Science Foundation Graduate Research Fellowship under Grant No. DGE-2039656.  Any opinions, findings, and conclusions or recommendations expressed in this material are those of the author(s) and do not necessarily reflect the views of the National Science Foundation.




\bibliography{ref}

\begin{thebibliography}{30}
\providecommand{\natexlab}[1]{#1}
\providecommand{\url}[1]{\texttt{#1}}
\expandafter\ifx\csname urlstyle\endcsname\relax
  \providecommand{\doi}[1]{doi: #1}\else
  \providecommand{\doi}{doi: \begingroup \urlstyle{rm}\Url}\fi

\bibitem[Carlini \& Wagner(2017)Carlini and Wagner]{carlini2017towards}
Carlini, N. and Wagner, D.
\newblock Towards evaluating the robustness of neural networks.
\newblock In \emph{2017 ieee symposium on security and privacy (sp)}, pp.\
  39--57. IEEE, 2017.

\bibitem[Carmon et~al.(2019)Carmon, Raghunathan, Schmidt, Liang, and
  Duchi]{carmon2019unlabeled}
Carmon, Y., Raghunathan, A., Schmidt, L., Liang, P., and Duchi, J.~C.
\newblock Unlabeled data improves adversarial robustness.
\newblock \emph{arXiv preprint arXiv:1905.13736}, 2019.

\bibitem[Croce \& Hein(2020{\natexlab{a}})Croce and Hein]{Croce020}
Croce, F. and Hein, M.
\newblock Provable robustness against all adversarial
  {\textdollar}l{\_}p{\textdollar}-perturbations for
  {\textdollar}p{\textbackslash}geq 1{\textdollar}.
\newblock In \emph{8th International Conference on Learning Representations,
  {ICLR} 2020, Addis Ababa, Ethiopia, April 26-30, 2020}. OpenReview.net,
  2020{\natexlab{a}}.
\newblock URL \url{https://openreview.net/forum?id=rklk\_ySYPB}.

\bibitem[Croce \& Hein(2020{\natexlab{b}})Croce and Hein]{croce2020reliable}
Croce, F. and Hein, M.
\newblock Reliable evaluation of adversarial robustness with an ensemble of
  diverse parameter-free attacks.
\newblock In \emph{International conference on machine learning}, pp.\
  2206--2216. PMLR, 2020{\natexlab{b}}.

\bibitem[Croce \& Hein(2021)Croce and Hein]{pmlr-v139-croce21a}
Croce, F. and Hein, M.
\newblock Mind the box: $l_1$-apgd for sparse adversarial attacks on image
  classifiers.
\newblock In Meila, M. and Zhang, T. (eds.), \emph{Proceedings of the 38th
  International Conference on Machine Learning}, volume 139 of
  \emph{Proceedings of Machine Learning Research}, pp.\  2201--2211. PMLR,
  18--24 Jul 2021.
\newblock URL \url{https://proceedings.mlr.press/v139/croce21a.html}.

\bibitem[Croce et~al.(2020)Croce, Andriushchenko, Sehwag, Debenedetti,
  Flammarion, Chiang, Mittal, and Hein]{croce2020robustbench}
Croce, F., Andriushchenko, M., Sehwag, V., Debenedetti, E., Flammarion, N.,
  Chiang, M., Mittal, P., and Hein, M.
\newblock Robustbench: a standardized adversarial robustness benchmark.
\newblock \emph{arXiv preprint arXiv:2010.09670}, 2020.

\bibitem[Dai et~al.(2022)Dai, Mahloujifar, and Mittal]{dai2022formulating}
Dai, S., Mahloujifar, S., and Mittal, P.
\newblock Formulating robustness against unforeseen attacks.
\newblock \emph{arXiv preprint arXiv:2204.13779}, 2022.

\bibitem[Gowal et~al.(2020)Gowal, Qin, Uesato, Mann, and
  Kohli]{gowal2020uncovering}
Gowal, S., Qin, C., Uesato, J., Mann, T., and Kohli, P.
\newblock Uncovering the limits of adversarial training against norm-bounded
  adversarial examples.
\newblock \emph{arXiv preprint arXiv:2010.03593}, 2020.

\bibitem[Gowal et~al.(2021)Gowal, Rebuffi, Wiles, Stimberg, Calian, and
  Mann]{gowal2021improving}
Gowal, S., Rebuffi, S.-A., Wiles, O., Stimberg, F., Calian, D.~A., and Mann,
  T.~A.
\newblock Improving robustness using generated data.
\newblock \emph{Advances in Neural Information Processing Systems},
  34:\penalty0 4218--4233, 2021.

\bibitem[Hsiung et~al.(2022{\natexlab{a}})Hsiung, Tsai, Chen, and
  Ho]{Hsiung2022towards}
Hsiung, L., Tsai, Y.-Y., Chen, P.-Y., and Ho, T.-Y.
\newblock Towards compositional adversarial robustness: Generalizing
  adversarial training to composite semantic perturbations.
\newblock \emph{arXiv preprint arXiv:2202.04235}, 2022{\natexlab{a}}.

\bibitem[Hsiung et~al.(2022{\natexlab{b}})Hsiung, Tsai, Chen, and
  Ho]{hsiung2022carben}
Hsiung, L., Tsai, Y.-Y., Chen, P.-Y., and Ho, T.-Y.
\newblock {CARBEN: Composite Adversarial Robustness Benchmark}.
\newblock In \emph{Proceedings of the Thirty-First International Joint
  Conference on Artificial Intelligence, {IJCAI-22}}. International Joint
  Conferences on Artificial Intelligence Organization, July 2022{\natexlab{b}}.

\bibitem[Huang et~al.(2021)Huang, Wang, Erfani, Gu, Bailey, and
  Ma]{huang2021exploring}
Huang, H., Wang, Y., Erfani, S., Gu, Q., Bailey, J., and Ma, X.
\newblock Exploring architectural ingredients of adversarially robust deep
  neural networks.
\newblock \emph{Advances in Neural Information Processing Systems},
  34:\penalty0 5545--5559, 2021.

\bibitem[Jin \& Rinard(2020)Jin and Rinard]{jin2020manifold}
Jin, C. and Rinard, M.
\newblock Manifold regularization for locally stable deep neural networks.
\newblock \emph{arXiv preprint arXiv:2003.04286}, 2020.

\bibitem[Kang et~al.(2019)Kang, Sun, Hendrycks, Brown, and
  Steinhardt]{kang2019robustness}
Kang, D., Sun, Y., Hendrycks, D., Brown, T., and Steinhardt, J.
\newblock Testing robustness against unforeseen adversaries.
\newblock \emph{arXiv preprint arXiv:1908.08016}, 2019.

\bibitem[Laidlaw \& Feizi(2019)Laidlaw and Feizi]{LaidlawF19}
Laidlaw, C. and Feizi, S.
\newblock Functional adversarial attacks.
\newblock \emph{Advances in neural information processing systems}, 32, 2019.

\bibitem[Laidlaw et~al.(2021)Laidlaw, Singla, and Feizi]{laidlaw2020perceptual}
Laidlaw, C., Singla, S., and Feizi, S.
\newblock Perceptual adversarial robustness: Defense against unseen threat
  models.
\newblock In \emph{9th International Conference on Learning Representations,
  {ICLR} 2021, Virtual Event, Austria, May 3-7, 2021}. OpenReview.net, 2021.
\newblock URL \url{https://openreview.net/forum?id=dFwBosAcJkN}.

\bibitem[Madaan et~al.(2020)Madaan, Shin, and Hwang]{madaan2020learning}
Madaan, D., Shin, J., and Hwang, S.~J.
\newblock Learning to generate noise for robustness against multiple
  perturbations.
\newblock \emph{CoRR}, abs/2006.12135, 2020.
\newblock URL \url{https://arxiv.org/abs/2006.12135}.

\bibitem[Madry et~al.(2018)Madry, Makelov, Schmidt, Tsipras, and
  Vladu]{madry2017towards}
Madry, A., Makelov, A., Schmidt, L., Tsipras, D., and Vladu, A.
\newblock Towards deep learning models resistant to adversarial attacks.
\newblock In \emph{6th International Conference on Learning Representations,
  {ICLR} 2018, Vancouver, BC, Canada, April 30 - May 3, 2018, Conference Track
  Proceedings}. OpenReview.net, 2018.
\newblock URL \url{https://openreview.net/forum?id=rJzIBfZAb}.

\bibitem[Maini et~al.(2020)Maini, Wong, and Kolter]{MainiWK20}
Maini, P., Wong, E., and Kolter, J.~Z.
\newblock Adversarial robustness against the union of multiple perturbation
  models.
\newblock In \emph{Proceedings of the 37th International Conference on Machine
  Learning, {ICML} 2020, 13-18 July 2020, Virtual Event}, volume 119 of
  \emph{Proceedings of Machine Learning Research}, pp.\  6640--6650. {PMLR},
  2020.
\newblock URL \url{http://proceedings.mlr.press/v119/maini20a.html}.

\bibitem[Rebuffi et~al.(2021)Rebuffi, Gowal, Calian, Stimberg, Wiles, and
  Mann]{rebuffi2021fixing}
Rebuffi, S.-A., Gowal, S., Calian, D.~A., Stimberg, F., Wiles, O., and Mann, T.
\newblock Fixing data augmentation to improve adversarial robustness.
\newblock \emph{arXiv preprint arXiv:2103.01946}, 2021.

\bibitem[Rice et~al.(2020)Rice, Wong, and Kolter]{rice2020overfitting}
Rice, L., Wong, E., and Kolter, Z.
\newblock Overfitting in adversarially robust deep learning.
\newblock In \emph{International Conference on Machine Learning}, pp.\
  8093--8104. PMLR, 2020.

\bibitem[Sehwag et~al.(2021)Sehwag, Mahloujifar, Handina, Dai, Xiang, Chiang,
  and Mittal]{sehwag2021robust}
Sehwag, V., Mahloujifar, S., Handina, T., Dai, S., Xiang, C., Chiang, M., and
  Mittal, P.
\newblock Robust learning meets generative models: Can proxy distributions
  improve adversarial robustness?
\newblock In \emph{International Conference on Learning Representations}, 2021.

\bibitem[Szegedy et~al.(2014)Szegedy, Zaremba, Sutskever, Bruna, Erhan,
  Goodfellow, and Fergus]{szegedy2013intriguing}
Szegedy, C., Zaremba, W., Sutskever, I., Bruna, J., Erhan, D., Goodfellow,
  I.~J., and Fergus, R.
\newblock Intriguing properties of neural networks.
\newblock In Bengio, Y. and LeCun, Y. (eds.), \emph{2nd International
  Conference on Learning Representations, {ICLR} 2014, Banff, AB, Canada, April
  14-16, 2014, Conference Track Proceedings}, 2014.
\newblock URL \url{http://arxiv.org/abs/1312.6199}.

\bibitem[Tram{\`e}r \& Boneh(2019)Tram{\`e}r and Boneh]{TB19}
Tram{\`e}r, F. and Boneh, D.
\newblock Adversarial training and robustness for multiple perturbations.
\newblock In \emph{Conference on Neural Information Processing Systems
  (NeurIPS)}, 2019.
\newblock URL \url{https://arxiv.org/abs/1904.13000}.

\bibitem[Tsai et~al.(2022)Tsai, Hsiung, Chen, and Ho]{tsai2022compositional}
Tsai, Y.-Y., Hsiung, L., Chen, P.-Y., and Ho, T.-Y.
\newblock Towards compositional adversarial robustness: Generalizing
  adversarial training to composite semantic perturbations, 2022.

\bibitem[Tsipras et~al.(2019)Tsipras, Santurkar, Engstrom, Turner, and
  Madry]{tsipras2019robustness}
Tsipras, D., Santurkar, S., Engstrom, L., Turner, A., and Madry, A.
\newblock Robustness may be at odds with accuracy.
\newblock In \emph{International Conference on Learning Representations},
  number 2019, 2019.

\bibitem[Wu et~al.(2021)Wu, Chen, Cai, He, and Gu]{wu2021wider}
Wu, B., Chen, J., Cai, D., He, X., and Gu, Q.
\newblock Do wider neural networks really help adversarial robustness?
\newblock \emph{Advances in Neural Information Processing Systems},
  34:\penalty0 7054--7067, 2021.

\bibitem[Xiao et~al.(2018)Xiao, Zhu, Li, He, Liu, and Song]{XiaoZ0HLS18}
Xiao, C., Zhu, J., Li, B., He, W., Liu, M., and Song, D.
\newblock Spatially transformed adversarial examples.
\newblock In \emph{6th International Conference on Learning Representations,
  {ICLR} 2018, Vancouver, BC, Canada, April 30 - May 3, 2018, Conference Track
  Proceedings}. OpenReview.net, 2018.
\newblock URL \url{https://openreview.net/forum?id=HyydRMZC-}.

\bibitem[Zhang et~al.(2019)Zhang, Yu, Jiao, Xing, Ghaoui, and
  Jordan]{zhang2019theoretically}
Zhang, H., Yu, Y., Jiao, J., Xing, E.~P., Ghaoui, L.~E., and Jordan, M.~I.
\newblock Theoretically principled trade-off between robustness and accuracy.
\newblock In Chaudhuri, K. and Salakhutdinov, R. (eds.), \emph{Proceedings of
  the 36th International Conference on Machine Learning, {ICML} 2019, 9-15 June
  2019, Long Beach, California, {USA}}, volume~97 of \emph{Proceedings of
  Machine Learning Research}, pp.\  7472--7482. {PMLR}, 2019.
\newblock URL \url{http://proceedings.mlr.press/v97/zhang19p.html}.

\bibitem[Zhang et~al.(2018)Zhang, Isola, Efros, Shechtman, and
  Wang]{zhang2018lpips}
Zhang, R., Isola, P., Efros, A.~A., Shechtman, E., and Wang, O.
\newblock The unreasonable effectiveness of deep features as a perceptual
  metric.
\newblock In \emph{2018 {IEEE} Conference on Computer Vision and Pattern
  Recognition, {CVPR} 2018, Salt Lake City, UT, USA, June 18-22, 2018}, pp.\
  586--595. Computer Vision Foundation / {IEEE} Computer Society, 2018.
\newblock \doi{10.1109/CVPR.2018.00068}.
\newblock URL
  \url{http://openaccess.thecvf.com/content\_cvpr\_2018/html/Zhang\_The\_Unreasonable\_Effectiveness\_CVPR\_2018\_paper.html}.

\end{thebibliography}
\bibliographystyle{icml2023}

\newpage
\appendix
\onecolumn
\section{Categorization of existing defenses against multiple attacks}
\label{app:categorization}
We now categorize existing defenses for multiattack robustness into the different knowledge settings explored and provide a brief description of each defense.
\subsection{Full Knowledge}
The works under the full knowledge category generally study the problem of achieving robustness against unions of attacks (typically $\ell_p$ norms).

\textbf{AVG and MAX training \citep{TB19}}
\citet{TB19} propose 2 methods for adversarial training with a set of different attack types.  In both methods, for every training example, the learner generates adversarial examples with respect to all attacks in the set.  With AVG training, the loss used for backpropagation is taken to be the average of losses across this set of adversarial examples, while with MAX training, the loss is taken to be the maximum loss over the set of adversarial examples.  Since these techniques are targeted towards robustness against the same set of attacks as used during training, this defense is designed as a defense with full knowledge.

\textbf{Multiple steepest descent (MSD) \citep{MainiWK20}} \citet{MainiWK20} improve upon the MAX training algorithm from \citet{TB19} specifically for robustness against union of $\ell_1$, $\ell_{\infty}$, and $\ell_2$ norm attacks.  For these norms, adversarial examples are generally obtained through multiple steps of gradient-based optimization (PGD).  Instead of applying 3 rounds of PGD to obtain $\ell_1$, $\ell_{\infty}$, and $\ell_2$ attacks and then taking the maximum loss for backpropagation, MSD unrolls the PGD steps and at each step of PGD chooses the the PGD update with respect to the norm that maximizes the loss after the update.  This technique also falls under the category of full knowledge because during evaluation, the models are tested on the same attacks as used during the training process.

\textbf{Stochastic adversarial training (SAT) \citep{madaan2020learning}} \citet{madaan2020learning} propose stochastic adversarial training (SAT), where to achieve robustness against a set of attacks, for each input an attack from that set is chosen at random.  The authors then combine this with a regularization which enforces similar distributions of prediction probabilities for clean inputs, adversarial examples, and noisy inputs, with noise generated through their proposed meta-noise generator (MNG).  Since this technique also sees all the attack types in training as during evaluation, this defense also falls under the full knowledge setting.

\subsection{Partial Knowledge}
In practice, there can be mismatch between the attacks used by the learner and attacks used during test-time by the adversary, which motivates studying the partial knowledge setting.  Recently, two works have begun investigating defending in the partial knowledge attacks.  In general, the problem of defending against unforeseen attacks falls under this category.

\textbf{Perceptual adversarial training \citep{laidlaw2020perceptual}} \citet{laidlaw2020perceptual} propose a adversarial training technique called perceptual adversarial training (PAT) which uses attacks that are based on LPIPS metric.  LPIPS \citep{zhang2018lpips} is a distance metric that is based on distances between feature maps when images are passed through a trained neural network (ie. AlexNet) and has been demonstrated to be more perceptually aligned than $\ell_p$ distance metrics.  \citet{laidlaw2020perceptual} show that models trained using PAT can exhibit nontrivial robustness against $\ell_p$ attacks, ReColor attacks \citep{LaidlawF19}, and StAdv attacks \citep{XiaoZ0HLS18}.  Since these attacks were not used during training, this defense falls under the partial knowledge setting.

\textbf{Variation regularization \citep{dai2022formulating}} \citet{dai2022formulating} propose a regularization technique called variation regularization for reducing drop in robust accuracy to unseen threat models.  This regularization technique any two perturbed inputs from the train-time threat model have similar predicted logits.  They, then combine this regularization method with adversarial training methods (such as PAT and PGD adversarial training) and evaluate the regularized model on attacks such as $\ell_p$ attacks, ReColor attacks \citep{LaidlawF19}, and StAdv attacks \citep{XiaoZ0HLS18}, which are outside of the train-time threat model.  Thus, this technique falls under the category of partial knowledge.

\subsection{No Knowledge}
\paragraph{Manifold regularization \citep{jin2020manifold}} Currently, to the best of our knowledge \citet{jin2020manifold} is the only defense which utilizes no knowledge of the test-time threat model.  \citet{jin2020manifold} propose using two regularization terms with standard training, one which reduces the hamming distance of activation patterns between perturbed images and one that reduces the $\ell_2$ Lipschitz constant of the network.  The perturbations used for regularization are not adversarial, instead they are random.  They show that their technique is able to achieve nontrivial $\ell_{\infty}$, $\ell_2$, and LPIPS robustness.  Since adversarial examples are not used during training, this defense falls under the no knowledge category.

\section{Comparison to Existing Evaluation Methods}
Previous works in multiattack robustness generally utilize 4 different approaches for evaluating robustness. In this section, we discuss these techniques and compare our evaluation method to these techniques.

\textbf{Accuracy on individual attack types }Most works in multiattack robustness report robust accuracy on individual attacks at some chosen attack strength $\epsilon$.  For example, works on adversarial training with multiple $\ell_p$ norms \citep{madaan2020learning, MainiWK20, TB19} report robust accuracies for the $\ell_p$ attacks used during training.  While this is approach provides the most information about robustness on individual attacks, typically these numbers are reported for only a few attack types at a single attack strength per attack type.  In our work, we evaluate 9 different attack types with 20 levels of attack strength leading to a larger scope in evaluation.  In our performance visualizations (See Appendix \ref{app:performance_visualizations}), we also allow users to see the CR computed across individual attack types so that users are still able to understand relative performance across each attack type.

\textbf{Accuracy on the union of different attacks }Another commonly reported value is the accuracy across the union of different attack types, which is obtained by considering an image incorrectly classified if any of the attacks in the attack set succeed.  While this metric is a good approximation of worst case robustness, this metric is commonly reported for only a few attack types at a single attack strength per attack type.  This metric is also does not take into account the inherent difficulty of each attack which can bias scores.  For example, consider a setting where one attack $P$ in the evaluation set is inherently more difficult than the rest and the best model for this attack can do no better than random guessing.  In this case, we would always expect the union accuracy to be highly biased by $P$ and always have value less than $\frac{1}{K}$ where $K$ is the number of classes.  Our metric $\text{CR}_{\text{ind-worst}}$ addresses this bias by weighting the robust accuracy of the defense by $\frac{1}{\text{acc}^*(P)}$.

\textbf{Average accuracy across attacks }Another value reported by papers in multiattack robustness is average accuracy across attacks.  For example, \citet{laidlaw2020perceptual} report average accuracy across unseen attacks to demonstrate improved robustness against unseen attacks.  Similar to union accuracy, this metric can also be biased by attack difficulty.  Our metric $\text{CR}_{\text{ind-avg}}$ addresses this bias by weighting the robust accuracy of the defense by $\frac{1}{\text{acc}^*(P)}$.

\textbf{mUAR metric \citep{kang2019robustness} }\citet{kang2019robustness} introduce a metric called mUAR for evaluating robustness against unseen attacks.  Specifically, this value is defined as follows:
\begin{definition}[mUAR \citep{kang2019robustness}] Let $K$ be a set of different attack types $P$.  Let $\text{acc}(h, P, \epsilon)$ denotes the robust accuracy of defended model $h$ using attack $P$ with attack strength $\epsilon$.  Let $\text{acc}^*(P, \epsilon)$ denote the best accuracy obtainable from a model in $\mathcal{H}$.  For each $P_i \in K, i \in \{1 ... |K|\}$, let $\mathcal{E}_i$ be a corresponding set of attack strengths.  Then, for a model $h$, 

\begin{multicols}{2}
  \begin{equation}
    \text{UAR}(h, P, \mathcal{E}) = 100 \times \frac{\sum_{\epsilon \in \mathcal{E}} \text{acc}(h, P, \epsilon)}{\sum_{\epsilon \in \mathcal{E}} \text{acc}^*(P, \epsilon)}
  \end{equation} \break
  \begin{equation}
    \text{mUAR}(h) = \frac{1}{|K|} \sum_{i = 1}^{|K|} \text{UAR}(h, P_i, \mathcal{E}_i)
  \end{equation}
\end{multicols}

\end{definition}
From the definition of UAR, for a single attack type the aggregated robust accuracies across attack strengths $\epsilon$ are weighted by the aggregate best accuracy attainable, which addresses the problem of bias from evaluated across different values of $\epsilon$.  However, when considering multiple attacks mUAR weights the scores of each attack equally, so this score can be still be biased by the difficulties of each attack type.  In comparison, our CR metrics are weighted across different attack types as well.  We also note that by using a different definition of multiattack error and using a single attack type in our evaluation set, we can obtain the UAR metric from CR (see Appendix \ref{app:multiattack_err} for more discussion).

\textbf{CARBEN \citep{hsiung2022carben} }\citep{hsiung2022carben} propose a benchmark for measuring compositional robustness called CARBEN.  In CARBEN evaluates models by optimizing the attack order of a set of attacks at a single attack strength (specifically $\ell_{\infty}$, hue, saturation, rotation, brightness, contrast) and reporting the robust accuracy of the model after performing that sequence of attacks.  In general, we find that hue, saturation, rotation, brightness, and contrast attacks are much weaker than existing (and less perceptible) attacks such as StAdv and ReColor, so the accuracies from the CARBEN benchmark does not reflect multiattack robustness well.  Additionally, we evaluate on multiple attack strengths for each attack and use CR metrics due to potential bias from using accuracy.

\section{Description of Attacks Used and Evaluation Procedure}
\label{app:attack_desc}
\begin{figure}[]
    \centering
    \includegraphics[width=\textwidth]{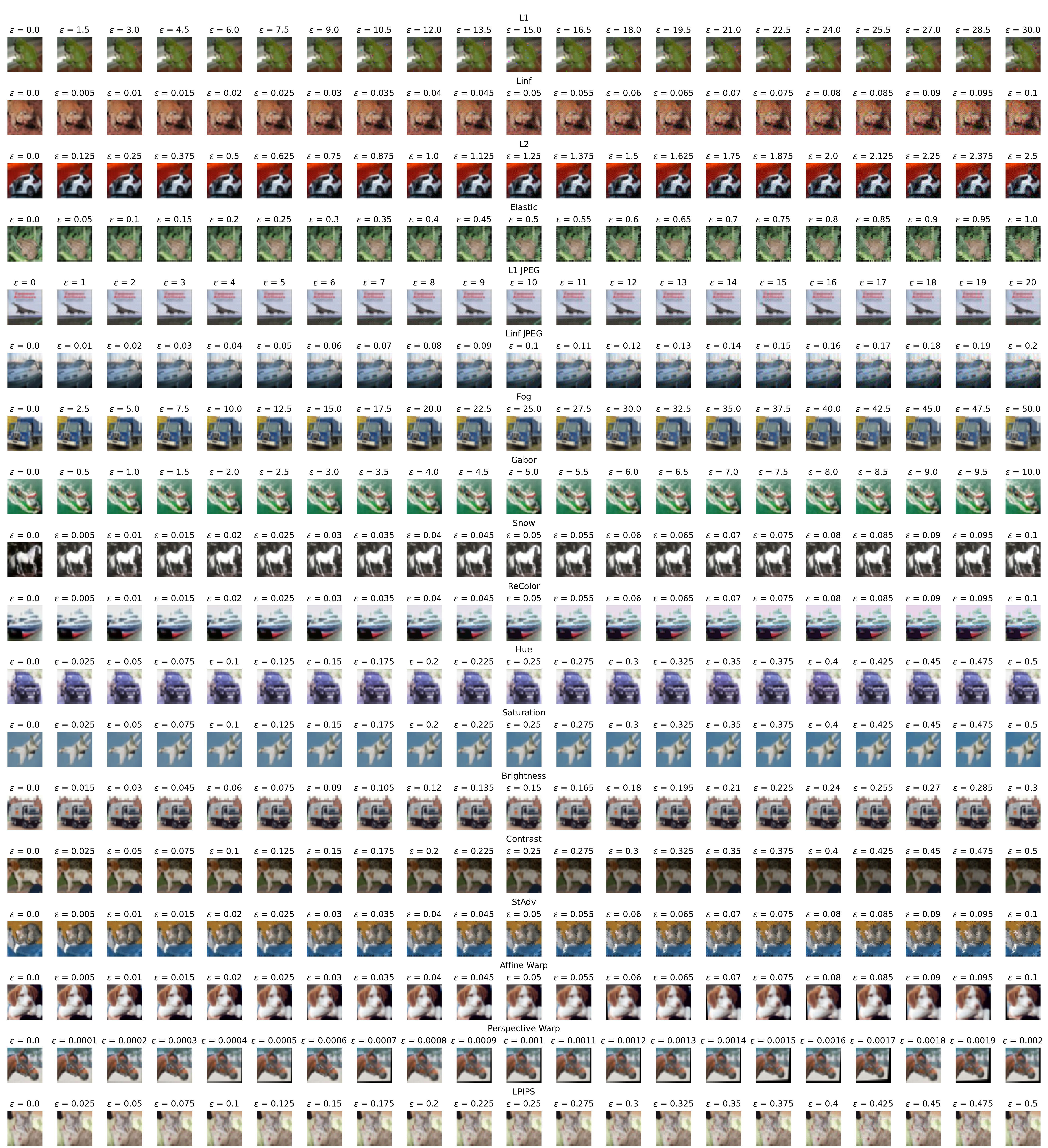}
    \caption{Samples of all attacks at each attack strength $\epsilon$ used in evaluation.}
    \label{fig:attack_samples}
\end{figure}

In this section, we describe in more depth the attacks used by the benchmark and evaluation procedure for computing CR and stability scores.  We include samples of each attack at the strengths evaluated in Figure \ref{fig:attack_samples}

\subsection{Attack descriptions}
\label{app:attack_desc2}
\textbf{$\ell_p$ attacks} The most commonly studied form of robustness is robustness to $\ell_p$ attacks, mainly $\ell_1$, $\ell_2$, and $\ell_{\infty}$ attacks.  AutoAttack \citep{croce2020reliable} is a commonly used package for evaluating the robustness which includes a combination of 3 white box attacks (APGD-CE, APGD-T, and FAB-T) and a black box attack (Square).  Since we evaluate at 20 different attack strengths for each attack type, we evaluate with only the APGD-T and FAB-T attacks to reduce evaluation time.  For $\ell_1$ attacks, we evaluate with attack strength $\epsilon \in (0, 30]$ in increments of 1.5.  For $\ell_2$ attacks, we evaluate with $\epsilon \in (0, 2.5]$ in increments of 0.125.  For $\ell_{\infty}$ attacks, we evaluate with $\epsilon \in (0, 0.1)$ in increments of 0.005.

\textbf{Spatial transformation attacks} For the leaderboard, we measure robustness to spatial transformations using StAdv attack \citep{XiaoZ0HLS18}.  This attack generates adversarial examples by optimizing for a per pixel flow field $f$, where $f_i$ corresponds to the displacement vector of the $i$th pixel of the image.  This flow field is obtained by solving $\argmin_f \ell_{\text{adv}}(x, f) + \tau \ell_{\text{flow}}(f) $ 
where $\ell_{\text{adv}}$ is the CW objective \citep{carlini2017towards} and $\ell_{\text{flow}}$ is a regularization term that controls the smoothness of the change. $\tau$ is a hyperparameter controlling regularization strength. For StAdv attacks, we evaluate with $\epsilon \in (0, 0.1]$ in increments of $0.005$ and set $\tau = 0.0025 / \epsilon$.
  
Outside of StAdv attacks, we also allow users to see robust accuracies for attacks that apply global spatial transformations including affine warp and perspective warp (though these are not included in the leaderboard ranking as the attacks are much easier than StAdv).  An affine warp is a transformation captured by a matrix of the form  $M_{\text{affine}} = \begin{pmatrix}
x_{1,1} & x_{1,2} & x_{1,3}\\
x_{2,1} & x_{2,2} & x_{2,3}\\
0 & 0 & 1
\end{pmatrix}$ and the pixel coordinates of the resulting image is obtained by multiplying $M_{\text{affine}}$ to each pixel coordinate vector $(x, y, 1)^T$.  Affine transformations capture translations, rotations, scaling, and shear.  To generate adversarial affine transformations, we use PGD to optimize over $M_{\text{affine}}$ and apply an $\ell_{\infty}$ constraint to $M_{\text{affine}}$.  For affine attacks, we use $\ell_{\infty}$ bounds in range (0, 0.1) with increments of 0.005.  Perspective warps capture more transformations than affine warps and can be parameterized by a matrix $M_{\text{perspective}} = \begin{pmatrix}
x_{1,1} & x_{1,2} & x_{1,3}\\
x_{2,1} & x_{2,2} & x_{2,3}\\
x_{3,1} & x_{3,2} & 1
\end{pmatrix}$.  We apply the same method to optimize over $M_{\text{perspective}}$ and apply an $\ell_{\infty}$ constraint.  For perspective attacks, we use $\ell_{\infty}$ bounds in range (0, 0.002) with increments of 0.0001.

\textbf{Color shift attacks} For the leaderboard, we measure robustness to color shifts via ReColor attacks \citep{LaidlawF19}.  The approach to generating ReColor attacks is similar to the objective of StAdv attacks: $\argmin_f \ell_{\text{adv}}(x, f) + \tau \ell_{\text{flow}}(f)$, where $f$ now models a function that maps between colors and $\ell_{\text{flow}}(f)$ is a regularization term that ensures that neighboring pixels will have colors changed in a similar manner.  For ReColor attacks, we evaluate with $\epsilon \in (0, 0.1]$ in increments of $0.005$ and set $\tau = 0.0036 / \epsilon$.

We also allow users to see evaluations for global color changes including hue shifts, brightness changes, contrast changes, and saturation changes (but these scores are not used for computing leaderboard ranking as they are much weaker attacks than ReColor).  All of these color changes can be parameterized by a single scalar parameter, and we use PGD to optimize this parameter as in \citet{tsai2022compositional}.  For hue and saturation attacks, we consider changes in the parameter from (0, 0.5] with increments of 0.025.  For brightness, we consider changes from (0, 0.3] with increments of 0.015, and for contrast, we consider changes from (0, 0.5] with increments of 0.025.

\textbf{UAR attacks \citep{kang2019robustness}} \citet{kang2019robustness} propose a set of attacks for evaluating unforeseen robustness including attacks such as elastic attacks, $\ell_1$ JPEG attacks, $\ell_{\infty}$ JPEG attacks, snow, fog, and Gabor attacks.  Of these attacks, elastic attacks, $\ell_1$ JPEG attacks, and $\ell_{\infty}$ JPEG attacks are targeted towards the CIFAR-10 dataset, so we incorporate these attacks into leaderboard ranking.  Elastic attacks are a spatial attack based off of StAdv where the flow field $f$ is obtained by smoothing a vector $W$ by a Gaussian kernel and optimizing over $W$ under the constraint that $||W||_{\infty} \le \epsilon$.  For elastic attacks, we consider $\epsilon \in [0, 1)$ in increments of 0.05.  $\ell_1$ (or $\ell_{\infty}$) JPEG attacks optimize for $\ell_1$ (or $\ell_{\infty}$) bounded adversarial examples in the JPEG-encoded space of compressed images.  For $\ell_1$ JPEG attacks, we consider $\ell_1$ bounds in range (0, 20] in increments of 1.  For $\ell_{\infty}$ JPEG attacks, we consider $\ell_{\infty}$ bounds in range (0, 0.2] in increments of 0.01.

While they are not included in leaderboard ranking, we allow users to see the scores for snow, fog, and Gabor attacks on the leaderboard site.  For snow attacks, we provide evaluations for attack strengths in range (0, 0.1) in increments of 0.005.  For fog attacks, we provide evaluations for attack strengths in range (0, 50) in increments of 2.5.  For gabor attacks, we provide evaluations for attack strengths in range (0, 10) in increments of 0.5.

\textbf{Bounded LPIPS attacks \citet{laidlaw2020perceptual}} LPIPS distance \citep{zhang2018lpips}, is a distance metric based on distances between feature maps of trained models (ie. AlexNet) and has been shown to be more perceptually aligned than $\ell_p$ metrics.  \citet{laidlaw2020perceptual} introduce 2 attacks, perceptual PGD (PPGD) and Lagrange perceptual attack (LPA) based on this distance.  PPGD optimizes for adversarial examples in a way that is analogous to PGD: at each iteration, an optimal first order step is taken and then projected so that it lies in the LPIPS bound.  LPA moves the distance constraint into the objective function via Lagrangian relaxation.  For evaluation, we measure robustness against LPIPS attacks by taking the accuracy against the union of PPGD and LPA attacks based off of AlexNet architecture.  We consider attacks of LPIPS bound in range (0, 0.5] in increments of 0.025.

\subsection{Additional evaluation details}
In this section, we describe additional details about how we perform metric computations.

\textbf{Training models for approximating optimal single attack accuracy } To approximate optimal single attack accuracy, for each attack, we train 3 ResNet-18 models using adversarial training.  We also train a set of 3 models with standard training (no attack).  For $\ell_1$ threat model, we train using adversarial examples generated via APGD \citep{pmlr-v139-croce21a}.  For $\ell_2$ and $\ell_{\infty}$ threat models, we use PGD-adversarial training.  For LPIPS threat model, we use perceptual adversarial training (PAT) which uses a fast approximation to the LPA attack \citet{laidlaw2020perceptual}.  For all other threat models, we the same attack generation method during training as used for evaluation as described in the previous section.  For all threat models (with the exception of $\ell_1$ threat model which uses settings from \citet{pmlr-v139-croce21a} and UAR attacks which use default settings from \citet{kang2019robustness}), we use 20 iterations to find adversarial examples with step size $\epsilon / 18$.  We train all models with batch size of 256 for 100 epochs and evaluate the model saved at the epoch which achieves highest robust accuracy on the test set.  We train models using SGD with initial learning rate of 0.1.  Learning rate drops to 0.01 after half of the training epochs and drops to 0.001 after 3/4 of the training epochs.

\textbf{Robust accuracy evaluation } For evaluating robust accuracy, with the exception of $\ell_p$ attacks and UAR attacks which use default evaluation setups from \citet{croce2020reliable} and \citet{kang2019robustness} respectively, we use 20 iterations to optimize over adversarial examples for non-LPIPS threat models and 40 iterations to optimize over adversarial examples for LPIPS threat model.  To obtain robust accuracies at multiple epsilon per attack type, we perform robustness evaluations in a binary search manner to find the smallest perturbation size at which the model misclassifies each image.  We aggregate this information across the entire test set to find the robust accuracy at each value of epsilon.

\textbf{Metric computation } For computing $\text{CR}_{\text{ind-avg}}$ and $\text{CR}_{\text{ind-worst}}$, we follow Definition \ref{def:crind} and take $K$ to be the set of 180 attacks described in \ref{app:attack_desc2} (9 attacks at 20 different values of $\epsilon$ each) with 1 additional attack representing no attacker ($\epsilon = 0$).  For $\text{CR}_{\text{ind-avg}}$, we assume that the distribution over all attacks is uniform.

For computing stability constant, we consider $K_{\text{learner}}$ to include attack strength $\epsilon$ from 0 (no attack) to the $\epsilon$ that is used by the defense that fall within the 20 values of $\epsilon$ that we used for evaluating robust accuracy.  For example, if a defense uses $\ell_2$ threat model with strength 0.5, and in our evaluation procedure for $\ell_2$, we evaluate with $\epsilon \in (0, 2.5]$ in increments of 0.125, we would consider $K_{\text{learner}}$ to contain no attack ($\epsilon = 2$) and $\ell_2$ attacks with $\epsilon \in \{0.125, 0.25, 0.375, 0.5\}$.  We then follow the equation in Definition \ref{def:stability} with $\alpha = 3\%$.  We note that we found that both reducing $\alpha$ to 1\% and increasing $\alpha$ generally does not influence the SC of defended models.  The only defense whose SC changed as a result of changing $\alpha$ was \citep{jin2020manifold} for which $K_{\text{learner}}$ does not contain any attacks (outside of $\epsilon=0$), for which decreasing $\alpha$ to 1\% reduces stability constant to 0 due to few $\text{acc}^*$ values that lie in the viscinity of standard training clean accuracy.  We choose 3\% to ensure that \citet{jin2020manifold} and other future defenses which may also be based on standard training will still have a value for SC for comparison.

\section{Defenses Present on Leaderboard}
\label{app:defenses}
Overall, we evaluate a total of 16 different models.  All defenses use one of the follow training threat models: no attack (standard training), $\ell_2$ with $\epsilon=\frac{128}{255}$, $\ell_{\infty}$ with $\epsilon=\frac{8}{255}$, (AlexNet) LPIPS with $\epsilon=1$, a combination of $\ell_1$ with $\epsilon=\frac{2000}{255}$, $\ell_2$ with $\epsilon=\frac{128}{255}$, and $\ell_{\infty}$ with $\epsilon=\frac{8}{255}$.  We describe the entries present on our leaderboard below:
\begin{itemize}
    \item Single attack adversarial training approaches: Since adversarial training with $\ell_2$ and $\ell_{\infty}$ is commonly studied, we include 2 entries (one with $\ell_2$ attacks, and one with $\ell_{\infty}$ attacks) for PGD adversarial training \citep{madry2017towards} and 2 entries for TRADES adversarial training \citep{zhang2019theoretically} (one with $\ell_2$ attacks, and one with $\ell_{\infty}$ attacks).  All 4 of these entries use ResNet-18 architecture.  Prior works have improved on the performance of $\ell_2$ and $\ell_{\infty}$ through the use of additional synthetic data.  One of these approaches is \citep{rebuffi2021fixing} which is currently the top performing approach on RobustBench \citep{croce2020robustbench}.  We include entries for two WRN-28-10 trained using \citep{rebuffi2021fixing} (one with $\ell_{\infty}$ threat model and the other with $\ell_2$ threat model).  The pretrained models for \citet{rebuffi2021fixing} models are available through RobustBench.
    
    \item Multiple attack adversarial training approaches: We also include entries for multiple defenses trained with a combination of $\ell_1$ with $\epsilon=\frac{2000}{255}$, $\ell_2$ with $\epsilon=\frac{128}{255}$, and $\ell_{\infty}$ with $\epsilon=\frac{8}{255}$.  These include 2 models using the training approach from \citet{madaan2020learning} (one which uses additional data via robust self-training \citep{carmon2019unlabeled} and one that does not use additional data), 1 model using the AVG approach in \citet{TB19}, 1 model using the MAX approach in \citet{TB19}, and 1 model using MSD from \citet{MainiWK20}.  These models are available through the code repository for \citet{madaan2020learning} \hyperlink{https://github.com/divyam3897/MNG_AC}{here}.  These models all use WRN-28-10 architecture.
    
    \item Variation regularization \citep{dai2022formulating}: \citet{dai2022formulating} propose variation regularization which can be applied on top of any train-time threat model.  We include 3 leaderboard entries for this defense, one using $\ell_{\infty}$ threat model, one using $\ell_2$ threat model, and one using LPIPS threat model.  The 
    $\ell_{\infty}$ and $\ell_2$ models both use ResNet-18 architecture while the LPIPS model uses ResNet-50 architecture.  These models are available through the code repository for \citet{dai2022formulating} \hyperlink{https://github.com/inspire-group/variation-regularization}{here}.
    
    \item Perceptual adversarial training (PAT) \citep{laidlaw2020perceptual}: We include an entry for PAT with the AlexNet-based LPIPS attacks.  This model uses ResNet-50 architecture and is available through the code repository for \citet{laidlaw2020perceptual} \hyperlink{https://github.com/cassidylaidlaw/perceptual-advex}{here}.
    
    \item Manifold regularization \citep{jin2020manifold}: \citet{jin2020manifold} propose a regularization technique that can be applied on top of standard training and does not use adversarial examples to compute.  We include a leaderboard entry for manifold regularization.  This model uses ResNet-18 architecture.  The pretrained model is available \hyperlink{https://www.dropbox.com/sh/jfqo2l1i2dyj3o9/AABWq8IQZmQlMLJ_ntMj-TYVa?dl=0}{here}.
\end{itemize}

\section{Additional CR Definitions}
\label{app:multiattack_err}
In the main body of the paper, we focused mainly on using $\text{err}_{\text{multi-ind}}$ as the multiattack error when defining CR.  Additionally, we can consider using $\text{err}_\text{multi-exp}$ and $\text{err}_{\text{multi-max}}$ which leads to 2 new definitions of CR:
\begin{equation}
    \text{CR}_{\text{exp}}(h; K) = 100 \times \frac{\mathbb{E}_{P\sim \mathcal{D}(K)} \text{acc}(h, P)}{\mathbb{E}_{P\sim \mathcal{D}(K)} \text{acc}^*(P)}
\end{equation}
\begin{equation}
    \text{CR}_{\text{max}}(h; K) = 100 \times \frac{\min_{P \in K} \text{acc}(h, P)}{\min_{P \in K} \text{acc}^*(P)}
\end{equation}

We note that for $\text{CR}_{\text{exp}}(h; K)$, when $\mathcal{D}(K)$ is uniform and $K$ contains only attacks of the same type, we obtain the UAR metric proposed by \citet{kang2019robustness}.

For leaderboard rankings, we opted to use the $\text{err}_\text{multi-ind}$ definitions of CR since more clearly compares robust accuracy on each specific attack to the corresponding robust accuracy of the optimal, making these metrics more interpretable, while $\text{CR}_{\text{exp}}$ and $\text{CR}_{\text{max}}$ both compare aggregates across all defense accuracies and across all optimal accuracies.  We find that $\text{CR}_{\text{exp}}$ and $\text{CR}_{\text{max}}$ both lead to higher scores relative to $\text{CR}_{\text{ind-avg}}$ and $\text{CR}_{\text{ind-worst}}$ respectively.  We also find that ranking by $\text{CR}_{\text{exp}}$ maintains the rankings of the top 3 best performing models compared to $\text{CR}_{\text{ind-avg}}$.  For $\text{CR}_{\text{max}}$, we find that the set of top 3 best performing models stays the same, but the rankings are reversed compared to $\text{CR}_{\text{ind-worst}}$.

\section{Additional Leaderboard Features}
\label{app:performance_visualizations}
While the scores present on the leaderboard allow us to easily compare the performance of defended models for multiattack robustness, it is hard to understand failure points of specific defenses by looking at the score alone.  To this end, we provide additional features that allow users to have a more in depth understanding of model performance.

\begin{figure*}[ht]
    \centering
    \includegraphics[width=\textwidth]{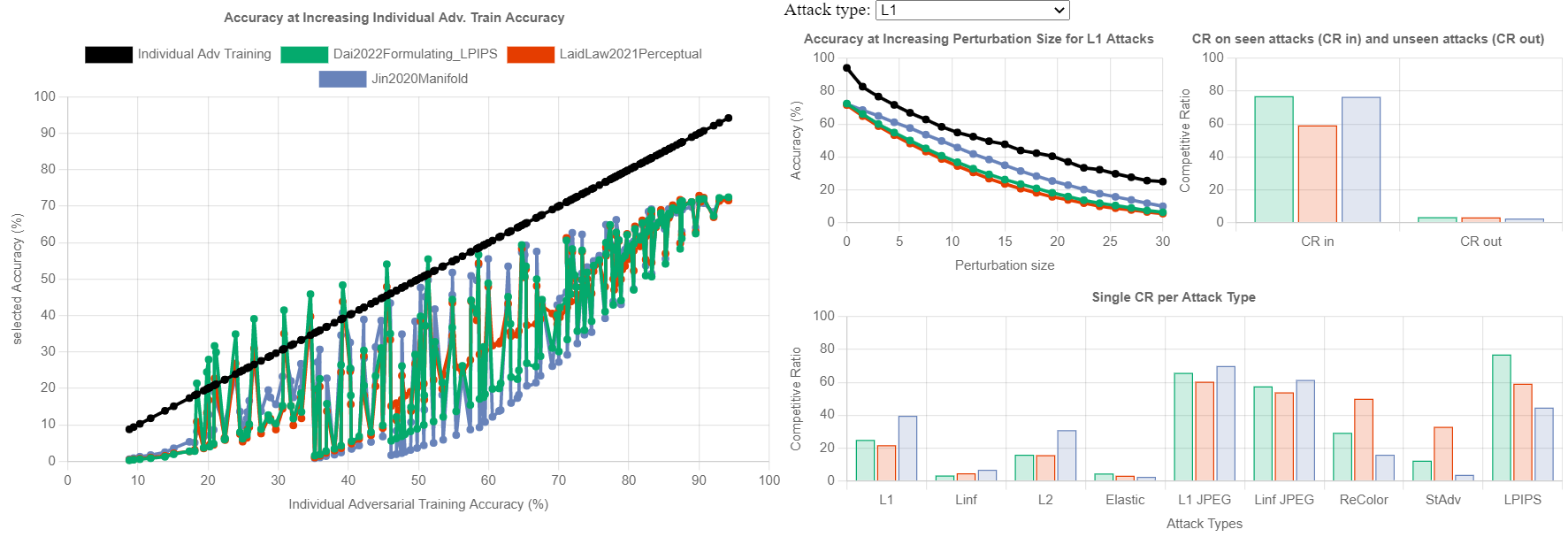}
    \caption{Sample performance visualizations provided on our leaderboard website.}
    \label{fig:sample_vis}
\end{figure*}

\textbf{User controls} Since some defenses are designed for robustness against a union of specific attack types, we allow users to control the types of attacks used in attack set $K$ used when computing metrics.  We note that leaderboard rankings, however, are independent of user selection since the goal of the leaderboard is to reflect performance across a diverse set of imperceptible attacks.

\textbf{Performance Visualizations} While metrics like CR and stability are useful for ranking and comparing the performance of different defenses, it is difficult to understand specific weaknesses of existing defenses, which may make it difficult to improve upon existing techniques.  For example, a defense may systematically fail on a particular attack type, but since CR and stability aggregate performance across multiple attack types, they are unable to convey this.  To address this, our leaderboard allows users to generate performance visualizations for specific defenses and compare performance visualizations for up to 5 different defenses.  We provide 4 types of performance visualizations: (1) a graph of the accuracy of the defense on each tested perturbation type against the accuracy of adversarial training directly on that perturbation type, (2) graphs of robust accuracy across attack strength $\epsilon$ for each perturbation type tested, (3) a bar chart of CR-in compared to CR-out (4) a bar chart of single-CR values computed for each perturbation type tested.  Examples of these visualizations are shown in Figure \ref{fig:sample_vis}.

\section{Experimental Setup for Adversarially Trained Models}
\label{app:setup}
To understand the impact of factors such as architecture size, additional data, and number of training epochs on multiattack performance, we train our own set of models using adversarial training on 3 different threat models: $\ell_2$ (with radius $\epsilon = 0.5$), $\ell_{\infty}$ (with $\epsilon =  \frac{8}{255}$), and LPIPS (with $\epsilon = 0.5$).  For $\ell_{\infty}$ and $\ell_2$ threat models, we train using 20 iterations of PGD with step size $\epsilon / 18$.  For LPIPS threat model we use PAT with 20 iterations for Fast LPA to find adversarial examples \citep{laidlaw2020perceptual}.  For all experiments, we train 3 trials. We train models using SGD with initial learning rate of 0.1.  Learning rate drops to 0.01 after half of the training epochs and drops to 0.001 after 3/4 of the training epochs. For all evaluations, we use the same set of attacks as described in Apendix \ref{app:attack_desc2}, but for $\ell_{\infty}$ and $\ell_2$ attacks, we use 10 step PGD to find adversarial examples (with step size $\epsilon / 8$), and for $\ell_1$ attacks we use APGD \citep{pmlr-v139-croce21a}.

\textbf{Architecture experiments} We train ResNet-18, ResNet-34, ResNet-50, and ResNet-101 models with batch size 256 for 100 epochs and evaluate at the model saved at the epoch achieving the highest robust accuracy on the test set.

\textbf{Extra training data experiments} We train ResNet-18 models with and without extra 1M (synthetic) training data from \citet{gowal2021improving}.  Models are trained in batches of 150 samples from the original training set and 350 samples from the extra training data.  We train for 100 epochs and evaluate at the model saved at the epoch achieving the highest robust accuracy on the test set.

\textbf{Training epoch experiments} We train ResNet-18 models with batch size of 256 for 200 epochs and save a copy of the model every 5 epochs.  We evaluate on this set of saved models.

\section{Additional Analysis}
\subsection{Stability of existing defenses}
\label{app:additional_defense_analysis}
Since the computation of stability depends on the choice of $K_{\text{learner}}$ for fair comparison, we should compare models which use the same threat model during training.  In Table \ref{tab:existing_stab}, we organize defenses present on our leaderboard by train-time threat model and report their corresponding stability constants.  We note that the only model missing from Table \ref{tab:existing_stab} is model using the manifold regularization defense from \citet{jin2020manifold} as it is the only model that does not use adversarial examples during training.

\begin{table}[h]
    \begin{subtable}[h]{0.45\textwidth}
        \centering
        \begin{tabular}{|c|c|}
        \hline
        Defense & Stability Constant \\
        \hline
        \citet{dai2022formulating} & 1801.00 \\
        \citet{rebuffi2021fixing} & 2056.50 \\
        \citet{zhang2019theoretically} & 2164.00 \\
        \citet{madry2017towards} & 2309.50 \\
        \hline
        \end{tabular}
       \caption{$\ell_{\infty}$}
        \label{subtab:linf_stab}
    \end{subtable}
    \hfill
    \begin{subtable}[h]{0.45\textwidth}
        \centering
        \begin{tabular}{|c|c|}
        \hline
        Defense & Stability Constant \\
        \hline
        \citet{rebuffi2021fixing} & 457.71 \\
        \citet{dai2022formulating} & 904.71 \\
        \citet{zhang2019theoretically} & 940.29 \\
        \citet{madry2017towards} & 1110.00 \\
        \hline
    \end{tabular}
        \caption{$\ell_2$}
        \label{subtab:l2_stab}
     \end{subtable}
     \hfill
     \begin{subtable}[h]{0.45\textwidth}
        \centering
        \begin{tabular}{|c|c|}
        \hline
        Defense & Stability Constant \\
        \hline
        \citet{dai2022formulating} & 4947.00 \\
        \citet{laidlaw2020perceptual} & 5823.00 \\
        \hline
    \end{tabular}
        \caption{LPIPS}
        \label{subtab:lpips_stab}
     \end{subtable}
     \hfill
     \begin{subtable}[h]{0.48\textwidth}
        \centering
        \begin{tabular}{|c|c|}
        \hline
        Defense & Stability Constant \\
        \hline
        \citet{madaan2020learning} (no RST) & 1494.00 \\
        \citet{MainiWK20} & 1803.00 \\
        \citet{madaan2020learning} (RST) & 1901.25 \\
        \citet{TB19} (MAX) & 2145.00	 \\
        \citet{TB19} (AVG) & 2502.00	 \\
        \hline
    \end{tabular}
        \caption{$\ell_1, \ell_2, \ell_{\infty}$}
        \label{subtab:lp_stab}
     \end{subtable}
    
     \caption{Stability constants of models present on the leaderboard}
     \label{tab:existing_stab}
\end{table}
We note that of all defenses tested, the defense from \citet{dai2022formulating} is specifically designed for improving stability (which \citet{dai2022formulating} refers to as unforeseen generalization gap), and we find that for $\ell_{\infty}$, $\ell_{2}$, and LPIPS threat models, the model using \citet{dai2022formulating} outperforms the corresponding baseline (\citep{madry2017towards} for $\ell_{\infty}$ and $\ell_2$ norms and \citep{laidlaw2020perceptual} for LPIPS).

\subsection{$\text{CR}_{\text{ind-worst}}$ per attack type analysis for LPIPS trained models}
In this section, we present computed $\text{CR}_{\text{ind-worst}}$ values across individual attack types for LPIPS trained models in Section \ref{sec:design_choice_impact}.
\subsubsection{Impact of architecture size }
\label{app:lpips_per_attack_arch}
\begin{figure*}[ht]
    \centering
     
    \begin{subfigure}[t]{0.3\textwidth}
    \includegraphics[width=\textwidth]{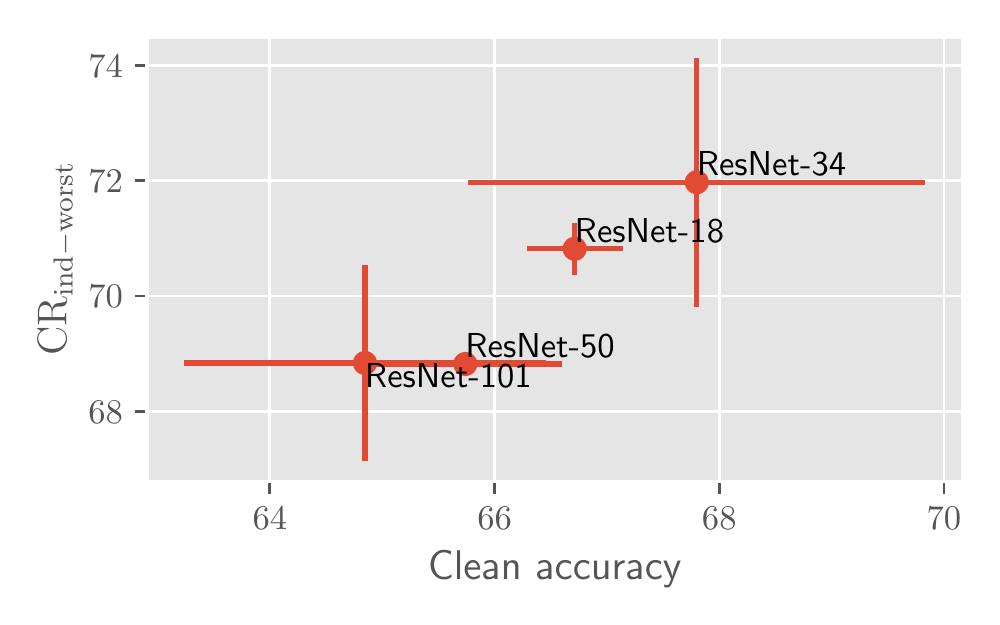}
        \caption{LPIPS}
    \label{subfig:arch_per_atk_lpips}
    \end{subfigure}
    \begin{subfigure}[t]{0.3\textwidth}
    \includegraphics[width=\textwidth]{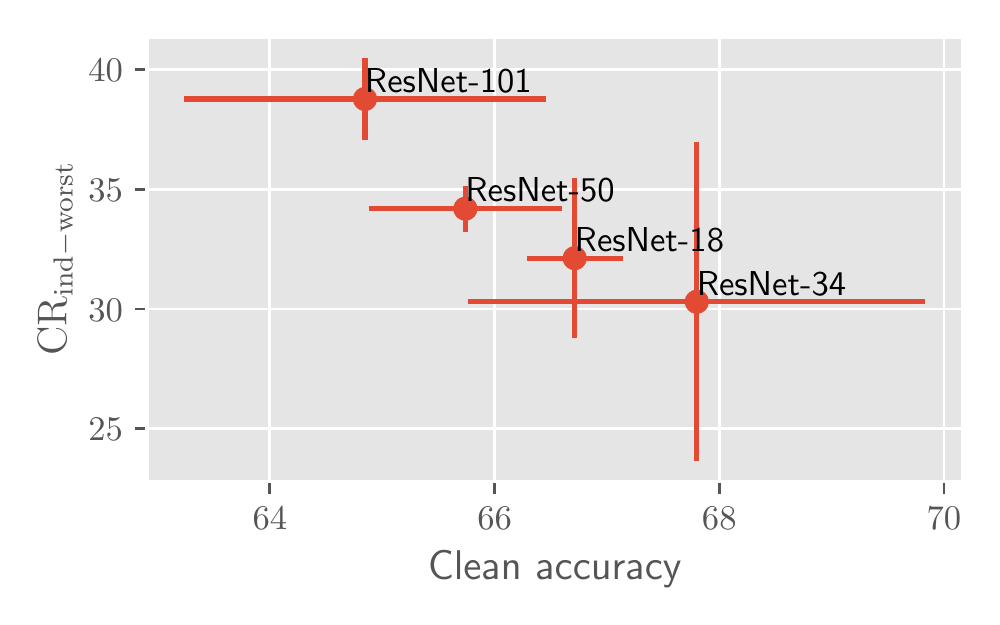}
        \caption{StAdv}
    \label{subfig:arch_per_atk_stadv}
    \end{subfigure}
    \begin{subfigure}[t]{0.3\textwidth}
    \includegraphics[width=\textwidth]{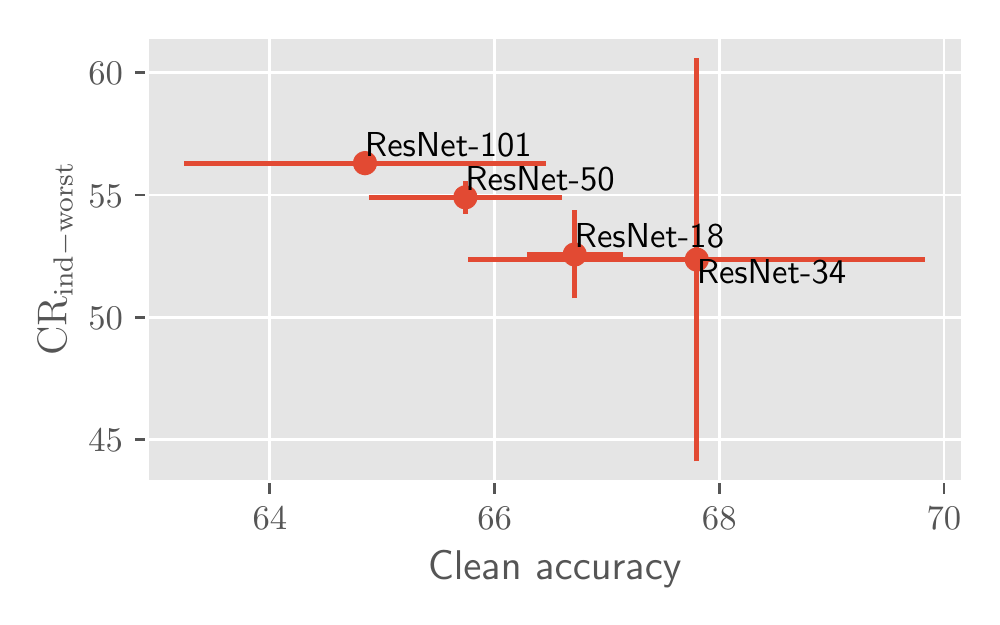}
    \caption{ReColor}
    \label{subfig:arch_per_atk_recolor}
    \end{subfigure}
    
    \begin{subfigure}[t]{0.3\textwidth}
    \includegraphics[width=\textwidth]{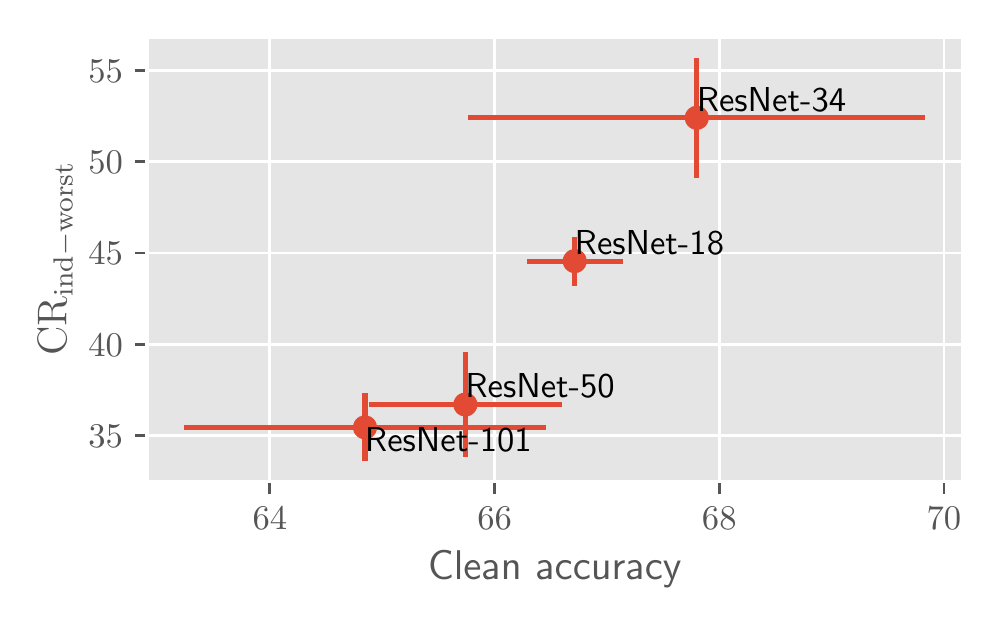}
    \caption{$\ell_1$}
    \label{subfig:arch_per_atk_l1}
    \end{subfigure}
    \begin{subfigure}[t]{0.3\textwidth}
    \includegraphics[width=\textwidth]{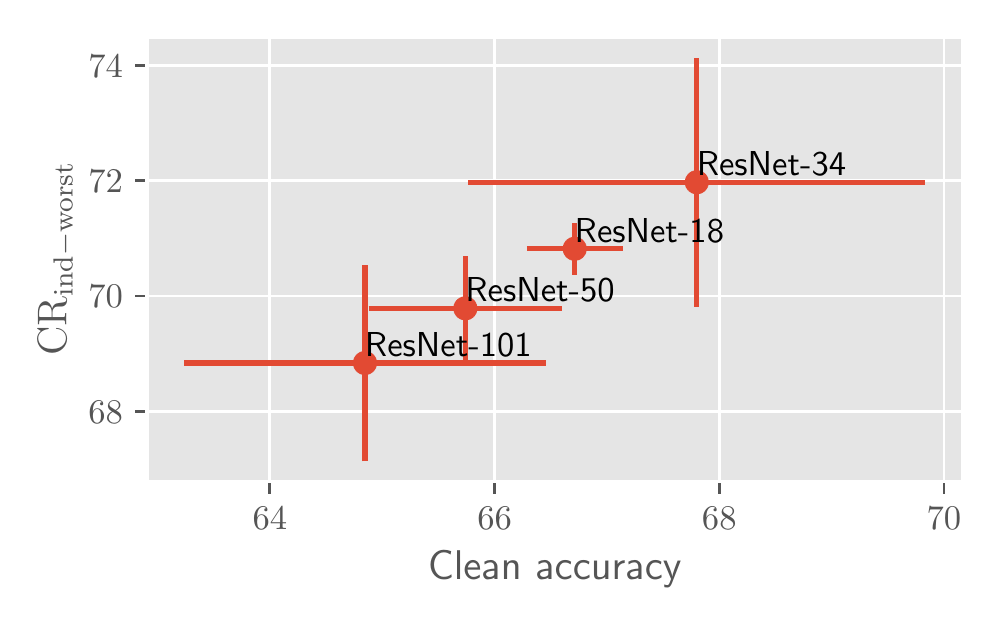}
    \caption{$\ell_2$}
        \label{subfig:arch_per_atk_l2}
    \end{subfigure}
    \begin{subfigure}[t]{0.3\textwidth}
    \includegraphics[width=\textwidth]{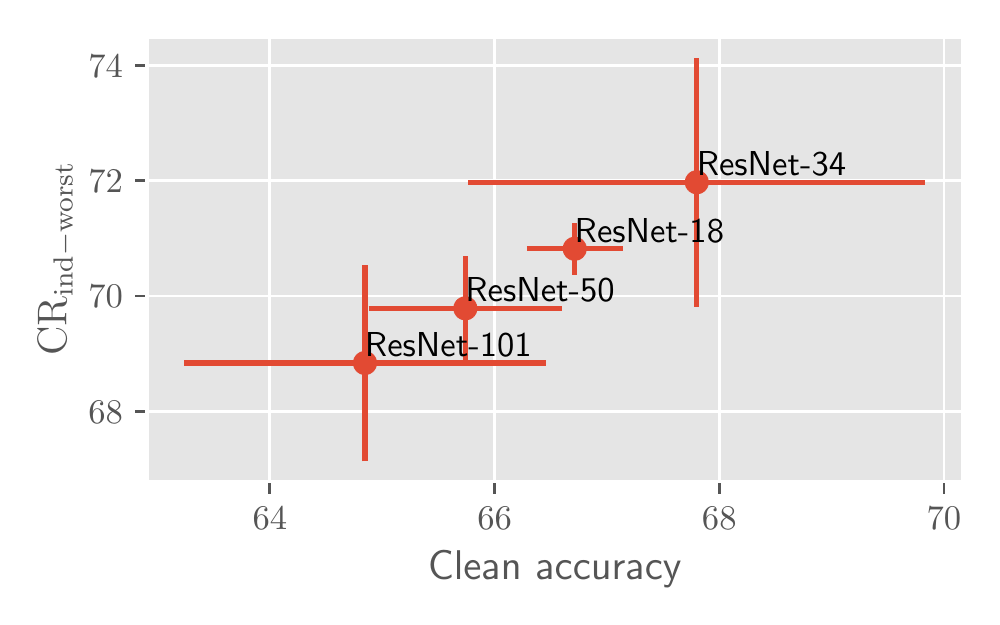}
    \caption{$\ell_\infty$}
    \label{subfig:arch_per_atk_linf}
    \end{subfigure}

    \begin{subfigure}[t]{0.3\textwidth}
    \includegraphics[width=\textwidth]{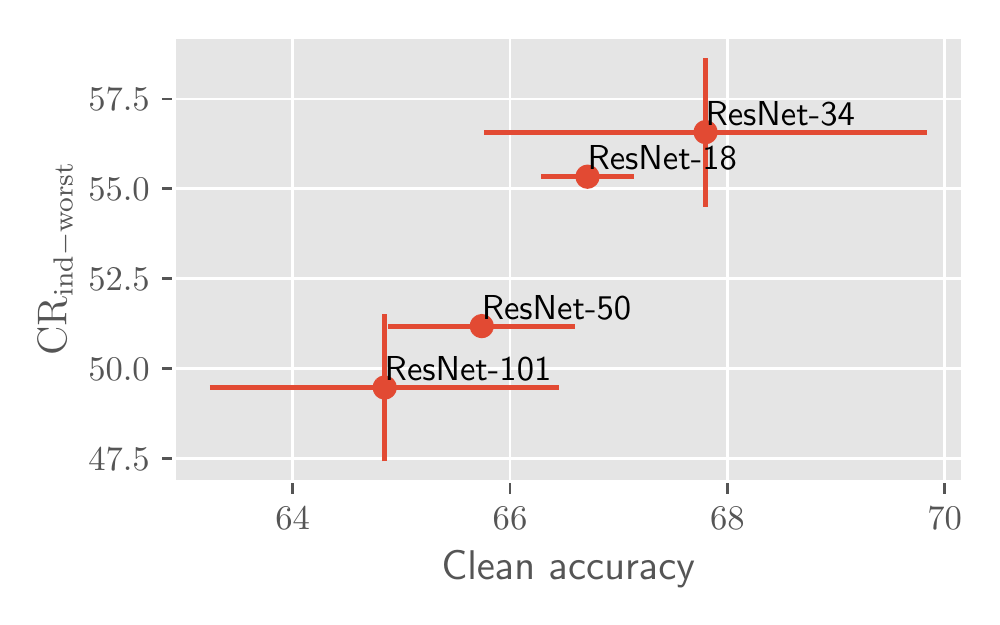}
        \caption{$\ell_1$-JPEG}

    \label{subfig:arch_per_atk_l1_jpeg}
    \end{subfigure}
    \begin{subfigure}[t]{0.3\textwidth}
    \includegraphics[width=\textwidth]{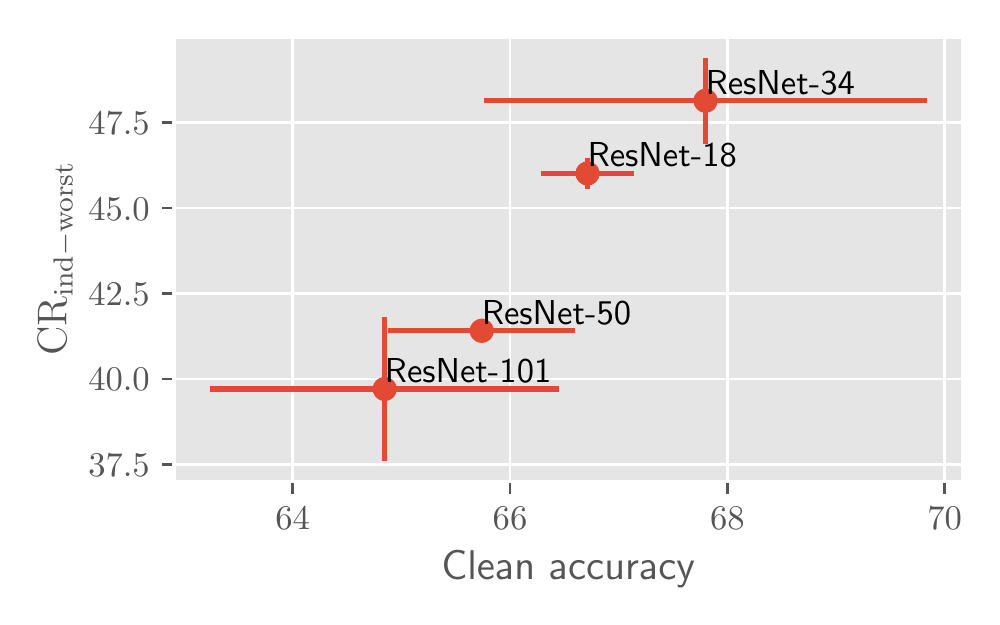}
        \caption{$\ell_\infty$-JPEG}
        \label{subfig:arch_per_atk_linf_jpeg}
    \end{subfigure}
    \begin{subfigure}[t]{0.3\textwidth}
    \includegraphics[width=\textwidth]{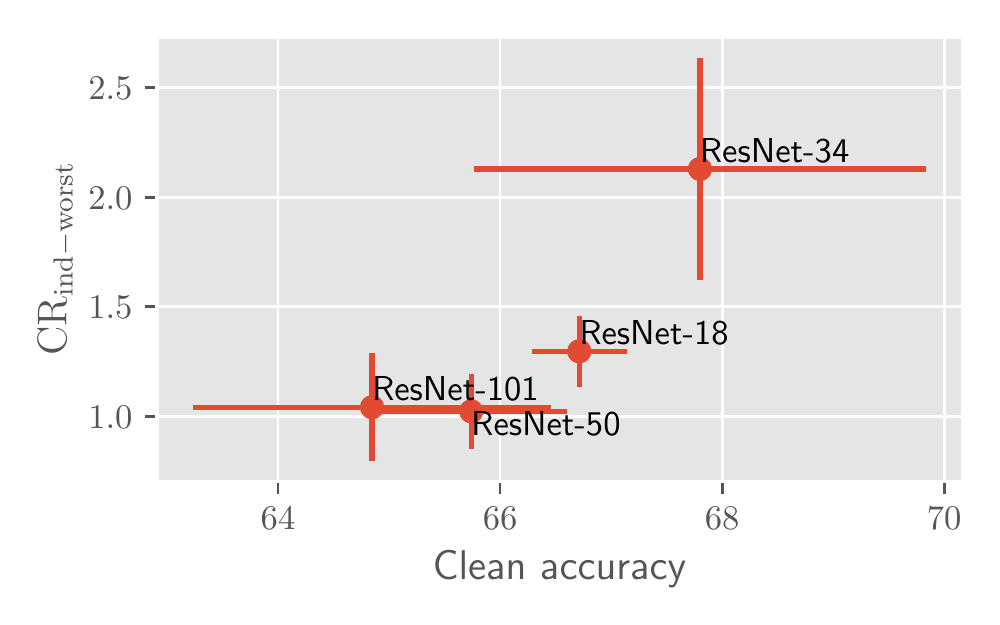}
        \caption{Elastic}
    \label{subfig:arch_per_atk_elastic}
    \end{subfigure}
    \caption{\noindent Impact of architecture size on $\text{CR}_{\text{ind-worst}}$ per attack type for models trained using LPIPS threat model with $\epsilon=0.5$ (via FastLPA training \citep{laidlaw2020perceptual})}

    \label{fig:per_atk_arch_impact}
\end{figure*}

In Figure \ref{fig:per_atk_arch_impact}, we plot the impact of architecture size on $\text{CR}_{\text{ind-worst}}$ for each attack type.  For StAdv and ReColor attacks, we find that $\text{CR}_{\text{ind-worst}}$ seems inversely correlated with accuracy and larger architectures tend to have higher $\text{CR}_{\text{ind-worst}}$ for those threat models.  For all other threat models, we observe that smaller architectures have better performance, which matches our observations in Section \ref{sec:design_choice_impact}.

\subsubsection{Impact of number of training epochs }
\label{app:lpips_per_attack_epochs}

\begin{figure*}[ht]
    \centering
     
    \begin{subfigure}[t]{0.3\textwidth}
    \includegraphics[width=\textwidth]{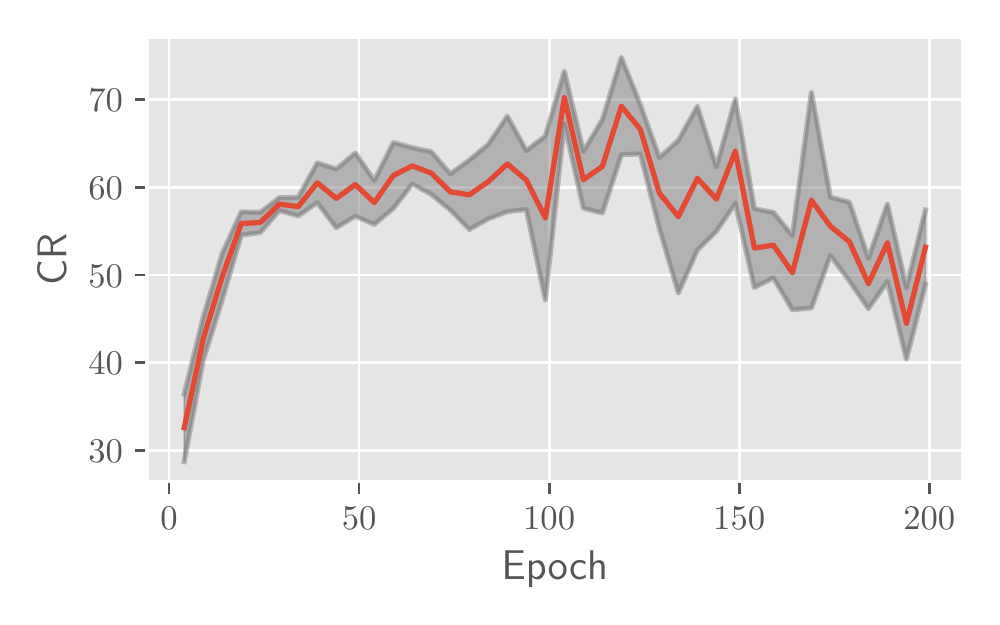}
        \caption{LPIPS}
    \label{subfig:epochs_per_atk_lpips}
    \end{subfigure}
    \begin{subfigure}[t]{0.3\textwidth}
    \includegraphics[width=\textwidth]{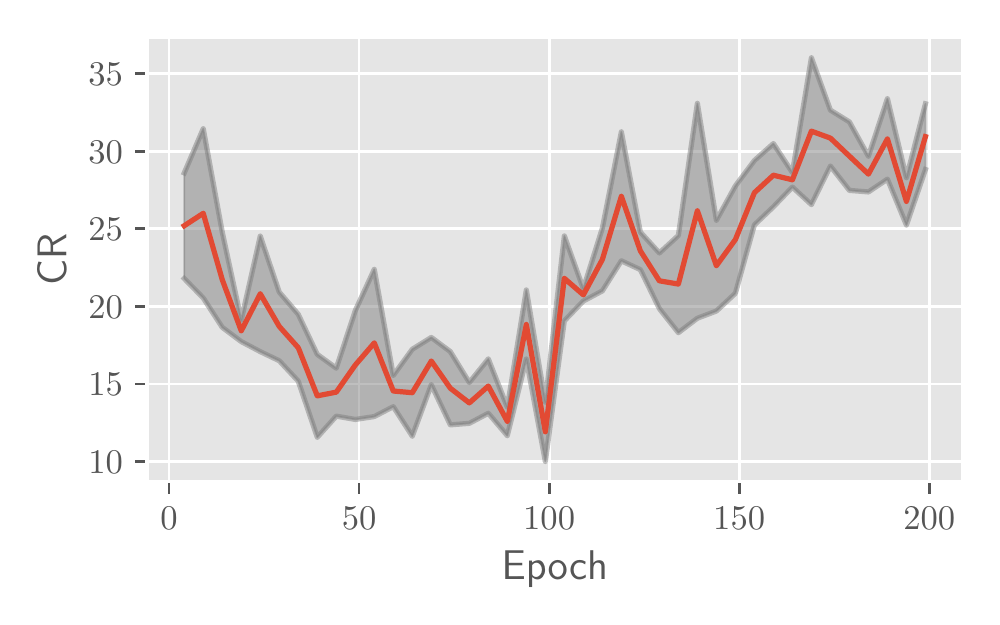}
        \caption{StAdv}
    \label{subfig:epochs_per_atk_stadv}
    \end{subfigure}
    \begin{subfigure}[t]{0.3\textwidth}
    \includegraphics[width=\textwidth]{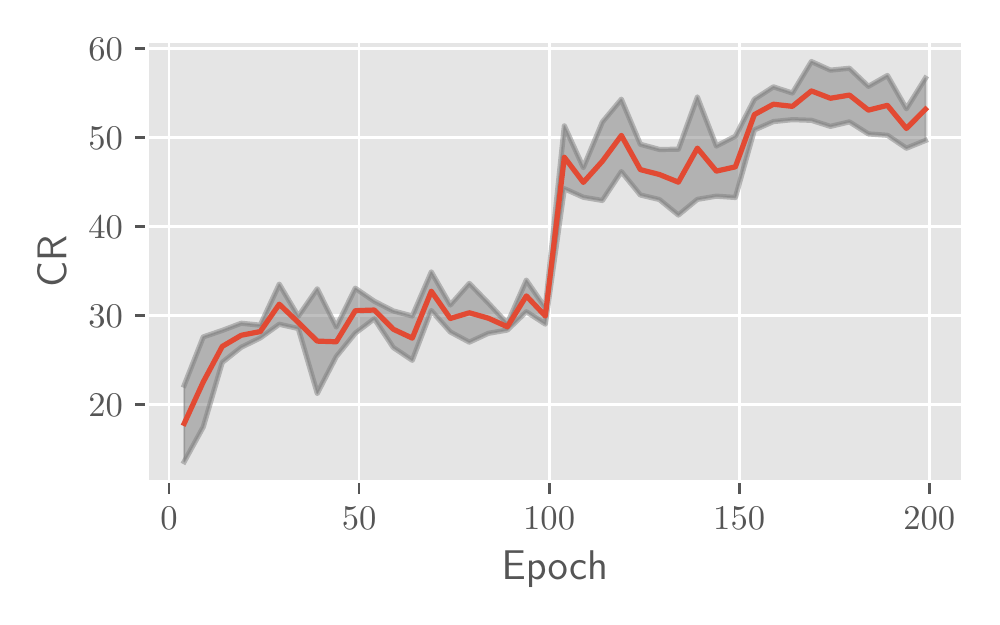}
    \caption{ReColor}
    \label{subfig:epochs_per_atk_recolor}
    \end{subfigure}
    
    \begin{subfigure}[t]{0.3\textwidth}
    \includegraphics[width=\textwidth]{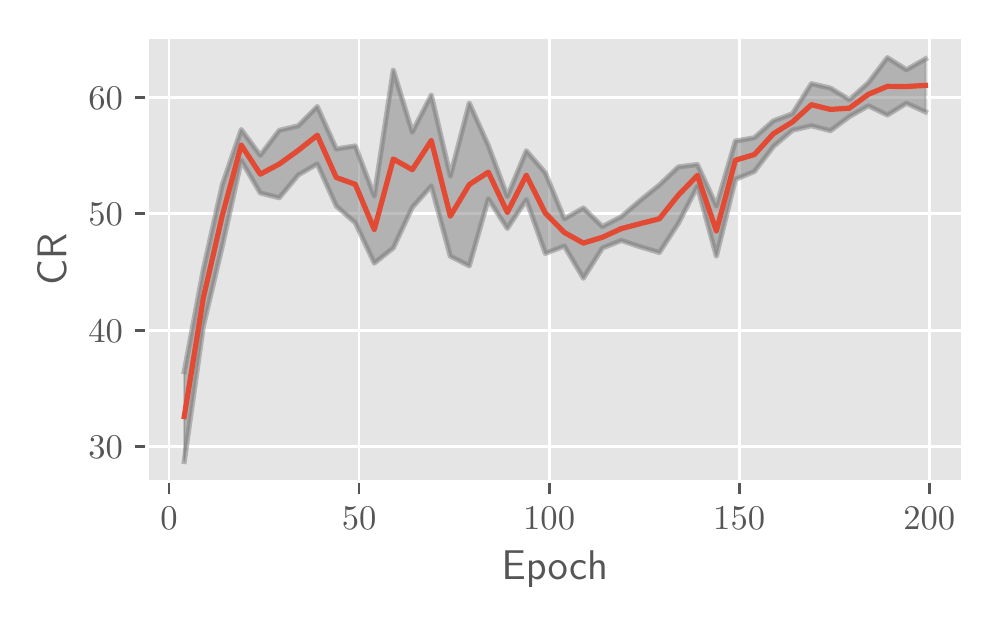}
    \caption{$\ell_1$}
    \label{subfig:epochs_per_atk_l1}
    \end{subfigure}
    \begin{subfigure}[t]{0.3\textwidth}
    \includegraphics[width=\textwidth]{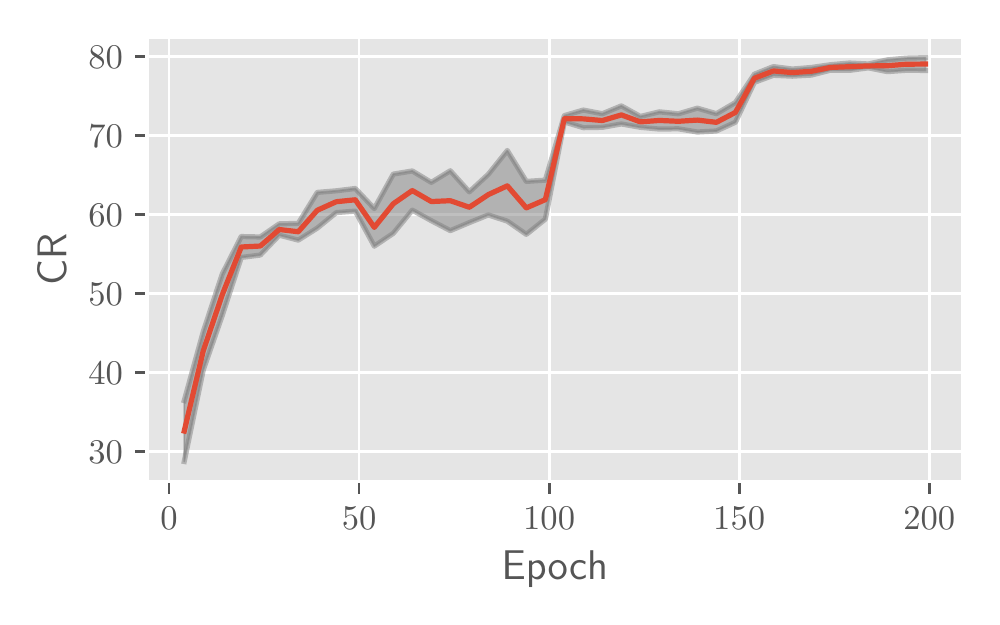}
    \caption{$\ell_2$}
        \label{subfig:epochs_per_atk_l2}
    \end{subfigure}
    \begin{subfigure}[t]{0.3\textwidth}
    \includegraphics[width=\textwidth]{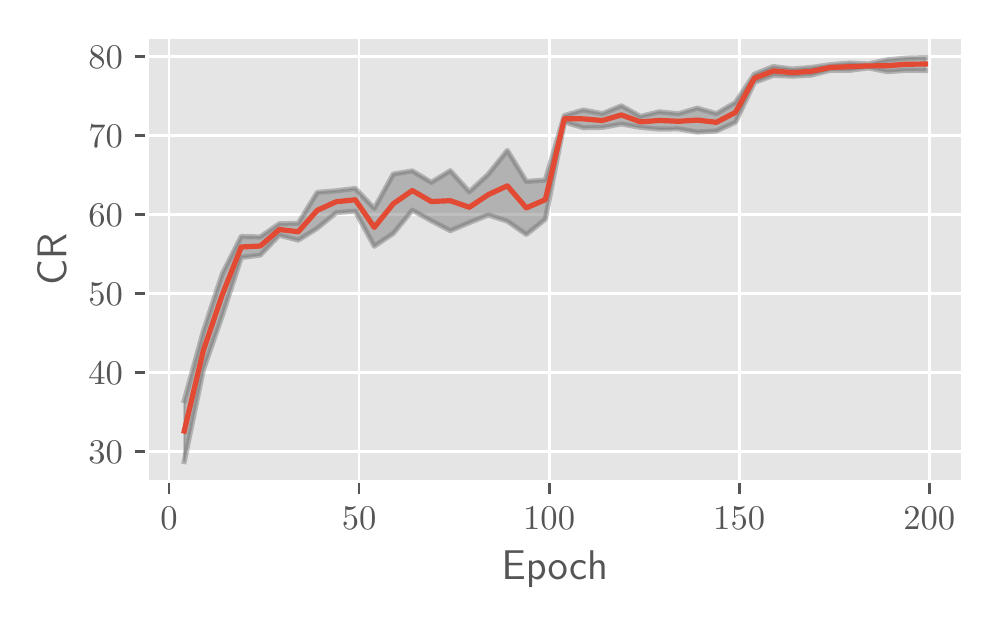}
    \caption{$\ell_\infty$}
    \label{subfig:epochs_per_atk_linf}
    \end{subfigure}

    \begin{subfigure}[t]{0.3\textwidth}
    \includegraphics[width=\textwidth]{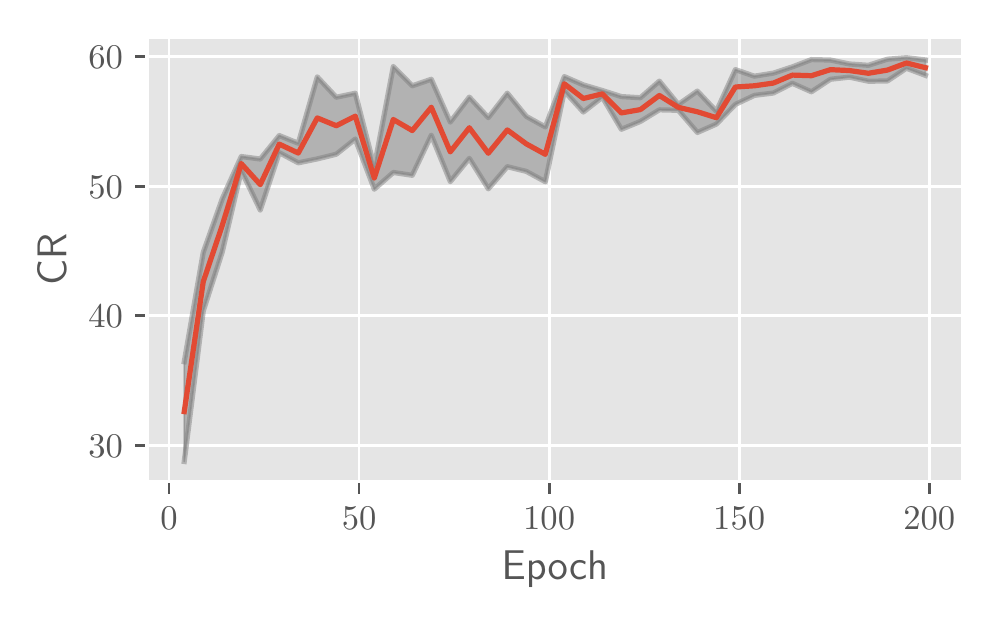}
        \caption{$\ell_1$-JPEG}

    \label{subfig:epochs_per_atk_l1_jpeg}
    \end{subfigure}
    \begin{subfigure}[t]{0.3\textwidth}
    \includegraphics[width=\textwidth]{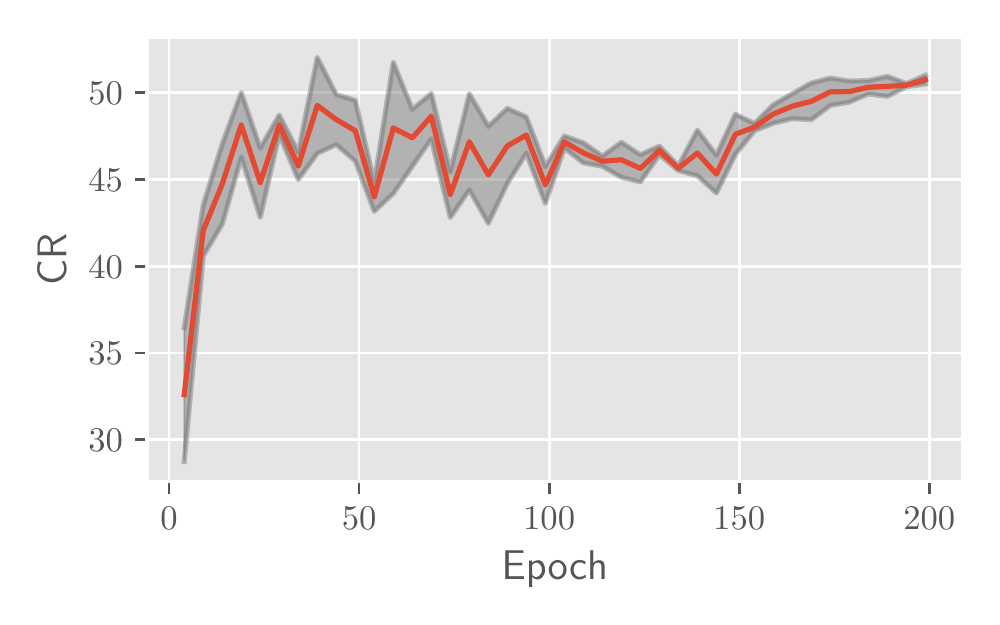}
        \caption{$\ell_\infty$-JPEG}
        \label{subfig:epochs_per_atk_linf_jpeg}
    \end{subfigure}
    \begin{subfigure}[t]{0.3\textwidth}
    \includegraphics[width=\textwidth]{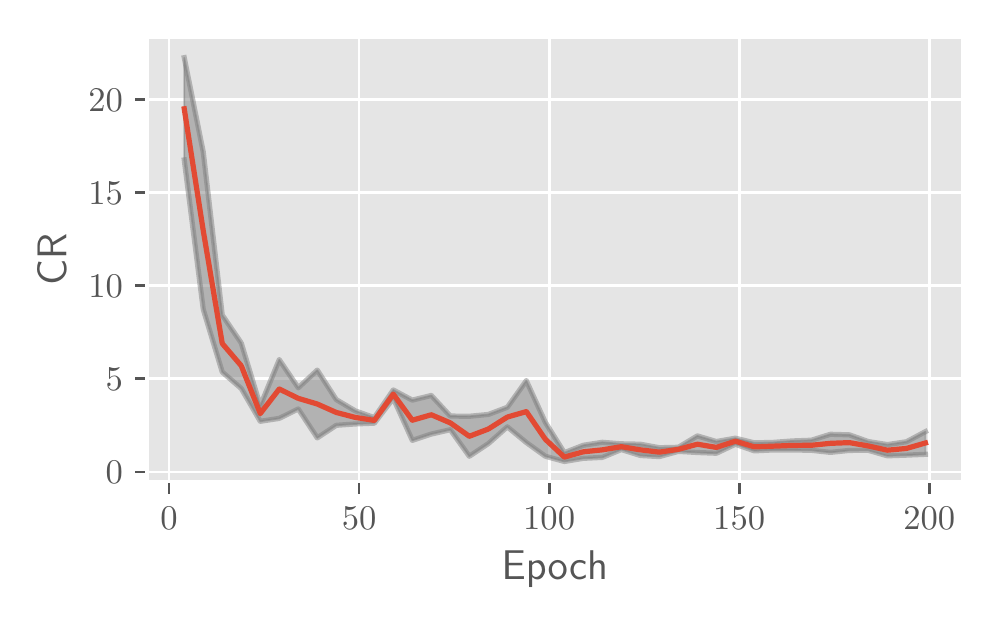}
        \caption{Elastic}
    \label{subfig:epochs_per_atk_elastic}
    \end{subfigure}
    \caption{\noindent Impact of number of training epochs on $\text{CR}_{\text{ind-worst}}$ per attack type for models trained using LPIPS threat model with $\epsilon=0.5$ (via FastLPA training \citep{laidlaw2020perceptual})}

    \label{fig:per_atk_epoch_impact}
\end{figure*}

In Figure \ref{fig:per_atk_epoch_impact}, we plot the impact of number of training epochs on $\text{CR}_{\text{ind-worst}}$ per attack.  We find that for LPIPS threat model (which is used during training), after about 100 epochs, additional training decreases CR on LPIPS attacks.  This suggests that after 100 epochs, the model starts to overfit on the training dataset.  For other threat models (except elastic attack), we find that CR is generally the highest at the last epoch of trining.  For elastic attacks, we find that CR drops during training.

\begin{figure*}[ht]
    \centering
     
    \begin{subfigure}[t]{0.3\textwidth}
    \includegraphics[width=\textwidth]{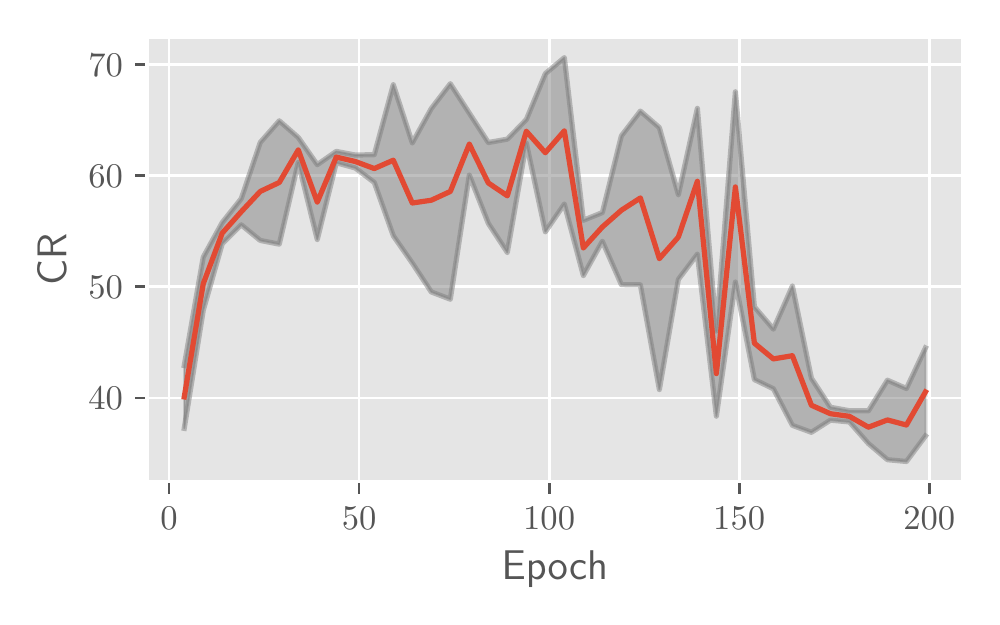}
        \caption{LPIPS}
    \label{subfig:el_epochs_per_atk_lpips}
    \end{subfigure}
    \begin{subfigure}[t]{0.3\textwidth}
    \includegraphics[width=\textwidth]{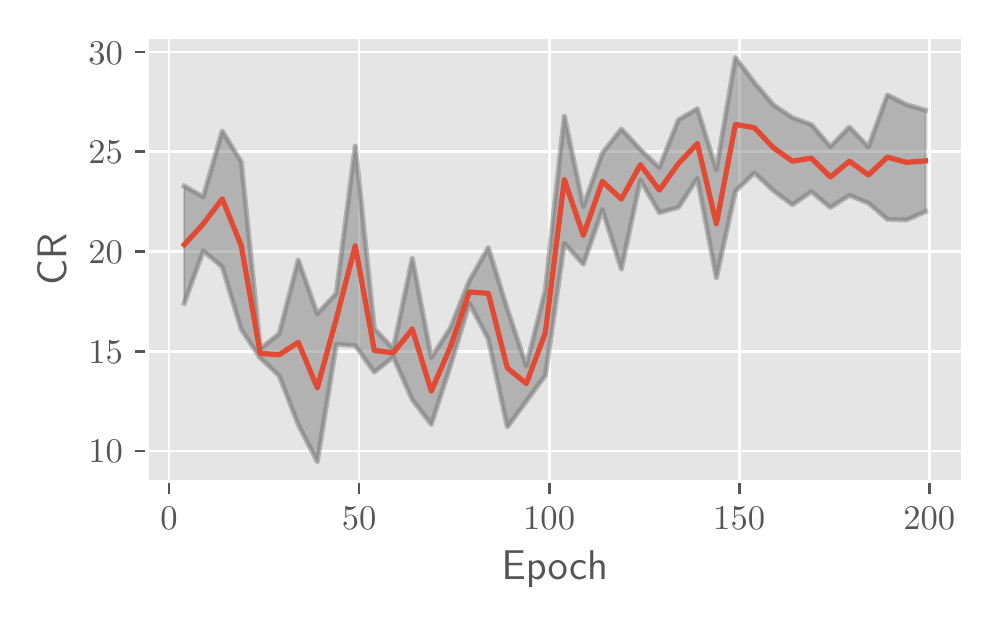}
        \caption{StAdv}
    \label{subfig:el_epochs_per_atk_stadv}
    \end{subfigure}
    \begin{subfigure}[t]{0.3\textwidth}
    \includegraphics[width=\textwidth]{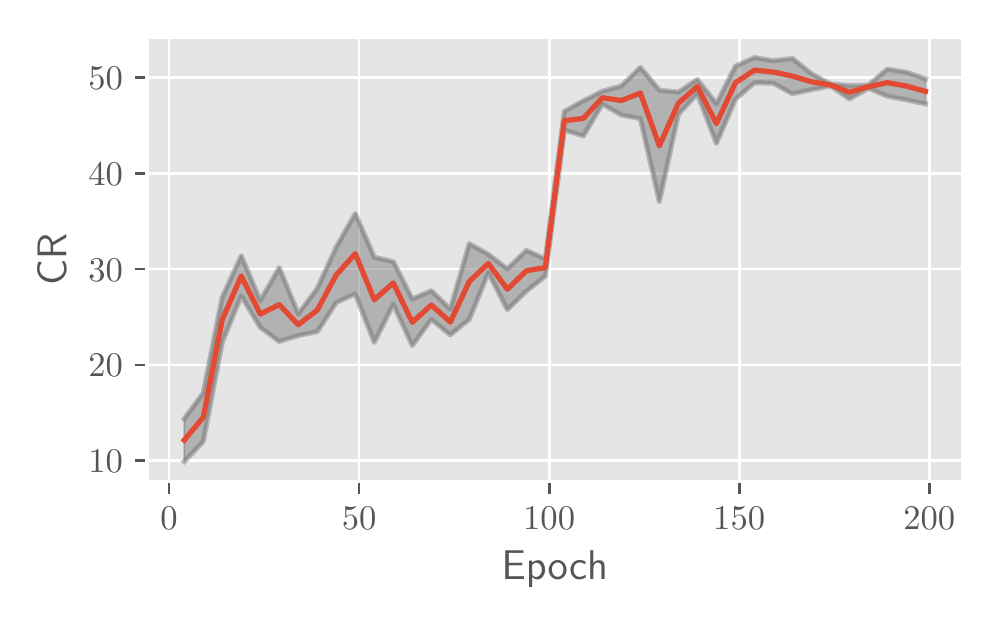}
    \caption{ReColor}
    \label{subfig:el_epochs_per_atk_recolor}
    \end{subfigure}
    
    \begin{subfigure}[t]{0.3\textwidth}
    \includegraphics[width=\textwidth]{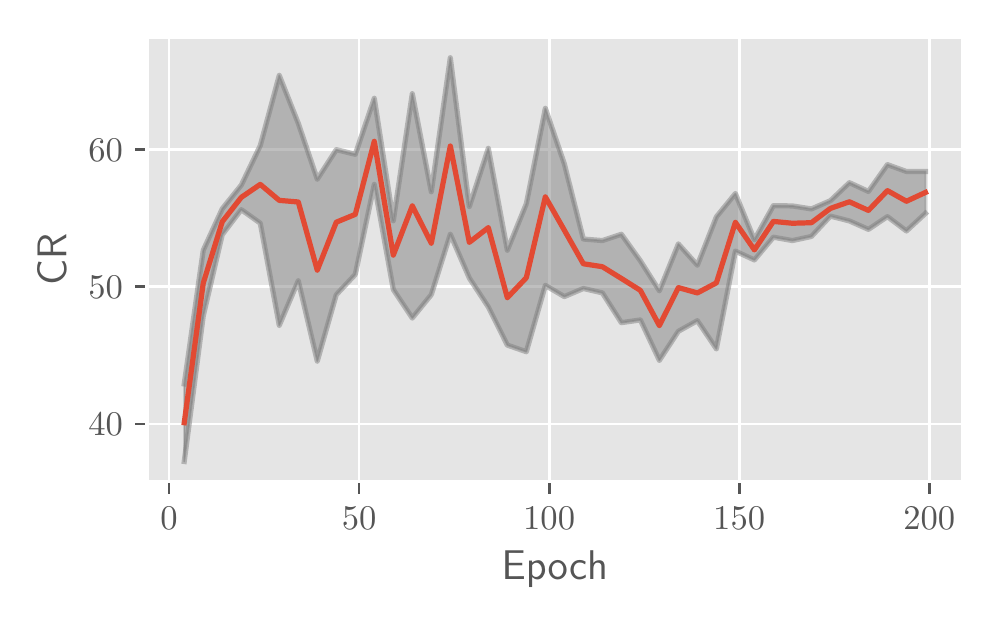}
    \caption{$\ell_1$}
    \label{subfig:el_epochs_per_atk_l1}
    \end{subfigure}
    \begin{subfigure}[t]{0.3\textwidth}
    \includegraphics[width=\textwidth]{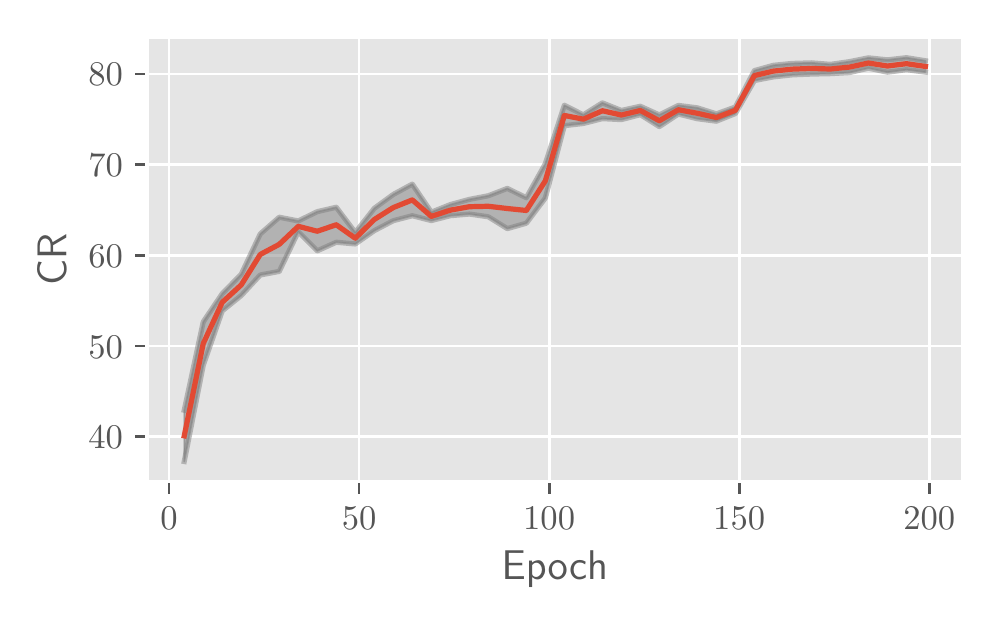}
    \caption{$\ell_2$}
        \label{subfig:el_epochs_per_atk_l2}
    \end{subfigure}
    \begin{subfigure}[t]{0.3\textwidth}
    \includegraphics[width=\textwidth]{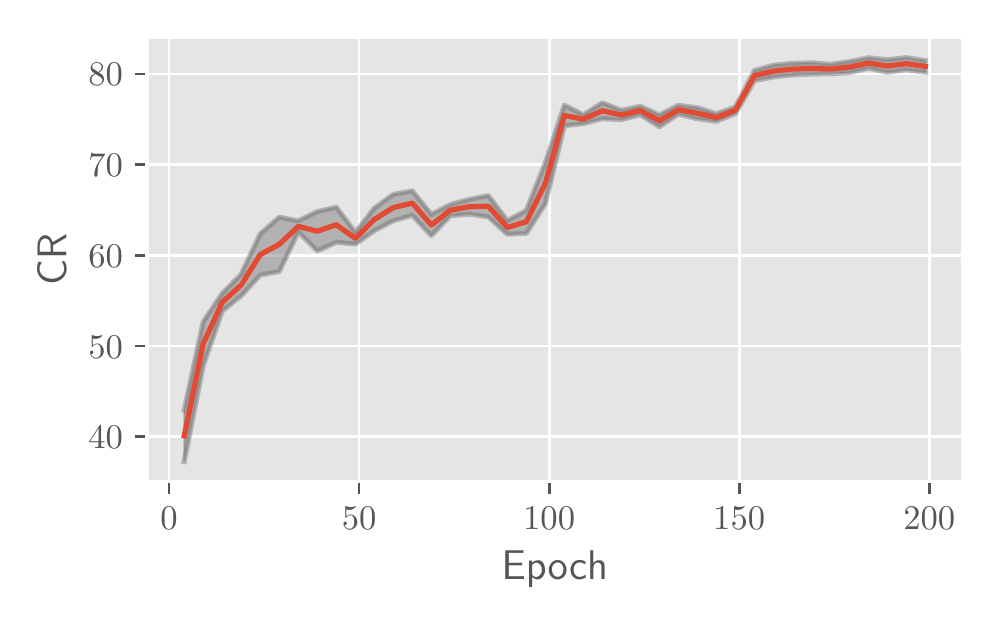}
    \caption{$\ell_\infty$}
    \label{subfig:el_epochs_per_atk_linf}
    \end{subfigure}

    \begin{subfigure}[t]{0.3\textwidth}
    \includegraphics[width=\textwidth]{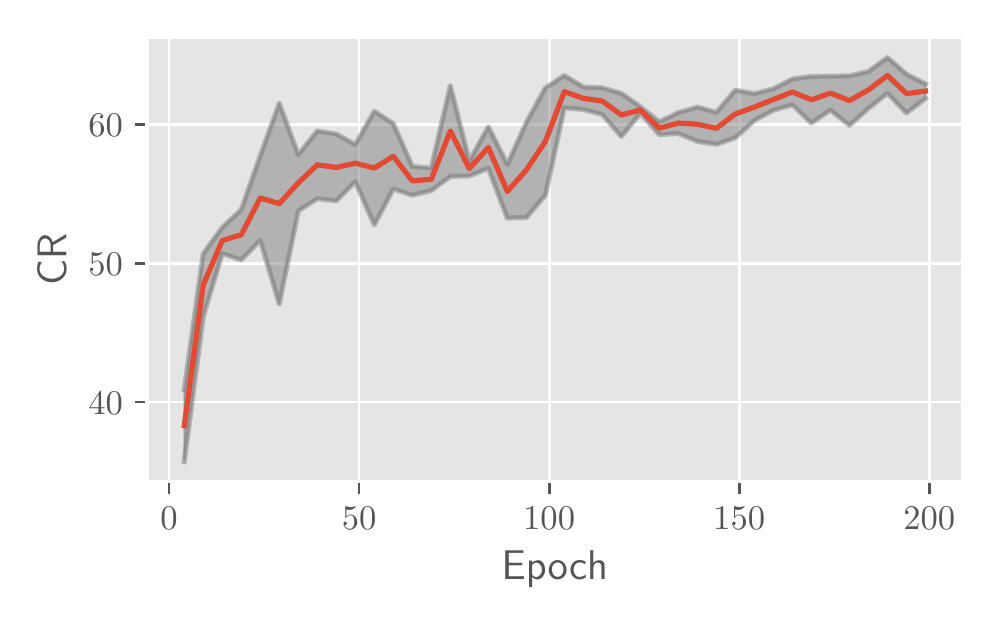}
        \caption{$\ell_1$-JPEG}

    \label{subfig:el_epochs_per_atk_l1_jpeg}
    \end{subfigure}
    \begin{subfigure}[t]{0.3\textwidth}
    \includegraphics[width=\textwidth]{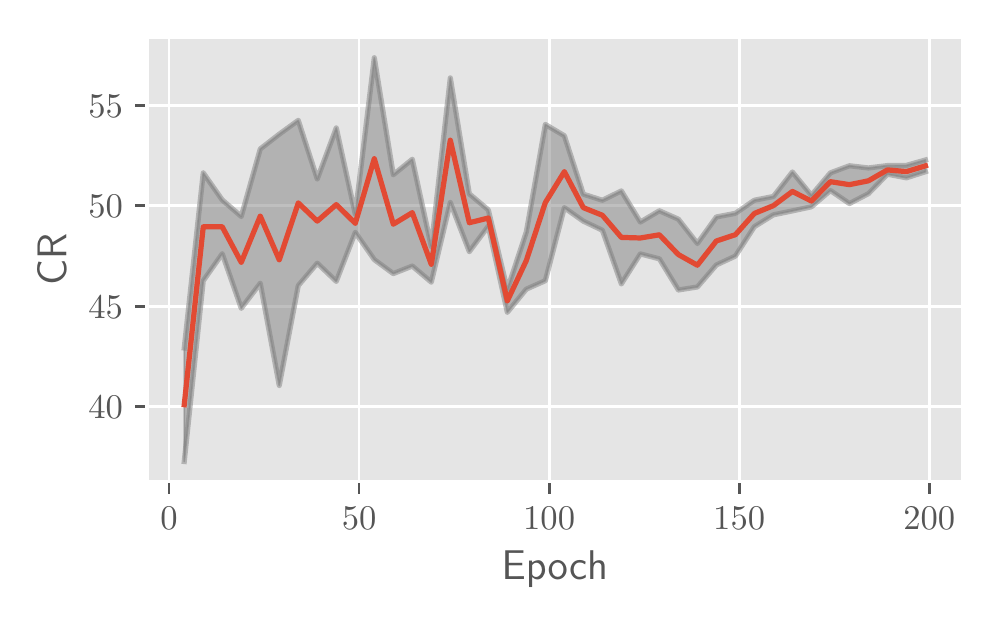}
        \caption{$\ell_\infty$-JPEG}
        \label{subfig:el_epochs_per_atk_linf_jpeg}
    \end{subfigure}
    \begin{subfigure}[t]{0.3\textwidth}
    \includegraphics[width=\textwidth]{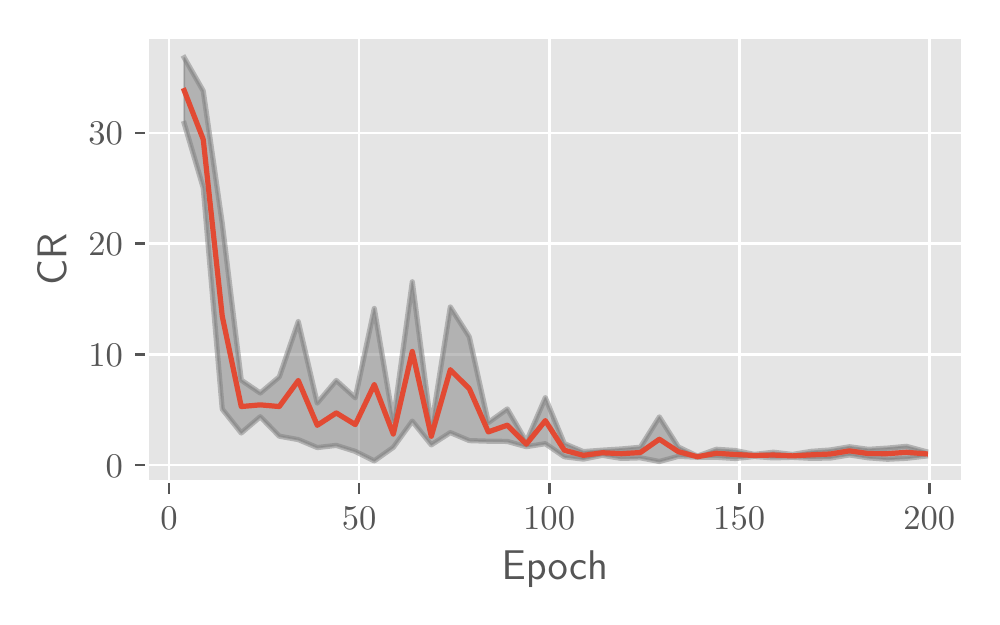}
        \caption{Elastic}
    \label{subfig:el_epochs_per_atk_elastic}
    \end{subfigure}
    \caption{\noindent Impact of number of training epochs on $\text{CR}_{\text{ind-worst}}$ per attack type for models trained with LPIPS threat model at $\epsilon=0.5$ and elastic threat model at $\epsilon=0.7$}

    \label{fig:el_per_atk_epoch_impact}
\end{figure*}

To see if incorporating elastic attacks into training changes the observed trend, we also incorporate elastic attacks with $\epsilon=0.7$ into training (along with the original LPIPS threat model).  We train using the maximum of elastic and LPIPS losses and provide plots in Figure \ref{fig:el_per_atk_epoch_impact}.  Interestingly, we do not observe much difference in training curves between including elastic into training and not including elastic in training.

\subsubsection{Impact of extra data}
\label{app:lpips_per_attack_data}
\begin{figure*}[ht]
    \centering
     
    \begin{subfigure}[t]{0.3\textwidth}
    \includegraphics[width=\textwidth]{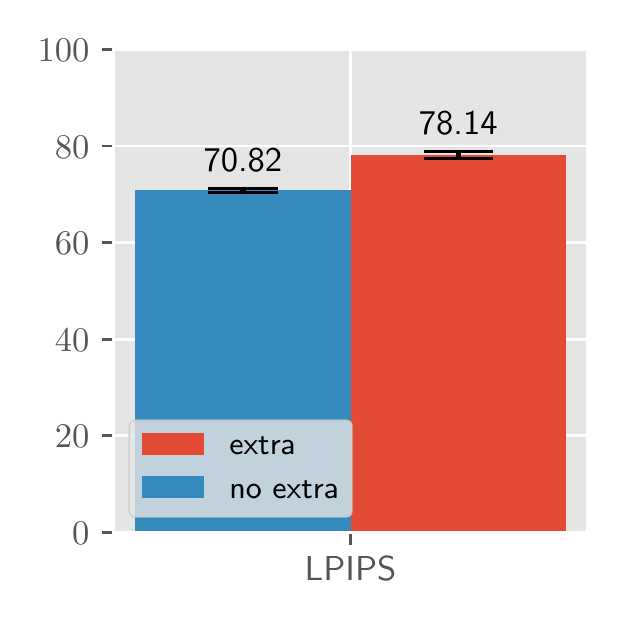}
    \label{subfig:extra_per_atk_lpips}
    \end{subfigure}
    \begin{subfigure}[t]{0.3\textwidth}
    \includegraphics[width=\textwidth]{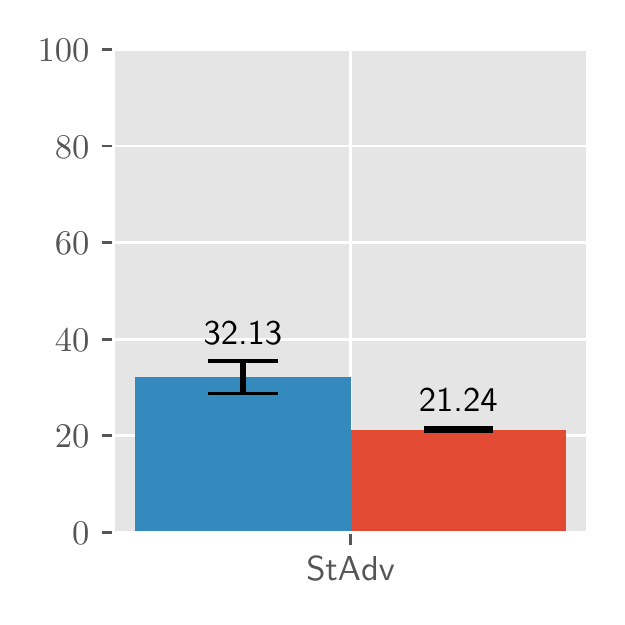}
        \label{subfig:extra_per_atk_stadv}
    \end{subfigure}
    \begin{subfigure}[t]{0.3\textwidth}
    \includegraphics[width=\textwidth]{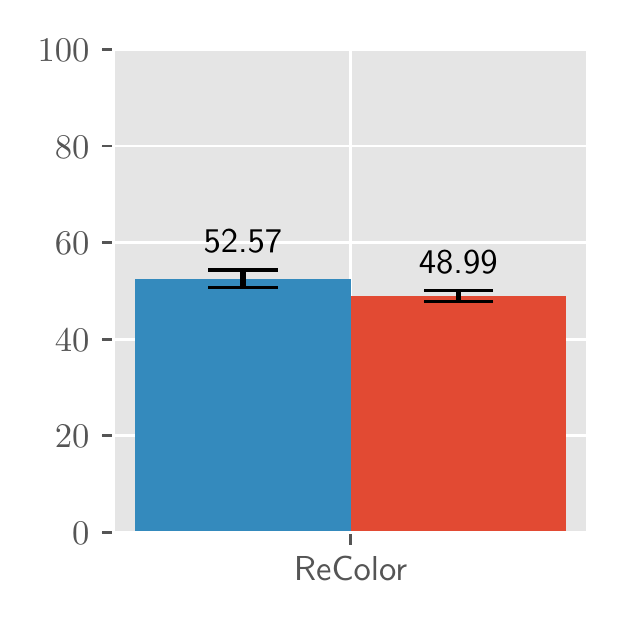}
    \label{subfig:extra_per_atk_recolor}
    \end{subfigure}
    
    \begin{subfigure}[t]{0.3\textwidth}
    \includegraphics[width=\textwidth]{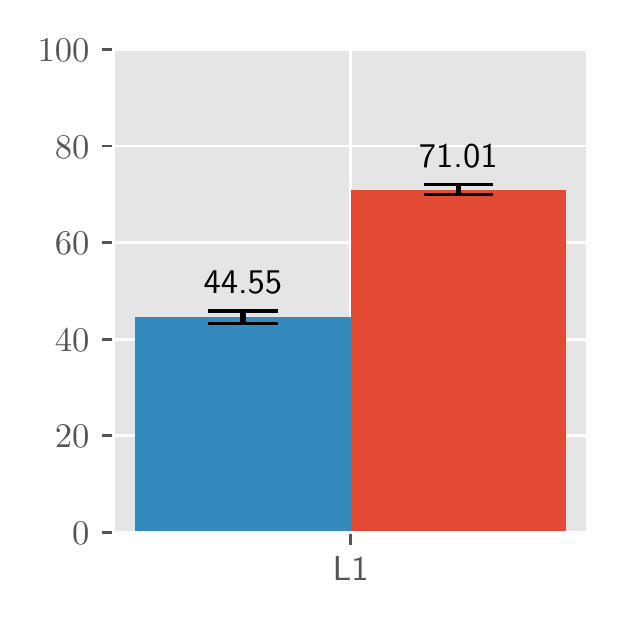}
    \label{subfig:extra_per_atk_l1}
    \end{subfigure}
    \begin{subfigure}[t]{0.3\textwidth}
    \includegraphics[width=\textwidth]{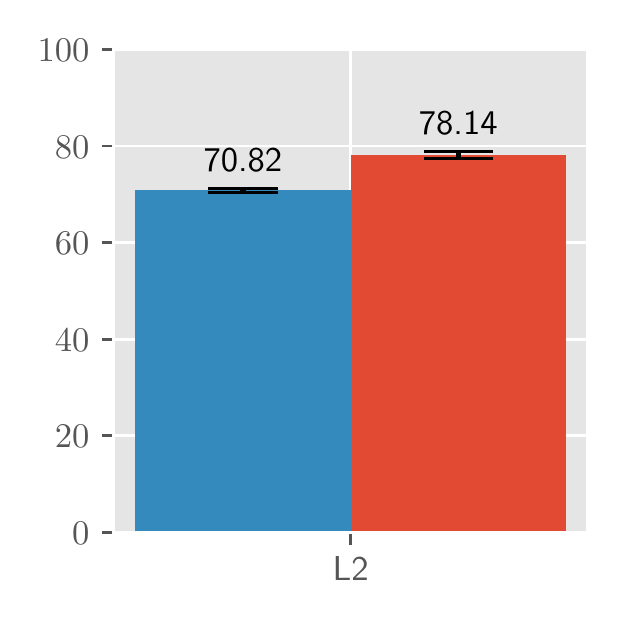}
        \label{subfig:extra_per_atk_l2}
    \end{subfigure}
    \begin{subfigure}[t]{0.3\textwidth}
    \includegraphics[width=\textwidth]{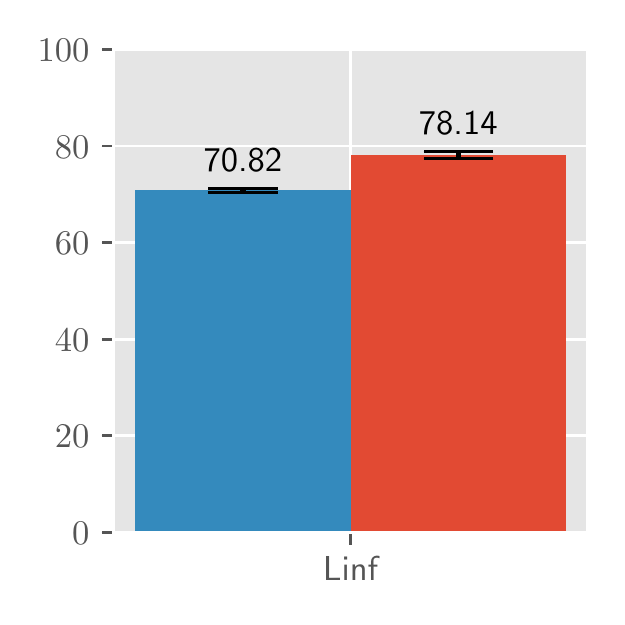}
    \label{subfig:extra_per_atk_linf}
    \end{subfigure}

    \begin{subfigure}[t]{0.3\textwidth}
    \includegraphics[width=\textwidth]{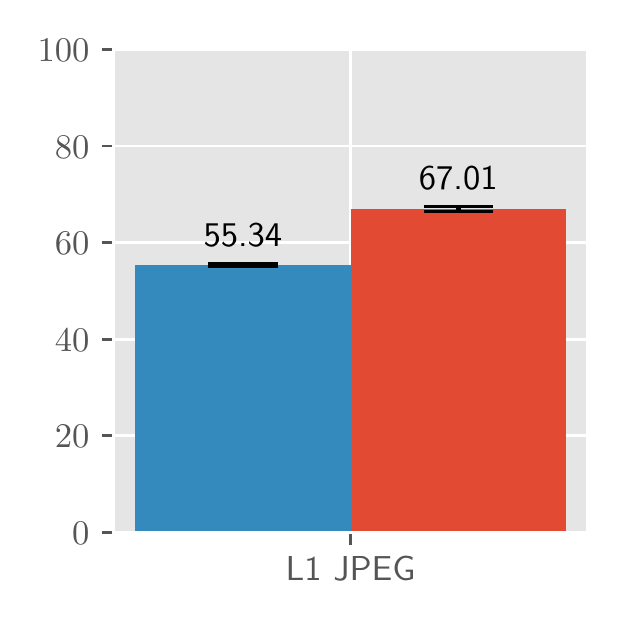}
    \label{subfig:extra_per_atk_l1_jpeg}
    \end{subfigure}
    \begin{subfigure}[t]{0.3\textwidth}
    \includegraphics[width=\textwidth]{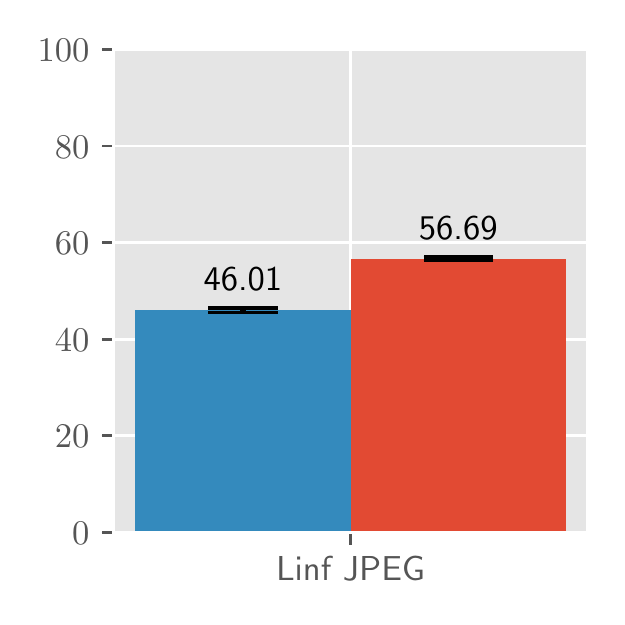}
        \label{subfig:extra_per_atk_linf_jpeg}
    \end{subfigure}
    \begin{subfigure}[t]{0.3\textwidth}
    \includegraphics[width=\textwidth]{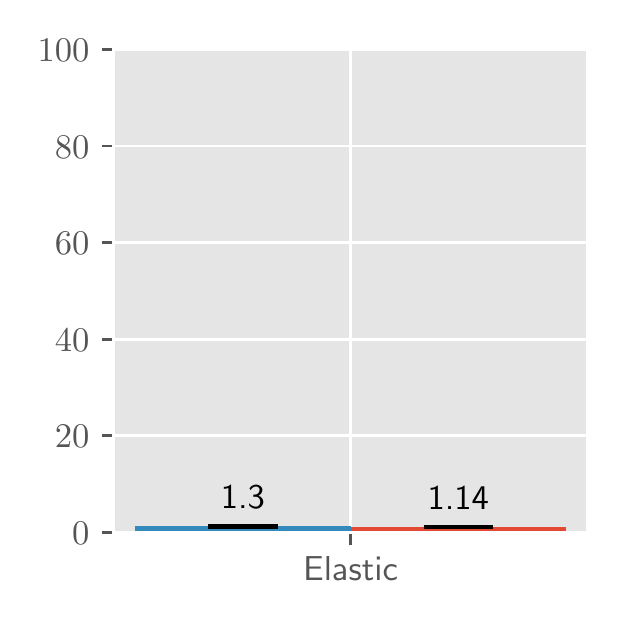}
    \label{subfig:extra_per_atk_elastic}
    \end{subfigure}
    \caption{\noindent Impact of extra training data on $\text{CR}_{\text{ind-worst}}$ per attack type for models trained using LPIPS threat model with $\epsilon=0.5$ (via FastLPA training \citep{laidlaw2020perceptual})}
    \label{fig:per_atk_extra_impact}
\end{figure*}

In Figure \ref{fig:per_atk_extra_impact}, we plot the impact of extra training data on $\text{CR}_{\text{ind-worst}}$ per attack.  We observe that for most attacks (LPIPS, $\ell_1$, $\ell_2$, $\ell_{\infty}$, $\ell_1$ JPEG and $\ell_\infty$ JPEG), extra data improves $\text{CR}_{\text{ind-worst}}$.  However, for some attacks, Specifically, StAdv, ReColor, and Elastic attacks extra data does not improve  $\text{CR}_{\text{ind-worst}}$.  In fact, for StAdv and ReColor, the drop in performance after incorporating extra data is significant.  Additionally, we find that the aggregate $\text{CR}_{\text{ind-worst}}$ trend for extra data observed in Section \ref{sec:design_choice_impact} is dominated by elastic attack performance.

\subsection{Additional results for adversarially trained models}
In this section, we present results for training with $\ell_{\infty}$, $\ell_2$ threat models, and the union of $\ell_1$, $\ell_2$, and $\ell_{\infty}$ attacks via stochastic adversarial training \citep{madaan2020learning} analogous to those present for LPIPS threat model in Section \ref{sec:design_choice_impact}.

\subsubsection{Analysis of models trained with $\ell_{\infty}$ source threat model} 
\label{app:adv_train_analysis}

\begin{figure*}[ht]
    \centering
     
    \begin{subfigure}[t]{0.3\textwidth}
    \includegraphics[width=\textwidth]{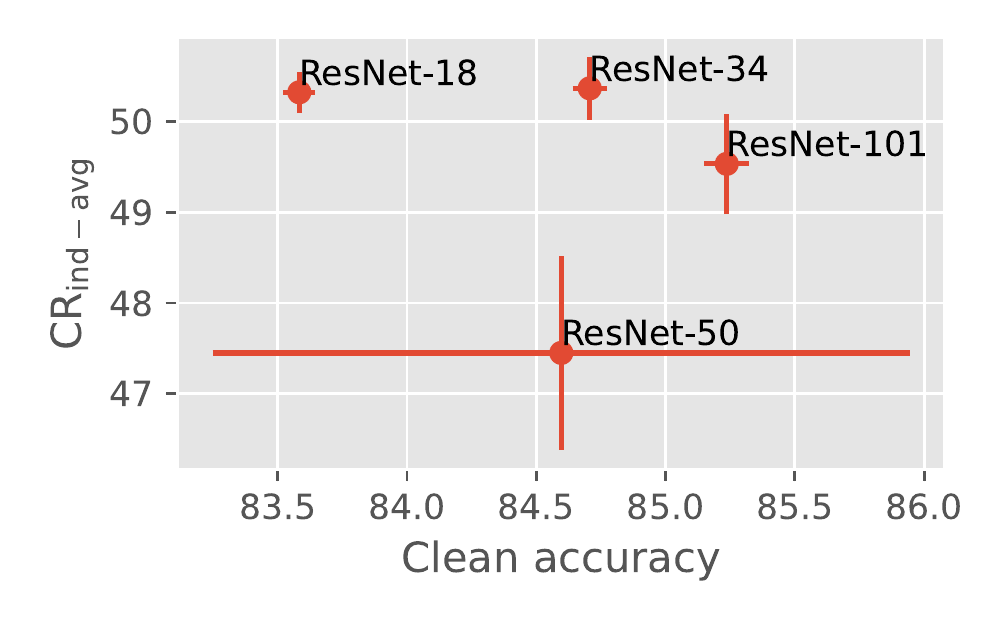}
    \caption{$\text{CR}_{\text{ind-avg}}$}
    \label{subfig:linf_arch_avg}
    \end{subfigure}
    \begin{subfigure}[t]{0.3\textwidth}
    \includegraphics[width=\textwidth]{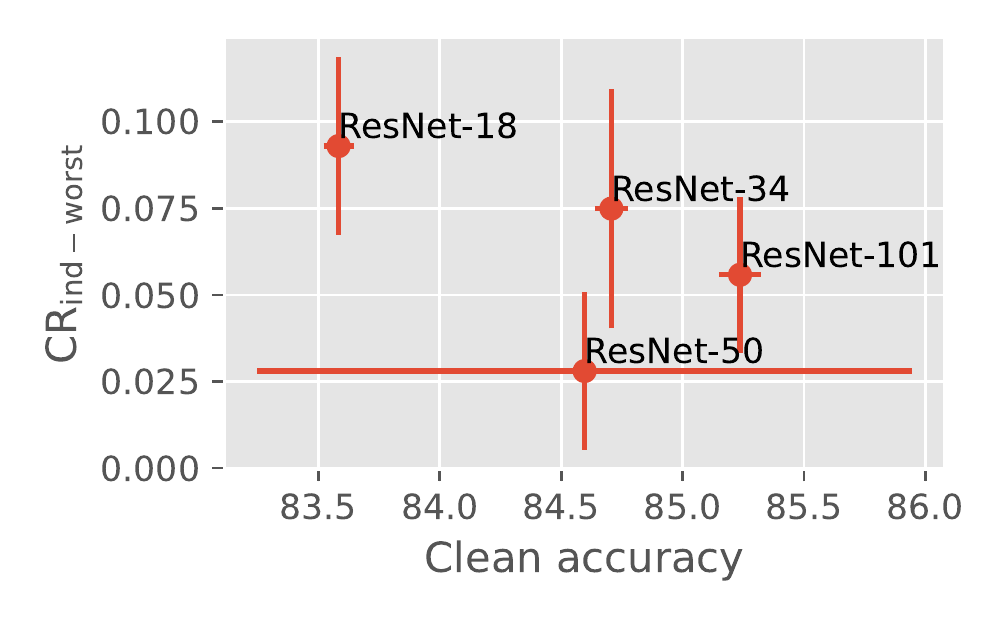}
    \caption{$\text{CR}_{\text{ind-worst}}$}
        \label{subfig:linf_arch_worst}
    \end{subfigure}
    \begin{subfigure}[t]{0.3\textwidth}
    \includegraphics[width=\textwidth]{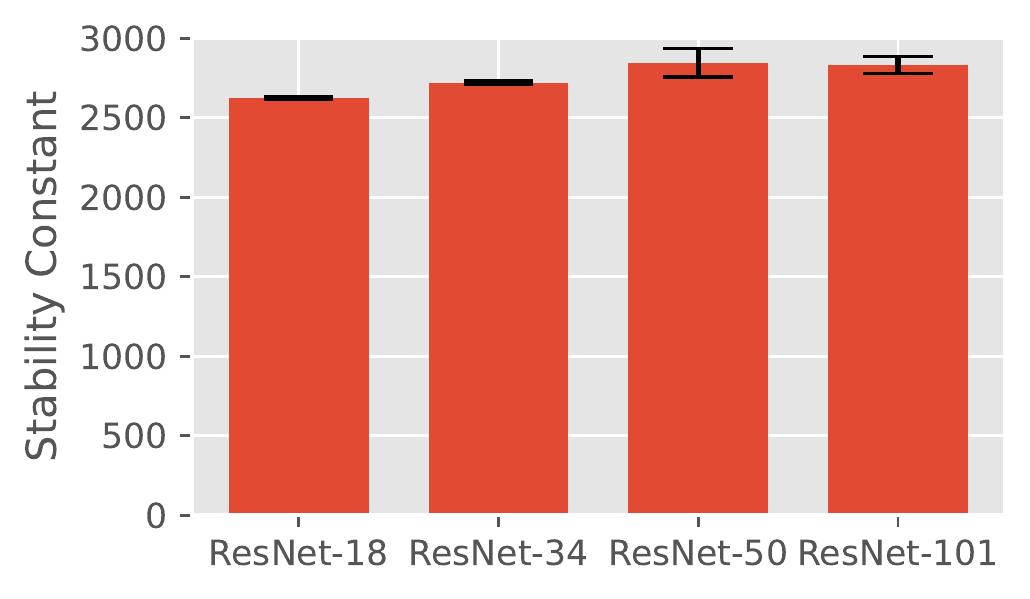}
    \caption{Stability Constant}
    \label{subfig:linf_arch_stab}
    \end{subfigure}
    \caption{\noindent \textbf{Impact of architecture size.} Figures (a) and (b): Clean accuracy vs CR for models trained  using PGD adversarial training with $\ell_{\infty}$ threat model with radius $\frac{8}{255}$.  Results are averaged over 3 trials and error bars are shown.  Higher values of CR indicate better performance. Figure (c): SC computed for models of each architecture. Lower SC indicates better performance.}
    \label{fig:linf_arch_impact}
\end{figure*}

\textbf{Impact of architecture size } In figure \ref{fig:linf_arch_impact}, we plot the performance of ResNet-18, ResNet-34, ResNet-50, and ResNet-101 architectures in terms of CR, clean accuracy, and stability constant.  From Figures \ref{subfig:linf_arch_avg} and \ref{subfig:linf_arch_worst}, we note that while larger ResNet architectures (in particular ResNet-101) is able to achieve much higher clean accuracy, smaller architectures (ResNet-18 and ResNet-34) are able to achieve higher CR score, suggesting that smaller architectures are more optimal for multiattack robustness.  Similarily, we find that these smaller architectures have smaller stability constant in Figure \ref{subfig:linf_arch_stab}, suggesting that there is less of a performance drop when shifting to unseen attacks compared to larger architectures.  This trend matches what was observed for LPIPS threat model in Section \ref{sec:design_choice_impact}.

\begin{figure*}[ht]
    \centering
     
    \begin{subfigure}[t]{0.34\textwidth}
    \includegraphics[width=\textwidth]{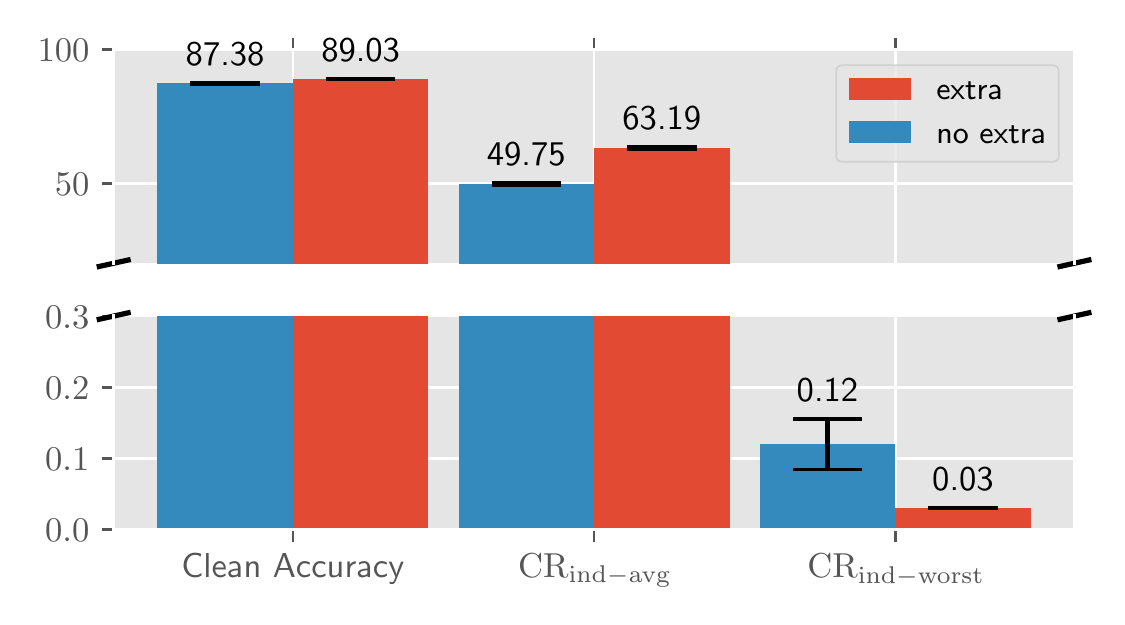}
    \caption{Clean Accuracy and CR}
    \label{subfig:linf_extra_avg}
    \label{subfig:linf_extra_worst}
    \end{subfigure}
    \begin{subfigure}[t]{0.3\textwidth}
    \includegraphics[width=\textwidth]{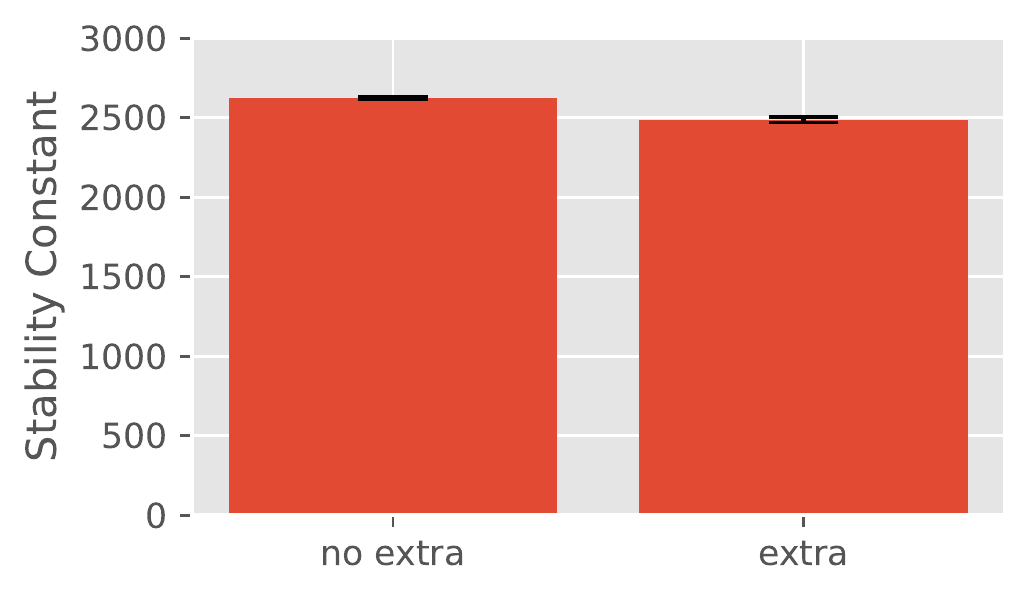}
    \caption{Stability Constant}
    \label{subfig:linf_extra_stab}
    \end{subfigure}
    \caption{\noindent \textbf{Impact of additional training data.} Figure (a): Clean accuracy and CR for ResNet-18 models trained using PGD adversarial training using PGD adversarial training with $\ell_{\infty}$ threat model with radius $\frac{8}{255}$.  Higher CR indicates better performance.  Results are averaged over 3 trials and error bars are shown. Figure (c): SC computed for models with and without additional training data. Lower SC indicates better performance.}
    \label{fig:linf_extra_impact}
\end{figure*}

\textbf{Impact of additional training data }In Figure \ref{fig:linf_extra_impact}, we plot CR, clean accuracy, and stability for ResNet-18 models trained with and without additional (synthetic) training data.  Similar to findings from Section \ref{sec:design_choice_impact}, we find that additional data improves both clean accuracy and $\text{CR}_{\text{ind-avg}}$.  However, there is no significant change in performance in terms of $\text{CR}_{\text{ind-worst}}$.  This suggests, that while on average, extra data can improve performance across the set of tested attacks, this is not necessarily the case for worst-case performance.  We find that for $\ell_{\infty}$ training, extra data does improve stability across attacks (stability constant in Figure \ref{subfig:linf_arch_stab} significantly decreases with extra training data), suggesting that for $\ell_{\infty}$ training, using additional data can decrease the drop in performance to unforeseen attacks.

\begin{figure*}[ht]
    \centering
     
    \begin{subfigure}[t]{0.3\textwidth}
    \includegraphics[width=\textwidth]{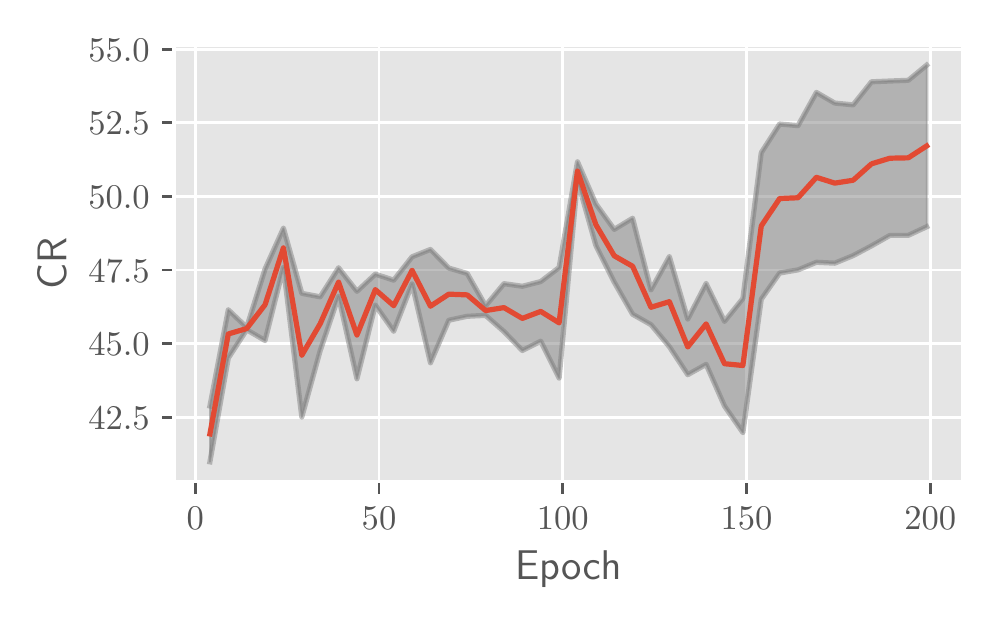}
    \caption{$\text{CR}_{\text{ind-avg}}$}
    \label{subfig:linf_es_avg}
    \end{subfigure}
    \begin{subfigure}[t]{0.3\textwidth}
    \includegraphics[width=\textwidth]{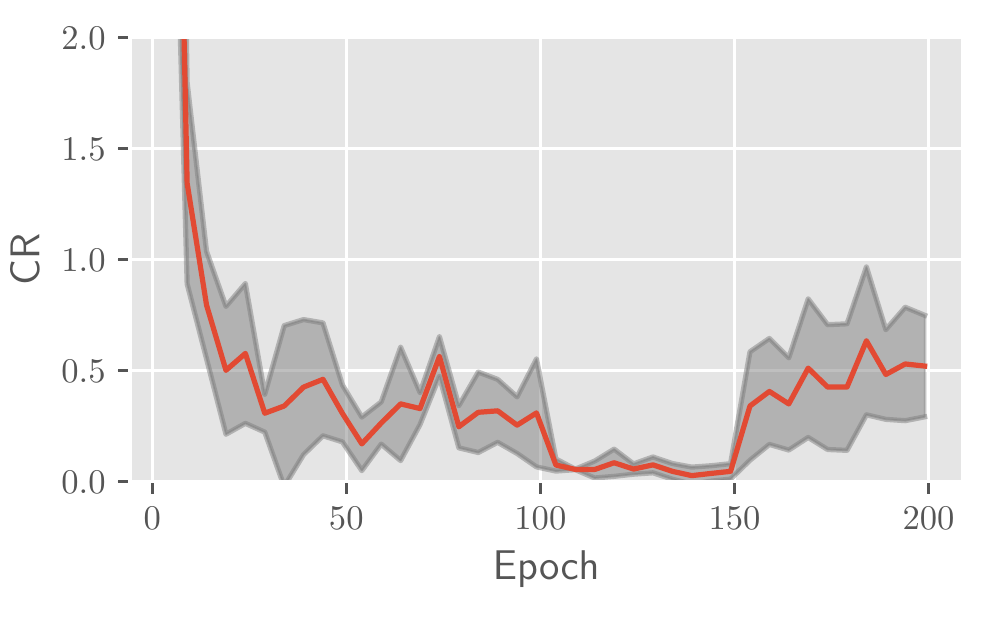}
    \caption{$\text{CR}_{\text{ind-worst}}$}
        \label{subfig:linf_es_worst}
    \end{subfigure}
    \begin{subfigure}[t]{0.3\textwidth}
    \includegraphics[width=\textwidth]{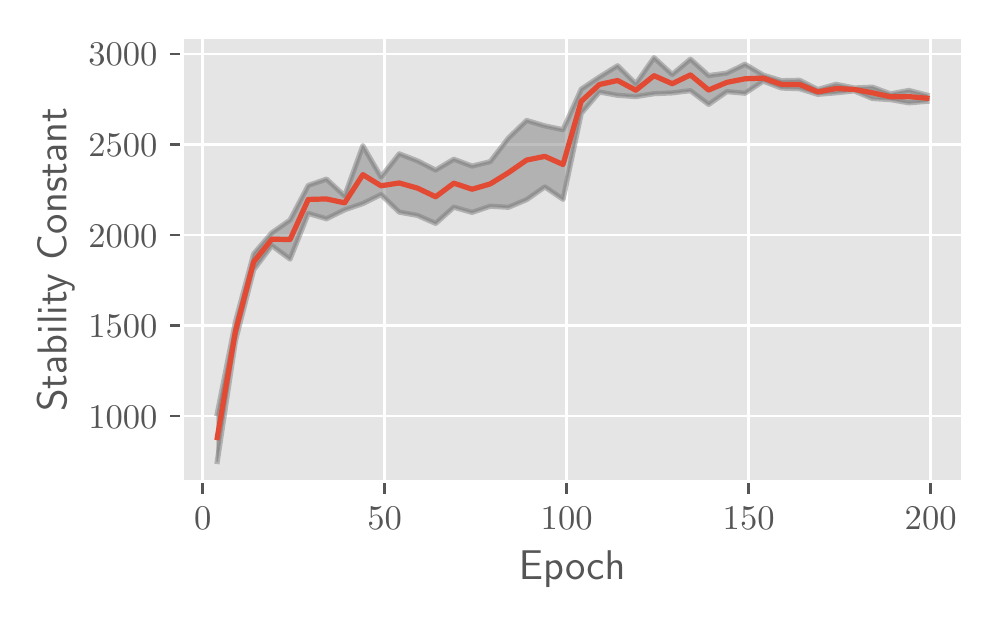}
    \caption{Stability Constant}
    \label{subfig:linf_es_stab}
    \end{subfigure}
    \caption{\noindent \textbf{Impact of number of training epochs.} CR and SC over epoch for models trained with $\ell_{\infty}$ threat model with radius $\frac{8}{255}$. The red line indicates the average over 3 runs while the grey band highlights indicate 1 standard deviation from the mean. Higher values of CR and lower values of SC indicate better performance.}
    \label{fig:linf_early_stop_impact}
\end{figure*}

\textbf{Impact of number of epochs } In Figure \ref{fig:linf_early_stop_impact}, we plot the impact of number of training epochs on CR and stability.  Similar to trends for training with LPIPS threat model in Section \ref{sec:design_choice_impact}, we find that longer training does improve average case performance.  For worst-case performance, we find that $\text{CR}_{\text{ind-worst}}$ drops quickly within the first 50 epochs of training and then stays relatively constant throughout the remainder of training.  This makes sense because at initialization the model is essentially randomly guessing so even on the worst-case attack, the model can still achieve about 10\% robust accuracy.  However, as the model trains it becomes more vulnerable to the worst-case (and likely unseen attack) leading to a large drop in worst-case robust accuracy, which causes CR to be near 0.  Similar to training with LPIPS threat model, we also find that stability constant increases during training, which suggests that as training continues, the drop in robustness between seen and unseen threat models increases.

\subsubsection{Analysis of models trained with $\ell_{2}$ source threat model} 

\begin{figure*}[ht]
    \centering
     
    \begin{subfigure}[t]{0.3\textwidth}
    \includegraphics[width=\textwidth]{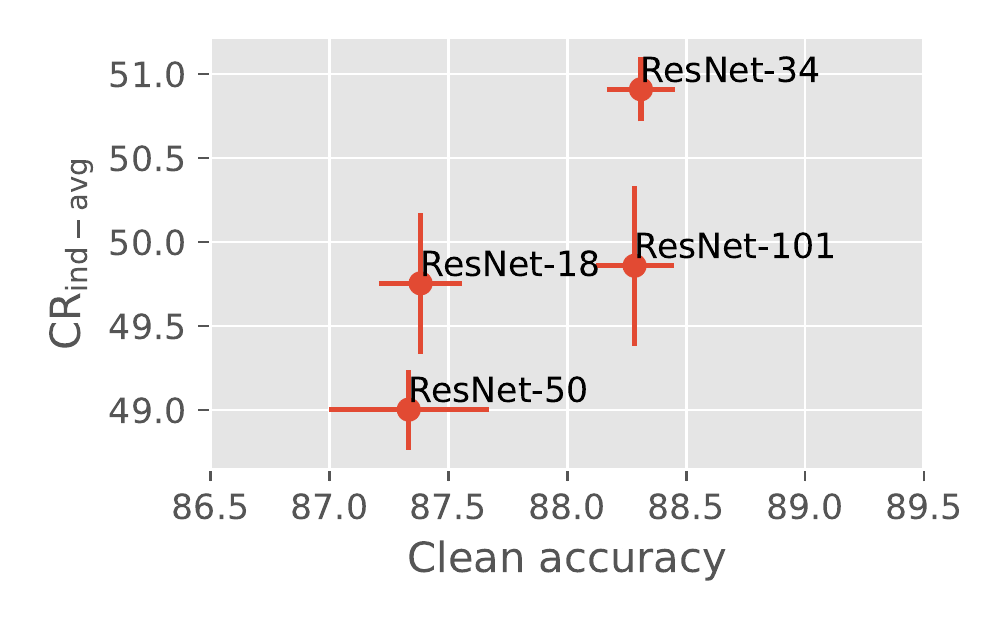}
    \caption{$\text{CR}_{\text{ind-avg}}$}
    \label{subfig:l2_arch_avg}
    \end{subfigure}
    \begin{subfigure}[t]{0.3\textwidth}
    \includegraphics[width=\textwidth]{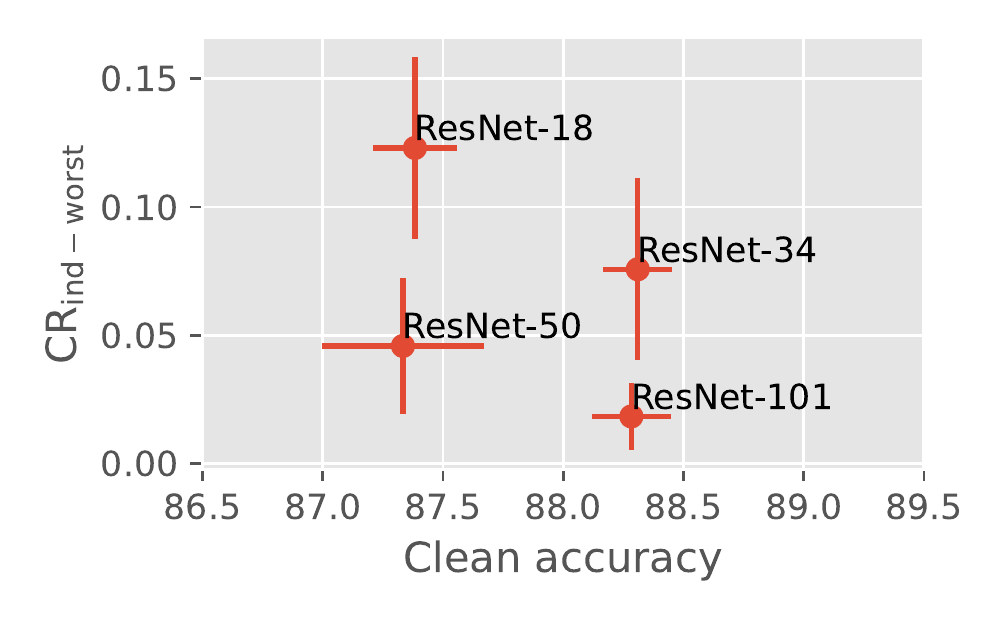}
    \caption{$\text{CR}_{\text{ind-worst}}$}
        \label{subfig:l2_arch_worst}
    \end{subfigure}
    \begin{subfigure}[t]{0.3\textwidth}
    \includegraphics[width=\textwidth]{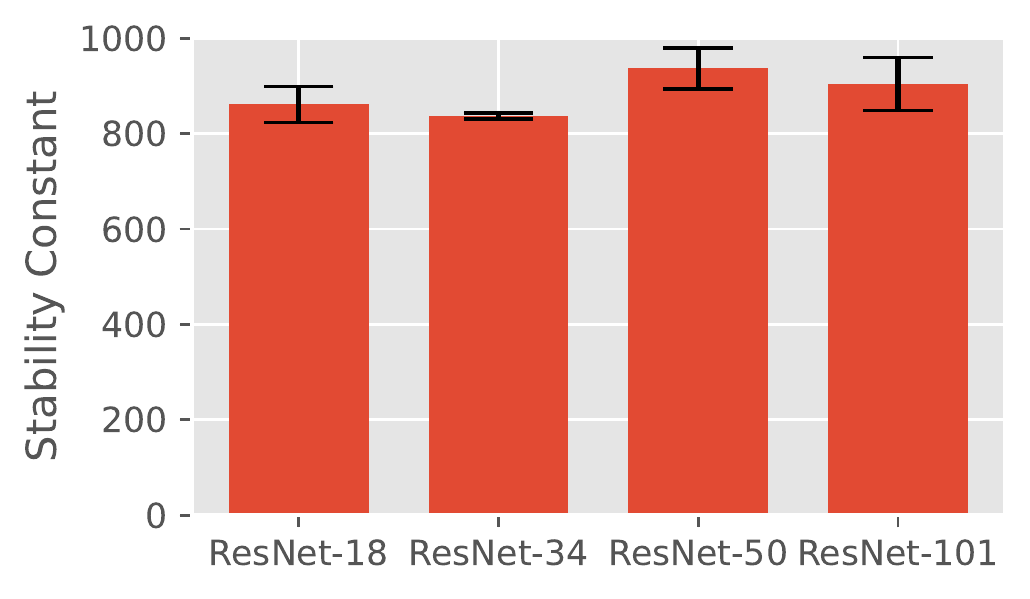}
    \caption{Stability Constant}
    \label{subfig:l2_arch_stab}
    \end{subfigure}
    \caption{\noindent \textbf{Impact of architecture size.} Figures (a) and (b): Clean accuracy vs CR for models trained  using PGD adversarial training with $\ell_{2}$ threat model with radius $0.5$. Results are averaged over 3 trials and error bars are shown.  Higher values of CR indicate better performance. Figure (c): SC computed for models of each architecture. Lower SC indicates better performance.}
    \label{fig:l2_arch_impact}
\end{figure*}

\textbf{Impact of architecture size }In Figure \ref{fig:l2_arch_impact}, we plot the CR, clean accuracy, and stability constant achieved by training ResNet-18, ResNet-34, ResNet-50, and ResNet-101.  Similar to trends for training with LPIPS and training with $\ell_{\infty}$ threat model, we find that smaller architectures (ResNet-18, ResNet-34) outperform larger models in terms of CR and stability constant, suggesting that smaller models are better when it comes to multiattack robustness.

\begin{figure*}[ht]
    \centering
     
    \begin{subfigure}[t]{0.34\textwidth}
    \includegraphics[width=\textwidth]{figs/extra_L2_both.pdf}
    \caption{Clean Accuracy and CR}
    \label{subfig:l2_extra_avg}
    \label{subfig:l2_extra_worst}
    \end{subfigure}
    \begin{subfigure}[t]{0.3\textwidth}
    \includegraphics[width=\textwidth]{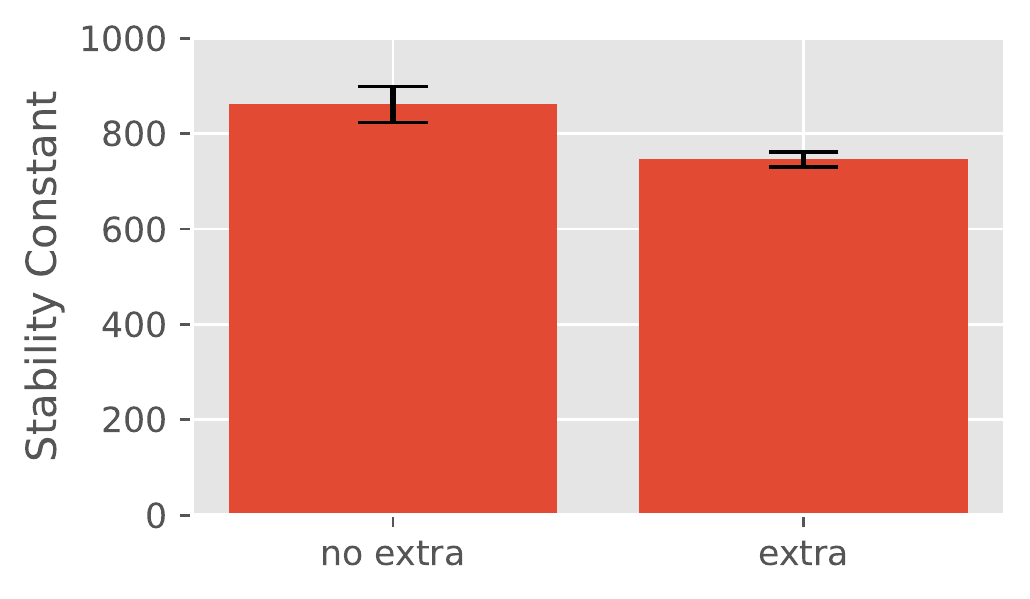}
    \caption{Stability Constant}
    \label{subfig:l2_extra_stab}
    \end{subfigure}
    \caption{
    \noindent \textbf{Impact of additional training data.} Figure (a): Clean accuracy and CR for ResNet-18 models trained using PGD adversarial training with $\ell_{2}$ threat model with radius $0.5$.  Higher CR indicates better performance.  Results are averaged over 3 trials and error bars are shown. Figure (c): SC computed for models with and without additional training data. Lower SC indicates better performance.}
    \label{fig:l2_extra_impact}
\end{figure*}

\textbf{Impact of additional training data} In Figure \ref{fig:l2_extra_impact}, we plot the clean accuracy, CR, and stability constant of ResNet-18 models trained with $\ell_2$ adversarial training.  Similar to what we observed for $\ell_{\infty}$ and LPIPS adversarial training, we find that extra data significantly improves $\text{CR}_{\text{ind-avg}}$, suggesting that extra data improves average robust performance over the set of attacks.  However, we find that using additional data harms worst-case multiattack robustness: in Figure \ref{subfig:l2_extra_worst}, we find that $\text{CR}_{\text{ind-worst}}$ decreases after including additional data during training.  This suggest that while on average robustness over the set of attacks increases with additional data, additional data does not uniformly improve performance over all attacks.  Observing stability constant in Figure \ref{subfig:l2_extra_stab}, we find that extra data helps decrease tability constant, suggesting that models trained with additional data exhibit less of a drop in robustness when evaluated on attacks outside of the $\ell_2$ threat model which have similar difficulty.

\begin{figure*}[ht]
    \centering
     
    \begin{subfigure}[t]{0.3\textwidth}
    \includegraphics[width=\textwidth]{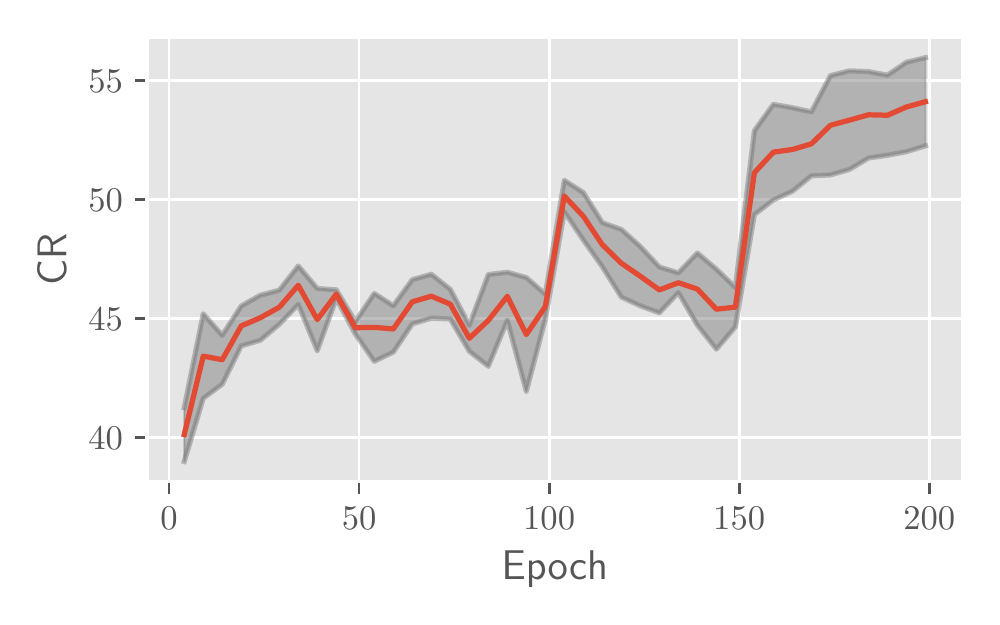}
    \caption{$\text{CR}_{\text{ind-avg}}$}
    \label{subfig:l2_es_avg}
    \end{subfigure}
    \begin{subfigure}[t]{0.3\textwidth}
    \includegraphics[width=\textwidth]{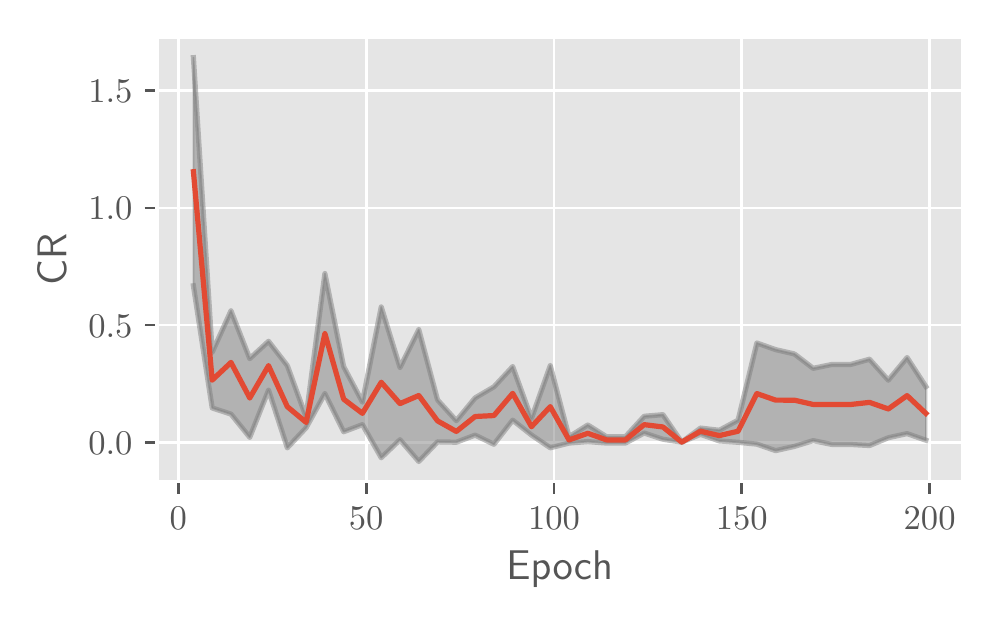}
    \caption{$\text{CR}_{\text{ind-worst}}$}
        \label{subfig:l2_es_worst}
    \end{subfigure}
    \begin{subfigure}[t]{0.3\textwidth}
    \includegraphics[width=\textwidth]{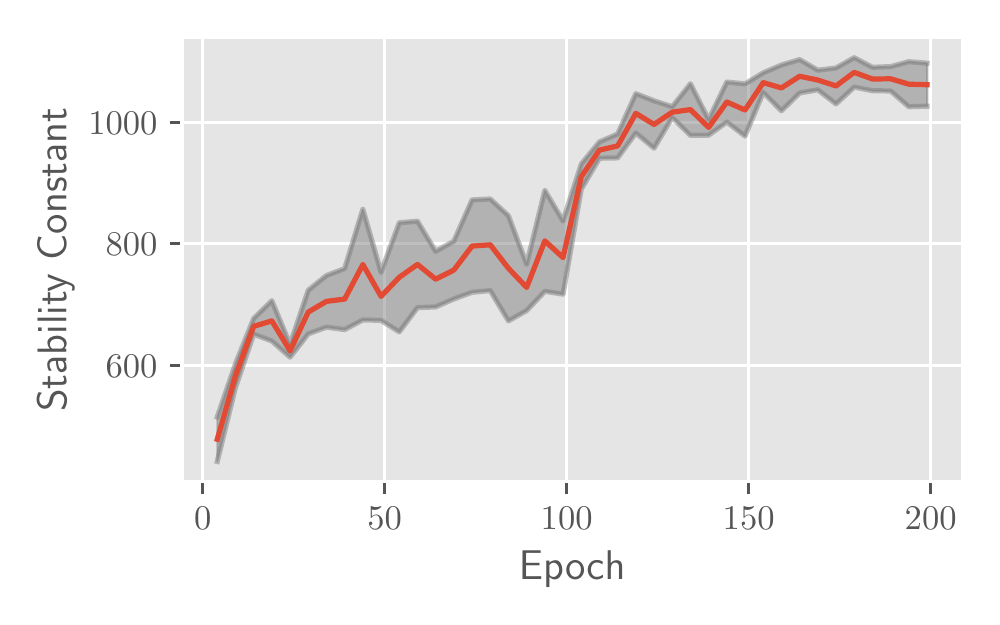}
    \caption{Stability Constant}
    \label{subfig:l2_es_stab}
    \end{subfigure}
    \caption{\noindent \textbf{Impact of number of training epochs.} CR and SC over epoch for models trained with $\ell_{2}$ threat model with radius $0.5$. The red line indicates the average over 3 runs while the grey band highlights indicate 1 standard deviation from the mean. Higher values of CR and lower values of SC indicate better performance.}
    \label{fig:l2_early_stop_impact}
\end{figure*}

\textbf{Impact of number of epochs }In Figure \ref{fig:l2_early_stop_impact}, we plot CR and stability constant over training epochs.  Similar to trends for LPIPS and $\ell_{\infty}$ threat models, we find that more training generally increases $\text{CR}_{\text{ind-avg}}$, suggesting better average case performance.  Additionally, we find that $\text{CR}_{\text{ind-worst}}$ drops quickly within the first 50 epochs of training and then remains generally constant throughout the remainder of training.  For stability, we find that stability constant gradually increases throughout training suggesting that as training progresses, the drop in performance across threat models increases.

\subsubsection{Analysis of models trained with $\ell_{1}$, $\ell_2$, and $\ell_{\infty}$ threat models}
In this section we report results for training on the union of $\ell_1$, $\ell_2$, and $\ell_{\infty}$ attacks via stochastic adversarial training (SAT) \citep{madaan2020learning}.  For these experiments, we use the same training setup as used in \citep{madaan2020learning} where for architecture size and additional data experiments we train for 30 epochs.

\begin{figure*}[ht]
    \centering
     
    \begin{subfigure}[t]{0.3\textwidth}
    \includegraphics[width=\textwidth]{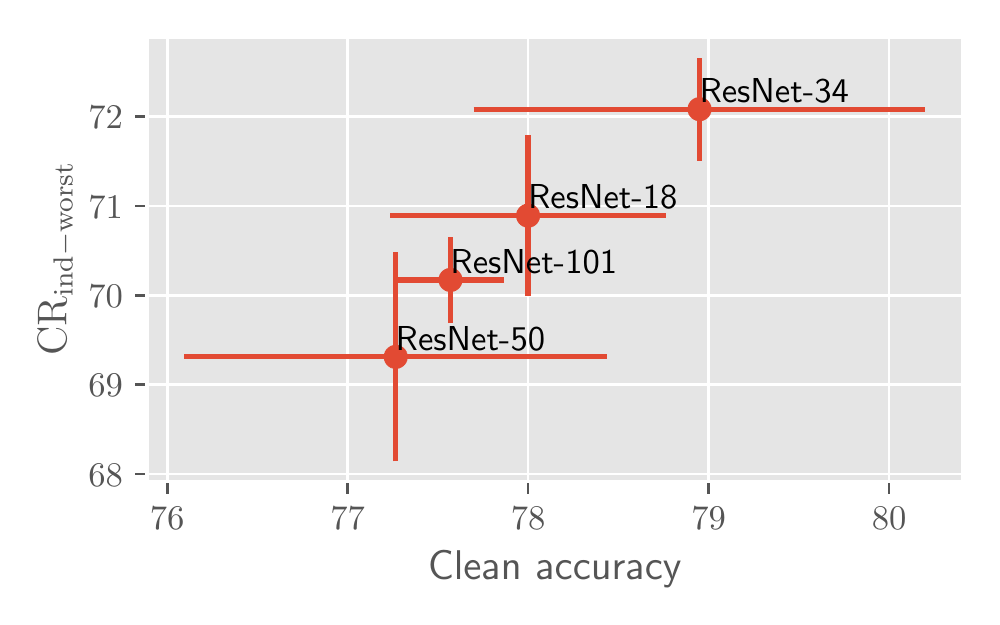}
    \caption{$\text{CR}_{\text{ind-avg}}$}
    \label{subfig:sat_arch_avg}
    \end{subfigure}
    \begin{subfigure}[t]{0.3\textwidth}
    \includegraphics[width=\textwidth]{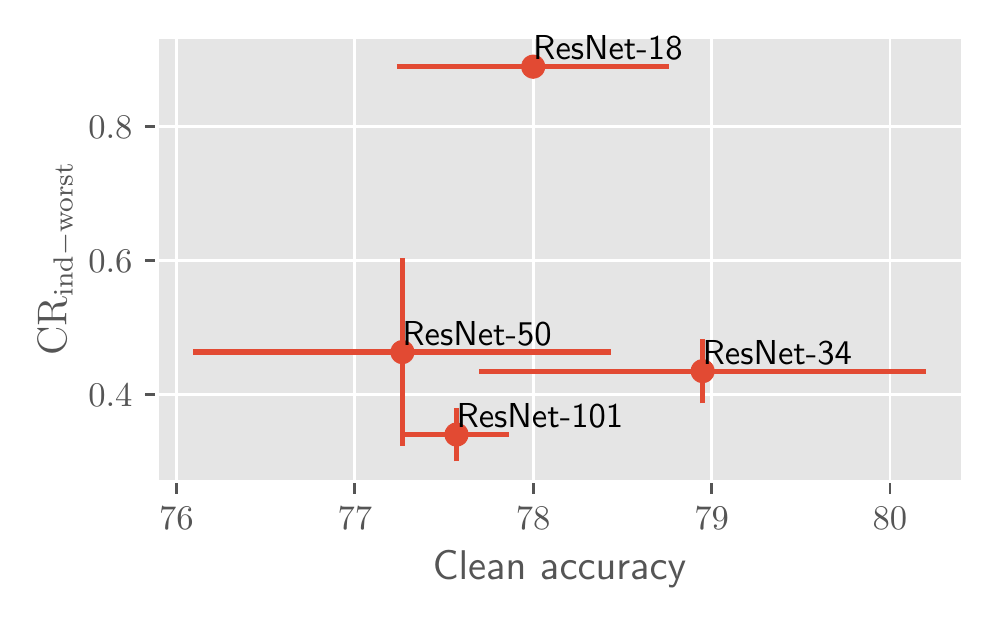}
    \caption{$\text{CR}_{\text{ind-worst}}$}
        \label{subfig:sat_arch_worst}
    \end{subfigure}
    \begin{subfigure}[t]{0.3\textwidth}
    \includegraphics[width=\textwidth]{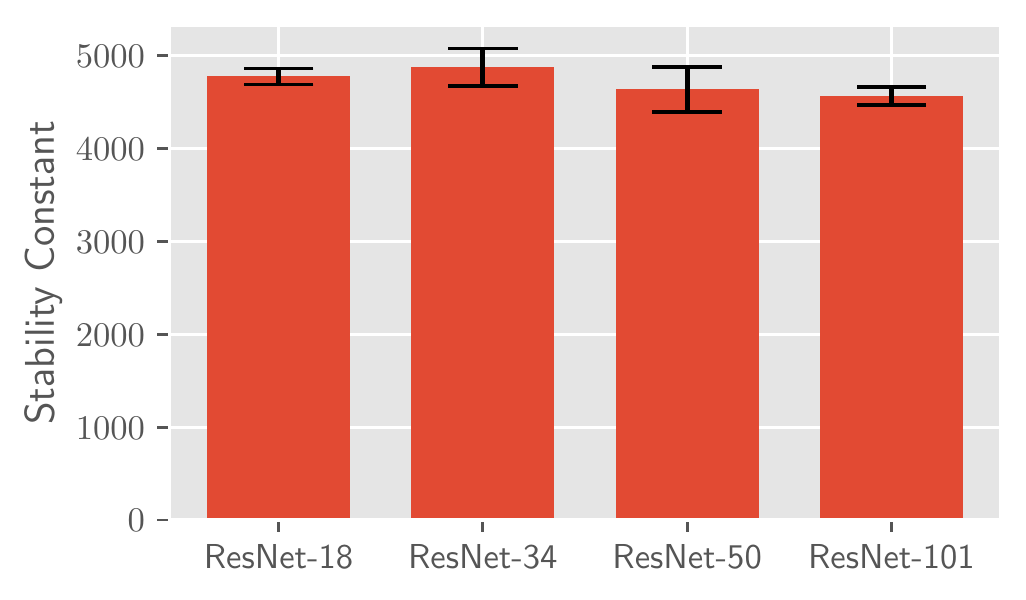}
    \caption{Stability Constant}
    \label{subfig:sat_arch_stab}
    \end{subfigure}
    \caption{\noindent \textbf{Impact of architecture size.} Figures (a) and (b): Clean accuracy vs CR for models trained  using stochastic adversarial training \citep{madaan2020learning}. Results are averaged over 3 trials and error bars are shown.  Higher values of CR indicate better performance. Figure (c): SC computed for models of each architecture. Lower SC indicates better performance.}
    \label{fig:sat_arch_impact}
\end{figure*}

\textbf{Impact of architecture size} In Figure \ref{fig:sat_arch_impact}, we plot the CR, clean accuracy, and stability constant achieved by training ResNet-18, ResNet-34, ResNet-50, and ResNet-101 models.  Similar to trends for training with LPIPS and training with $\ell_{\infty}$ threat model, we find that smaller architectures (ResNet-18, ResNet-34) outperform larger models in terms of CR (with ResNet-34 performing best in $\text{CR}_{\text{ind-avg}}$ and ResNet-18 performing best in $\text{CR}_{\text{ind-worst}}$) suggesting that smaller models are better when it comes to multiattack robustness when training with the union of $\ell_1$, $\ell_2$, and $\ell_{\infty}$ attacks.  In terms of stability constant, we do not see a significant trend across architecture size.

\begin{figure*}[ht]
    \centering
     
    \begin{subfigure}[t]{0.34\textwidth}
    \includegraphics[width=\textwidth]{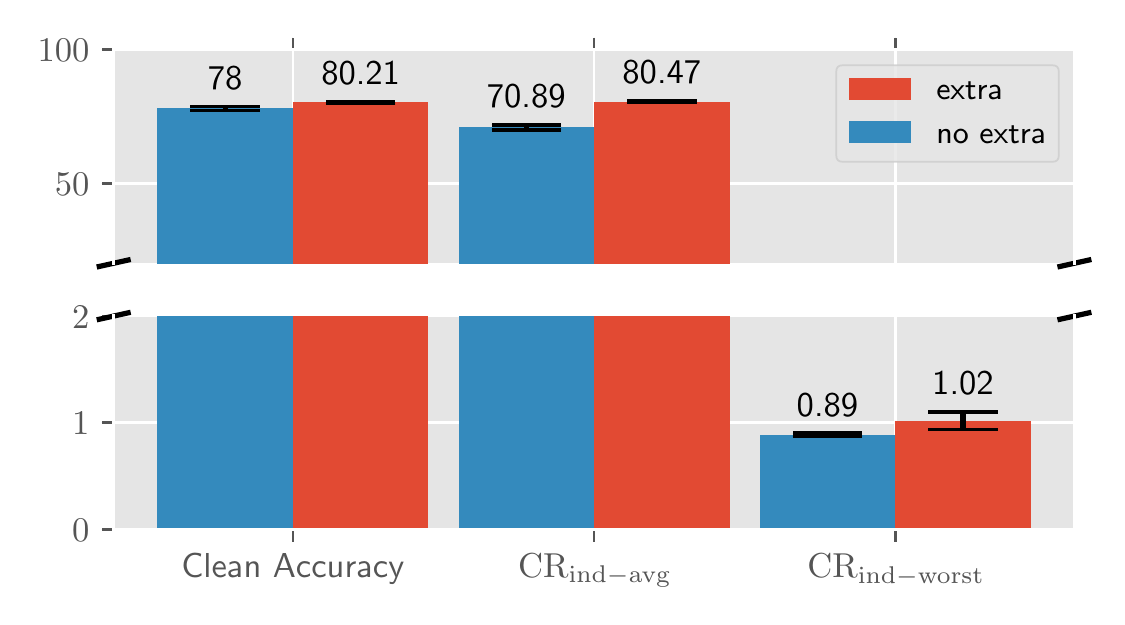}
    \caption{Clean Accuracy and CR}
    \label{subfig:sat_extra_avg}
    \label{subfig:sat_extra_worst}
    \end{subfigure}
    \begin{subfigure}[t]{0.3\textwidth}
    \includegraphics[width=\textwidth]{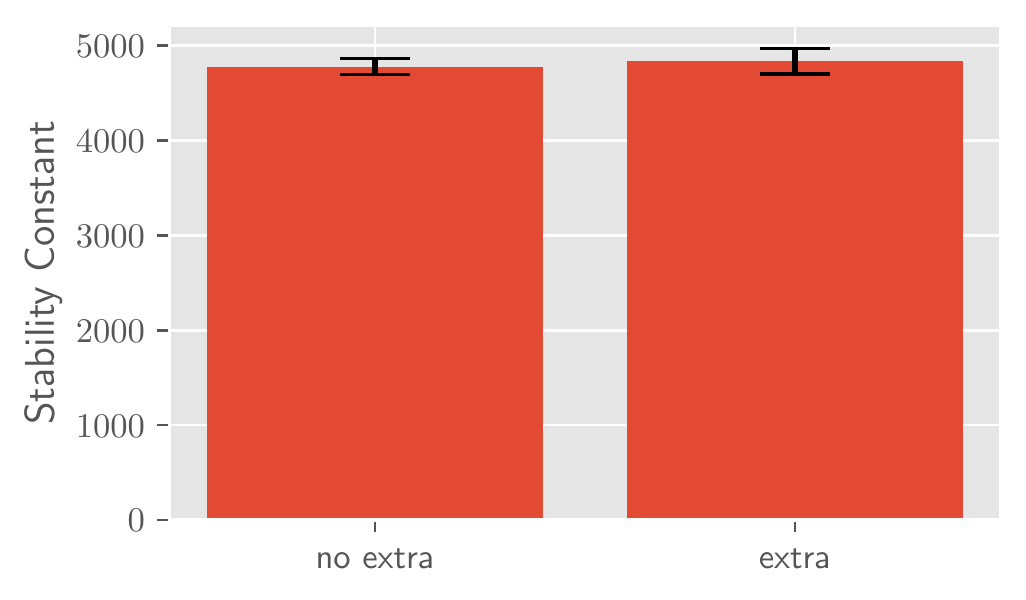}
    \caption{Stability Constant}
    \label{subfig:sat_extra_stab}
    \end{subfigure}
    \caption{
    \noindent \textbf{Impact of additional training data.} Figure (a): Clean accuracy and CR for ResNet-18 models trained using stochastic adversarial training \citep{madaan2020learning}.  Higher CR indicates better performance.  Results are averaged over 3 trials and error bars are shown. Figure (c): SC computed for models with and without additional training data. Lower SC indicates better performance.}
    \label{fig:sat_extra_impact}
\end{figure*}

\textbf{Impact of additional training data} In Figure \ref{fig:sat_extra_impact}, we plot the clean accuracy, CR, and stability constant of ResNet-18 models trained on the union of $\ell_1$, $\ell_2$, and $\ell_{\infty}$ attacks via stochastic adversarial training \citep{madaan2020learning}.  Similar to what we observed for other threat models used with adversarial training, we find that extra data significantly improves $\text{CR}_{\text{ind-avg}}$, suggesting that extra data improves average robust performance over the set of attacks.  We also find that for this threat model, extra data also improves $\text{CR}_{\text{ind-worst}}$, which differs from the trends observed for other threat models. There does not seem to be a significant change to stability constant for this training procedure with extra data.

\begin{figure*}[ht]
    \centering
     
    \begin{subfigure}[t]{0.3\textwidth}
    \includegraphics[width=\textwidth]{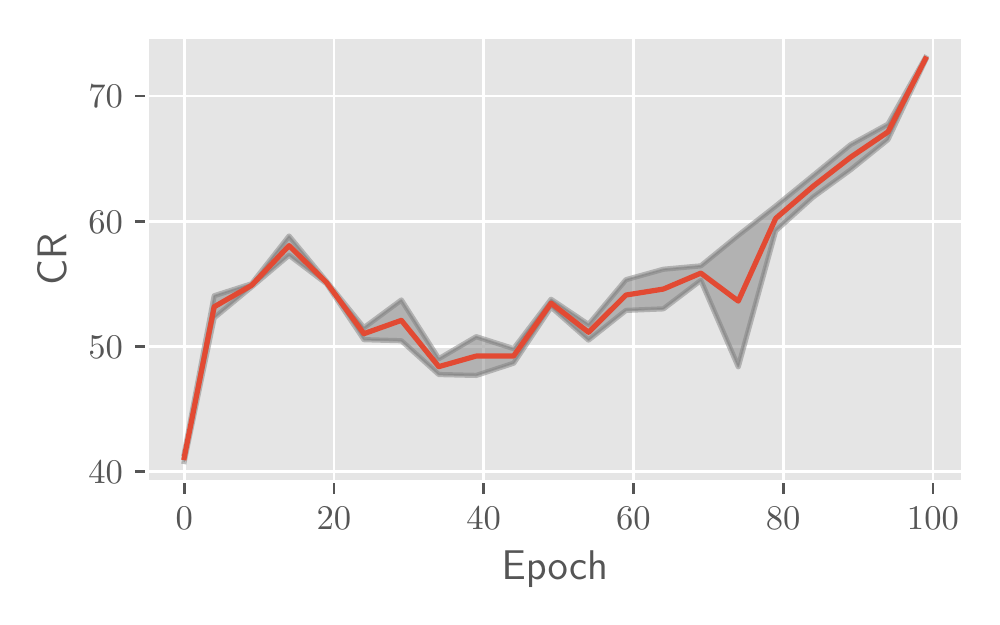}
    \caption{$\text{CR}_{\text{ind-avg}}$}
    \label{subfig:sat_es_avg}
    \end{subfigure}
    \begin{subfigure}[t]{0.3\textwidth}
    \includegraphics[width=\textwidth]{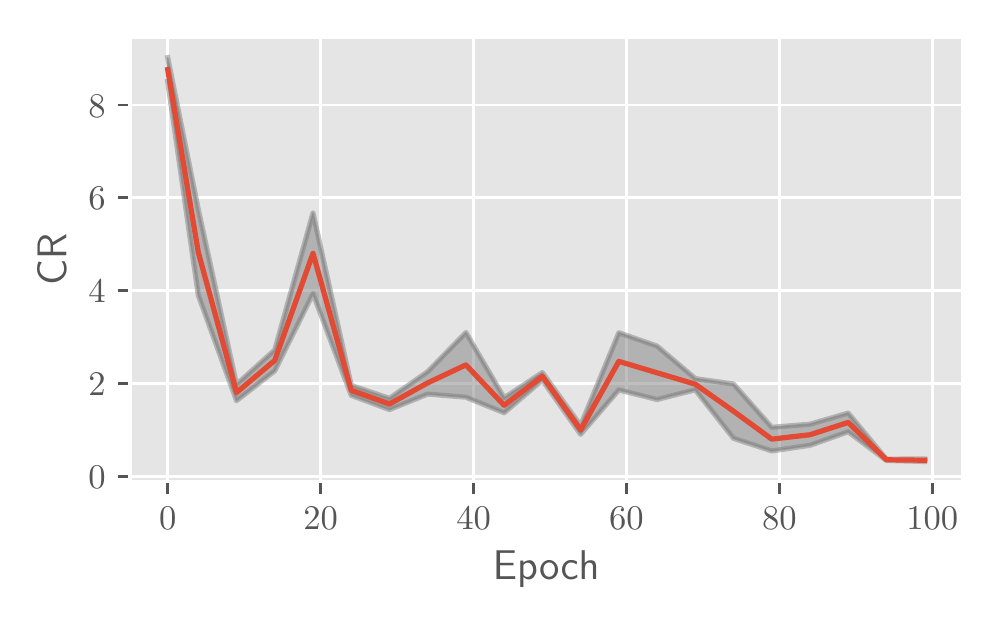}
    \caption{$\text{CR}_{\text{ind-worst}}$}
        \label{subfig:l2_es_worst}
    \end{subfigure}
    \begin{subfigure}[t]{0.3\textwidth}
    \includegraphics[width=\textwidth]{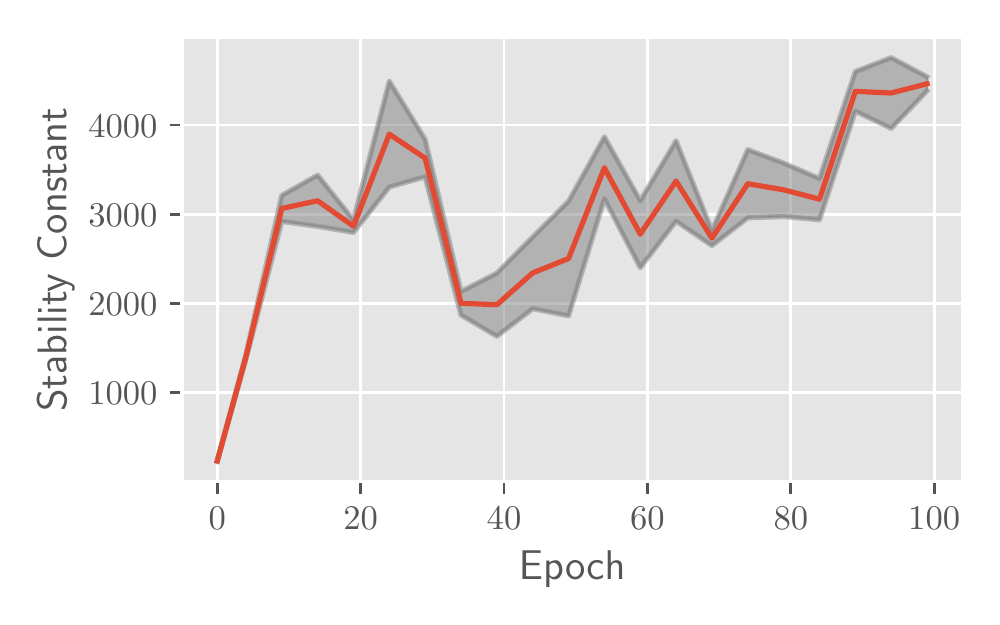}
    \caption{Stability Constant}
    \label{subfig:l2_es_stab}
    \end{subfigure}
    \caption{\noindent \textbf{Impact of number of training epochs.} CR and SC over epoch for models trained with stochastic adversarial training \citep{madaan2020learning} The red line indicates the average over 3 runs while the grey band highlights indicate 1 standard deviation from the mean. Higher values of CR and lower values of SC indicate better performance.}
    \label{fig:sat_early_stop_impact}
\end{figure*}

\textbf{Impact of number of epochs} In Figure \ref{fig:sat_early_stop_impact}, we plot CR and stability constant over training epochs.  Similar to trends for training on other threat models, we find that more training increases $\text{CR}_{\text{ind-avg}}$, suggesting better average case performance.  Additionally, we find that $\text{CR}_{\text{ind-worst}}$ drops quickly within the first 10 epochs of training and then remains gradually decreases throughout the remainder of training.  For stability, we find that stability constant gradually increases throughout training suggesting that as training progresses, the change in performance across threat models increases.

\end{document}